\documentclass{defaltmlr}

\usepackage{amsmath,amsfonts,bm}

\def\eqref#1{equation~\ref{#1}}

\def\1{\bm{1}}

\DeclareMathAlphabet{\mathsfit}{\encodingdefault}{\sfdefault}{m}{sl}
\SetMathAlphabet{\mathsfit}{bold}{\encodingdefault}{\sfdefault}{bx}{n}

\usepackage{url}
\usepackage[table]{xcolor}
\usepackage[dvipsnames]{xcolor}
\usepackage{graphicx}
\usepackage{geometry}
\usepackage{booktabs}
\usepackage[T1]{fontenc}
\usepackage{fontawesome}
\usepackage{fancyhdr}
\usepackage{tocloft}
\usepackage{enumitem}
\usepackage{etoc}
\usepackage{titletoc}
\usepackage{tcolorbox}
\usepackage{listings}
\newcommand\DoToC{%
  \startcontents
  \printcontents{}{1}{\noindent \textbf{\large{Table of Contents}}\vskip3pt\vskip5pt}
  \vskip3pt\vskip5pt
}

\usepackage{color}
\usepackage[dvipsnames]{xcolor}
\definecolor{JalapenoRed}{RGB}{183,21,64}
\definecolor{Belize}{RGB}{41,128,185}
\definecolor{Amour}{RGB}{238,82,83}
\usepackage{colortbl}

\newcommand{\High}{\textcolor[RGB]{247,85,115}{\gcmark~}}
\newcommand{\Medium}{\textcolor[RGB]{100,188,216}{$\triangle$~}}
\newcommand{\Low}{\textcolor[RGB]{117,196,119}{\rxmark~}}

\definecolor{myred}{RGB}{184,26,15}
\definecolor{mydarkred}{rgb}{0.6,0,0}
\definecolor{myblue}{HTML}{268BD2}
\definecolor{darkblue}{RGB}{0,20,115}
\definecolor{mybrown}{HTML}{B34D00}

\definecolor{bestcolor}{HTML}{2263aa}
\definecolor{secondcolor}{HTML}{2da02d}

\usepackage{natbib}
\usepackage{flushend}
\usepackage{subcaption}
\usepackage{caption}
\usepackage{tabularx}
\usepackage{graphicx}
\usepackage{amsfonts}
\usepackage{amsmath}
\usepackage{amsthm}
\usepackage{algorithm}
\usepackage{algorithmic}
\usepackage{multirow}
\usepackage{xcolor}
\usepackage{booktabs}
\usepackage{float}
\usepackage{balance}
\usepackage{enumitem}
\usepackage{pgfplots}
\setlist[itemize]{noitemsep, topsep=0pt, partopsep=0pt}

\usepackage{xspace}
\usepackage{threeparttable}
\usepackage{url}
\usepackage{makecell}
\usepackage{titletoc}

\usepackage{arydshln}

\usepackage{fontawesome}
\usepackage{pifont}

\usepackage{float}

\usepackage{hyperref}
\usepackage{cleveref}

\newcommand{\etc}{\textit{e}\textit{t}\textit{c}.}

\def\model{\texttt{ST-Booster}~}
\def\AirNineteen{\textit{AQI-19}}
\def\AirTwenty{\textit{AQI-22}}
\def\SpeedNineteen{\textit{Speed-19}}
\def\TrafficTwenty{\textit{Intensity-22}}
\newcommand{\firstres}[1]{{\textcolor{bestcolor}{\textbf{#1}}}}
\newcommand{\secondres}[1]{{\textcolor{secondcolor}{\underline{#1}}}}
\newcommand{\gcmark}{\textcolor[RGB]{18,220,168}{\checkmark}}
\newcommand{\rxmark}{\textcolor[RGB]{202,12,22}{\ding{55}}}

\newcommand{\ie}{\textit{i}.\textit{e}.}
\newcommand{\eg}{\textit{e}.\textit{g}.}

\newtheorem*{Pro*}{Problem}

\def\model{\texttt{{ExoST}}\xspace}

\usepackage{wrapfig}

\hypersetup{linkcolor=cyan}
\usepackage{mathtools}
\usepackage{ulem}
\usepackage{pifont}
\usepackage{bbding}

\usepackage{tcolorbox}
\usepackage{makecell}

\makeatletter
\let\orig@fnsymbol\@fnsymbol
\def\@fnsymbol#1{\ifcase#1\or\relax\else\orig@fnsymbol{#1}\fi}
\makeatother

\title{Select, then Balance: \\Exploring Exogenous Variable Modeling of Spatio-Temporal Forecasting}
\vspace{4cm}

\author{
\parbox{\textwidth}{
Wei Chen\textsuperscript{1,2}, Yuqian Wu\textsuperscript{1}, Yuanshao Zhu\textsuperscript{3}, Xixuan Hao\textsuperscript{2}, Shiyu Wang\textsuperscript{4}, Xiaofang Zhou\textsuperscript{2}, Yuxuan Liang$^*$\textsuperscript{1}
}
}

\affiliation{\textsuperscript{1}HKUST(GZ), \textsuperscript{2}HKUST, \textsuperscript{3}CityU, \textsuperscript{4}Byte Dance}

\abstract{
Spatio-temporal (ST) forecasting is critical for dynamic systems, yet existing methods predominantly rely on modeling a limited set of observed target variables. In this paper, we present the first systematic exploration of exogenous variable modeling for ST forecasting, a topic long overlooked in this field. We identify two core challenges in integrating exogenous variables: the inconsistent effects of distinct variables on the target system and the imbalance effects between historical and future data. To address these, we propose \model, a simple yet effective \underline{exo}genous variable modeling general framework highly compatible with existing \underline{ST} backbones that follows a ``select, then balance'' paradigm. Specifically, we design a latent space gated expert module to dynamically select and recompose salient signals from fused exogenous information. Furthermore, a siamese dual-branch backbone architecture captures dynamic patterns from the recomposed past and future representations, integrating them via a context-aware weighting mechanism to ensure dynamic balance. Extensive experiments on real-world datasets demonstrate the \model’s effectiveness, universality, robustness, and efficiency.
}

\correspondence{onedeanxxx@gmail.com, $^*$yuxliang@outlook.com}
\date{\sffamily September 06, 2025}

\begin{document}

\maketitle

\makeatletter
\let\@fnsymbol\orig@fnsymbol
\makeatother

\section{Introduction}

Spatio-temporal forecasting aims to learn the evolution of structured systems whose states vary across both space and time~\citep{shao2024exploring,chen2025eac,chen2025learning}. This task is fundamental to traffic management~\citep{avila2020data}, environmental monitoring~\citep{liang2023airformer}, and meteorology~\citep{bi2023accurate}. Recent advances have focused on spatio-temporal (ST) encoders~\citep{jin2023spatio,jin2024survey} that integrate graph neural operators~\citep{yu2017spatio,ma2025less} for learning spatial dependencies and sequence operators~\citep{wu2019graph,guo2019attention,gao2023spatio,han2024bigst} for modeling temporal patterns. \textit{However, existing methods predominantly focus on the dynamics of target variables, largely neglecting a critical source of auxiliary information: spatio-temporal exogenous variables.}

\begin{figure*}[t!]
  \centering
  \includegraphics[width=1.0\textwidth]{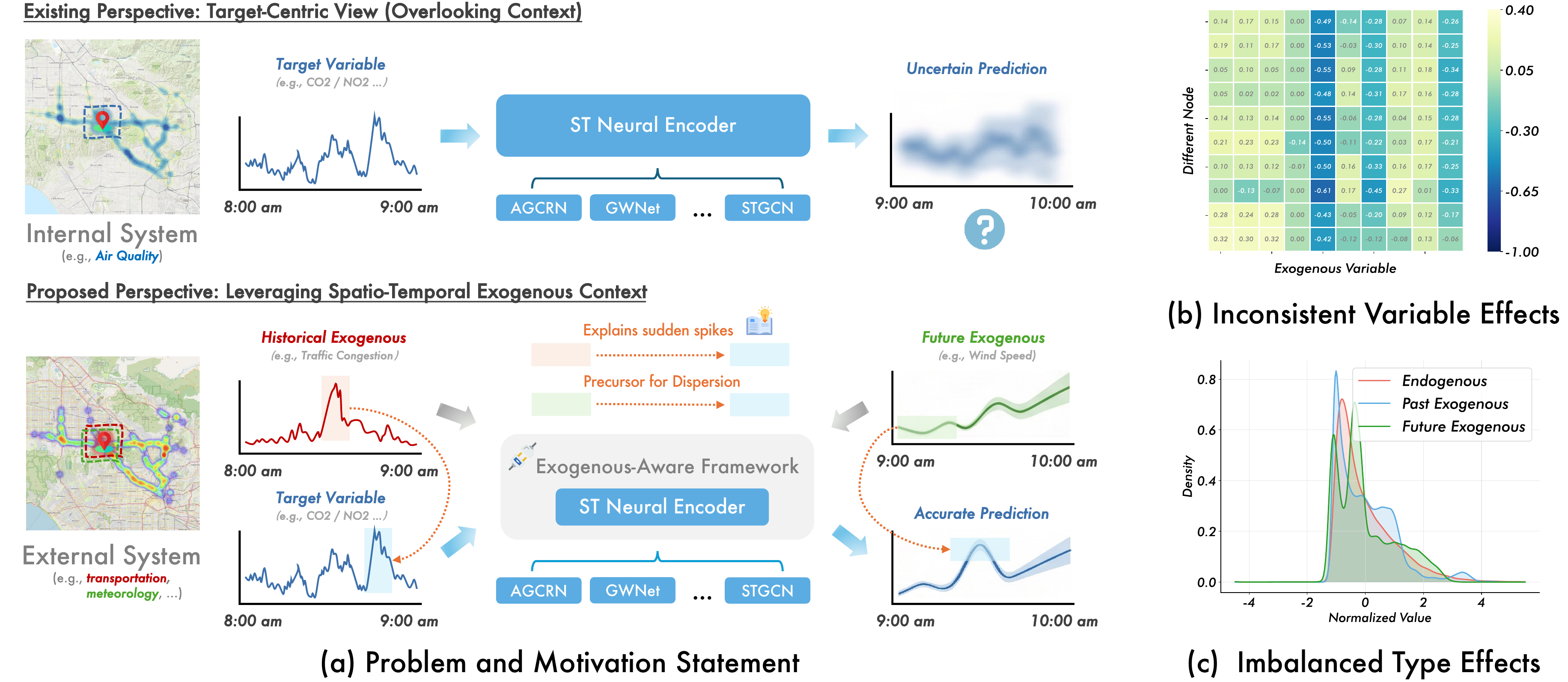}
  \vspace{-6mm}
  \caption{(a): Comparison between the target-centric view and the proposed exogenous-aware framework for ST forecasting. (b) and (c): A case study of inconsistent variable effects and imbalanced type effects in the \AirNineteen~dataset.}
  \label{fig:motivation_challenge}
  \vspace{-3mm}
\end{figure*}

Spatio-temporal exogenous variables encompass contextual data from external dynamic systems~\citep{berkani2023spatio}, including historical covariates and auxiliary future forecasts. These signals complement target predictions by reducing uncertainty and enhancing interpretability~\citep{olivares2023neural,wang2024timexer,huang2025exploiting,Arango2025ChronosX}. As illustrated in Figure~\ref{fig:motivation_challenge}~(a), air quality forecasting relies not only on historical pollutant indicators (\eg, NO$_2$, CO$_2$) but also on the environmental context. For instance, historical traffic congestion may explain sudden emission spikes, while future weather forecasts (\eg, wind speed) serve as precursors for pollutant dispersion or accumulation.

Despite the promise of exogenous information, their systematic utilization in ST forecasting remains surprisingly overlooked. We attribute this gap to two primary challenges:

\textit{Challenge \uppercase\expandafter{\romannumeral1}: Inconsistent Variable Effects:}
Exogenous variables exhibit heterogeneous correlations with prediction targets. As shown in Figure~\ref{fig:motivation_challenge}~(b)~(Pearson correlation coefficients between exogenous variables and the target variable across randomly stations in \AirNineteen), variables vary from highly predictive (\eg, traffic flow) to irrelevant or noisy (\eg, traffic incidents). Existing approaches typically employ a uniform concatenation strategy~\citep{giganti2024back, Arango2025ChronosX}, which fails to distinguish informative signals from noise. This lack of selective modeling limits representation capability and degrades predictive accuracy.

\textit{Challenge \uppercase\expandafter{\romannumeral2}: Imbalanced Type Effects}: In addition to inter-variable disparities, distinct distributional asymmetry exists between historical signals and proactive future forecasts. Figure~\ref{fig:motivation_challenge}~(c)~highlight the disparate distribution patterns of these data types in \AirNineteen. Current methods often rely on joint modeling or simple weighted fusion~\citep{das2023long,olivares2023neural,wang2024timexer,zhou2025crosslinear}, which fail to effectively decouple and balance the different impacts of past and future contexts on the target system.

To address these limitations, we propose \model, a novel framework governed by a ``select-then-balance'' paradigm designed to systematically exploit exogenous variables. \model features a modular design compatible with diverse spatio-temporal backbones. We introduce a Latent Space Gated Expert Module that projects endogenous and exogenous variables into a shared latent space, utilizing a gating mechanism to adaptively select informative factors while suppressing noise. To mitigate type imbalance, we design a dual-branch Siamese architecture that decouples the modeling of past and future exogenous contexts. These representations are processed by dual backbones and integrated via a context-aware balancer to generate the final prediction.

In summary, our key contributions are as follows:

\begin{itemize}[leftmargin=2.5mm,parsep=2.5pt]
\item \textit{Problem Level:} We present the first systematic formalization and exploration of exogenous variable modeling for ST forecasting, addressing a long-neglected gap therein.
\item \textit{Technical Level:} We identify the key challenges in modeling exogenous variables and propose \model, a simple yet effective general framework with targeted modules. \model demonstrates high compatibility with existing spatio-temporal backbones without any modification.
\item \textit{Evaluation Level:} We systematically evaluate \model across multiple aspects, verifying its efficacy, universality, robustness, design rationality, efficiency, and lightweight features. 
\end{itemize}

\section{Related Work}

\textbf{Spatio-Temporal Forecasting.}
Predicting future spatio-temporal signals from historical observations, a spatially extension of multivariate time series forecasting, has evolved significantly. Early statistical~\citep{box1970distribution,biller2003modeling} or independent time series models~\citep{nie2023time,wu2023timesnet,qiu2025duet} at each location proved insufficient for complex spatio-temporal dynamics~\citep{shao2024exploring}, which involve intricate spatial correlations and heterogeneity~\citep{chen2025eac,chen2025learning}. This led to hybrid architectures~\citep{li2017diffusion,wu2019graph,liu2023spatio,chen2025information,cao2025spatiotemporal,ma2025robust} that integrate spatial (\eg, graph operators for node interactions) and temporal (\eg, sequence operators for long-term trends) modeling to jointly learn rich spatio-temporal representations for accurate prediction. \textit{However, a critical limitation persists: these models largely focus on the system's internal dynamics while neglecting exogenous variables, external inputs that are independent of the system's state but directly influence its evolution, and whose inclusion is vital for robust and accurate predictions.}

\textbf{Exogenous Variables Modeling.}
Distinct from our ST forecasting setting, which models multiple systems as exogenous variables, existing literature primarily focuses on simple covariate modeling within multivariate time series forecasting to capture correlations between observations and target variables. Early statistical methods, such as ARIMAX~\citep{williams2001multivariate} and SARIMAX~\citep{vagropoulos2016comparison}, addressed the relationships among exogenous series, endogenous series, and autoregressive components. More recent deep learning approaches have investigated utilizing future exogenous values~\citep{das2023long,olivares2023neural}, handling time lags and missing data ~\citep{wang2024timexer,zhou2025crosslinear}, and leveraging language models to align semantic information for representation enhancement~\citep{liu2024spatial,huang2025exploiting}. While limited research integrates exogenous variables into ST forecasting, as summarized in Table~\ref{tab:comparison}, these efforts are restricted to specific ST tasks and model architectural designs~\citep{lindstrom2014flexible,zagouras2015role,dong2023spatiotemporal,giganti2024back,ma2025causal,ma2025spatiotemporal,wang2025unifying}. \textit{In contrast, we provide the first systematic formalization and exploration of modeling exogenous variables within this context, proposing a simple yet effective general framework.}

\begin{table*}[t!]
\centering
\renewcommand{\arraystretch}{1.4}
\caption{The advantages and differences of \model compared with existing methods. \High: Good, \Medium: Medium, \Low: Poor.}
\label{tab:comparison}
\resizebox{\linewidth}{!}{
\begin{tabular}{ccccccc}
\toprule
\textbf{Category} & \textbf{Example Works} & \textbf{Typical Integration} & \textbf{Universality} & \textbf{Simplicity} & \textbf{Robustness} & \textbf{Efficiency} \\
\cmidrule(lr){1-1}\cmidrule(lr){2-7}
\makecell[c]{Statistical Models} & ARIMAX~\citeyearpar{williams2001multivariate}, SARIMAX~\citeyearpar{vagropoulos2016comparison} & Linear Regression & \Low~(Strict Assump.) & \High & \Low & \Medium \\
\hline
\makecell[c]{Specific TS Models} & TiDE~\citeyearpar{das2023long}, TimeXer~\citeyearpar{wang2024timexer} & Concatenation / Cross-Attn & \Low~(TS specific) & \Low & \Medium & \High \\
\hline
\makecell[c]{LLM-based Models} & ST-LLM~\citeyearpar{liu2024spatial}, ExoLLM~\citeyearpar{huang2025exploiting} & Semantic Embedding & \High(Semantic Bridge) & \Low & \Medium & \Low \\
\hline
\makecell[c]{Specific ST Models} & SEEformer~\citeyearpar{dong2023spatiotemporal}, MAGCRN~\citeyearpar{giganti2024back} & Task-Specific Fusion & \Low~(ST specific) & \Low & \Medium & \High \\
\hline
\rowcolor{gray!8} General ST Framework & \textbf{\model (Ours)} & Select, then Balance & \High~(Any ST Backbone) & \High & \High & \High \\
\bottomrule
\end{tabular}
}
\vspace{-4mm}
\end{table*}

\section{Preliminaries}

\textbf{Definition 1.} \textit{(Spatio-Temporal Data).}
Spatio-temporal data records the evolution of dynamic systems across space and time. Formally, we represent this as a sequence $\mathcal{X}=\{X_t\}^{T}_{t=1}$, where ${X_t}\in{\mathbb{R}^{N \times F}}$ denotes the system state at time $t$. The spatial domain is modeled as a network $\mathcal{G}$ with $N$ nodes, where each node observes $F$ features (\eg, pollutant indicators such as NO$_2$, \etc~in air quality system).

\textbf{Definition 2.} \textit{(Exogenous Variables).}
Exogenous variables are external factors independent of the target system's internal dynamics, serving as auxiliary inputs. Denoted as $\mathcal{E}=\{E_t\}^{T}_{t=1}$, these variables share the structural format of ST data. $\mathcal{E}$ comprises past exogenous (\eg, historical traffic flow) and future exogenous variables (\eg, weather forecasts), distinguished by their temporal availability.

\textbf{Problem.} \textit{(ST Forecasting with Exogenous Variables).}
We aim to predict future spatio-temporal signals given historical observations, optional exogenous inputs, and a graph prior $\mathcal{G}$. Let $\mathcal{D} = \{(\mathbf{X}_i, \mathbf{Y}_i)\}_{i=1}^{|\mathcal{D}|} \sim \mathbb{P}$ denote the training set, where $\mathbf{X}_i \in \mathbb{R}^{N \times T_p \times F}$ and $\mathbf{Y}_i \in \mathbb{R}^{N \times T_f \times F}$ represent historical features (length $T_p$) and prediction targets (length $T_f$), respectively. Standard forecasting optimizes parameters $\theta^*$ by minimizing the expected loss:
$$
\theta^* = \arg\min_\theta \, \mathbb{E}_{(\mathbf{X}_{i}, \mathbf{Y}_{i}) \sim \mathbb{P}} \left[\mathcal{L}(f_\theta(\mathbf{X}_i, \mathcal{G}), \mathbf{Y}_i)\right],
$$
where $f_\theta$ is the model and $\mathcal{L}$ denotes the error metric. With exogenous inputs $\mathbf{E}_i = (\mathbf{E}_{i}^{p}, \mathbf{E}_i^{f})$, the objective changes to:
$$
\theta^* = \arg\min_\theta \, \mathbb{E}_{(\mathbf{X}_{i}, \mathbf{E}_{i}, \mathbf{Y}_{i}) \sim \mathbb{P}} \left[\mathcal{L}(f_\theta(\mathbf{X}_i, \mathbf{E}_i, \mathcal{G}), \mathbf{Y}_i)\right],
$$
where $\mathbf{E}_i^{p} \in \mathbb{R}^{N \times T_p \times F_p}$ and $\mathbf{E}_i^{f} \in \mathbb{R}^{N \times T_f \times F_f}$ correspond to past and future exogenous features with dimensions $F_p$ and $F_f$.
In practice, according to~\citep{jin2023spatio,jin2024survey,shao2024exploring}, the feature to be predicted is usually only the target variable, \ie, $F$ in $\mathbf{Y}_i$ is equal to 1.

\section{Methodology}\label{sec:method}

\begin{figure*}[htbp!]
  \centering
  \includegraphics[width=1.0\textwidth]{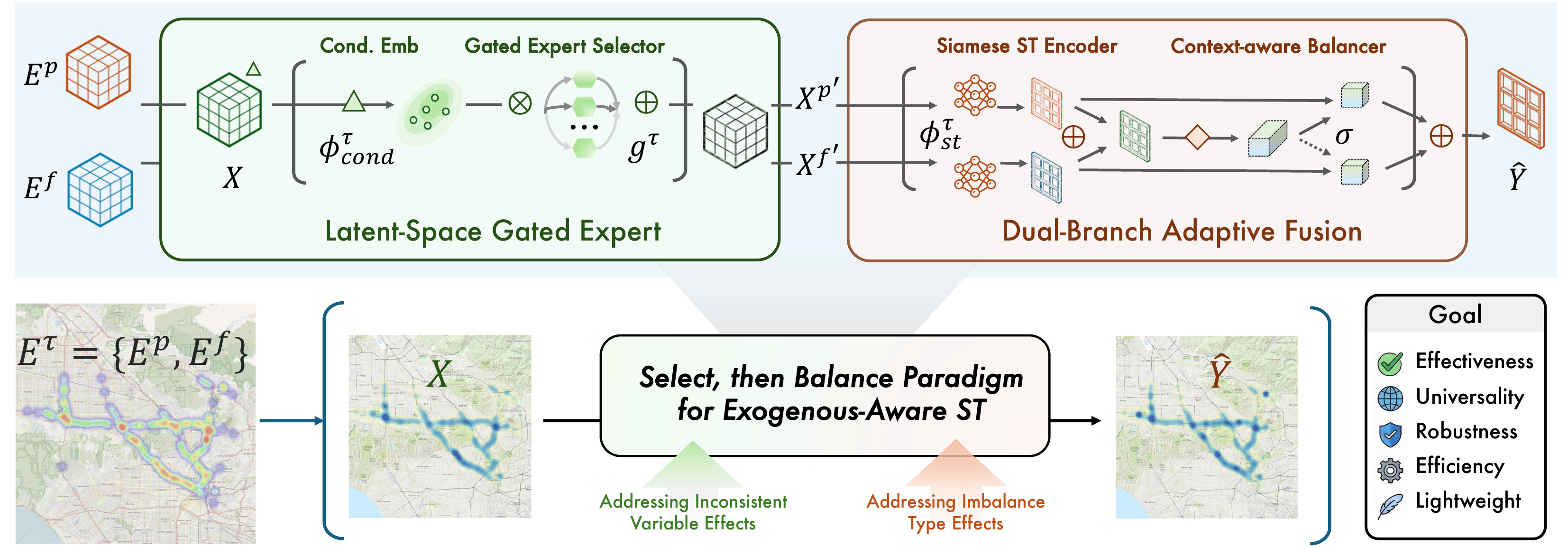}
  \vspace{-4mm}
  \caption{Overview of the proposed Exogenous-Aware Spatio-temporal forecasting framework \model.}
  \label{fig:framework}
  \vspace{-4mm}
\end{figure*}

The \model is built on the ``select-then-balance'' paradigm, as shown in Figure \ref{fig:framework}. The process consists of two consecutive stages, which are used to address the two core challenges in exogenous-aware ST forecasting mentioned above: (\romannumeral1) inconsistent variable effects and (\romannumeral2) unbalanced type effects. \textit{Both stages are fully differentiable and backbone-agnostic: any modern ST encoder can be plugged in without architectural change.}

\subsection{Select: Latent-Space Gated Expert}
\label{sec:select}

The selection stage aims to address inconsistent variable effects by extract compact, adaptive semantic representations from input endogenous variables and heterogeneous, multi-source exogenous variables, laying the foundation for the subsequent balancing stage. It operates in two key stages.

\textbf{Conditional Embedding.} Conditional embeddings specifically address the semantic space and dimension size alignment of endogenous variables with two types of exogenous variables. For each type $\tau \in \{p,f\}$ (\ie, \text{past}, \text{future}), we project the raw exogenous tensor $\mathbf{E}^{\tau}\in\mathbb{R}^{N\times T_{\tau}\times F_{\tau}}$ and historical input $\mathbf{X}\in\mathbb{R}^{N\times T_{p}\times F}$ into a condition latent space:
\begin{equation}
    \mathbf{X}^{\tau} = \phi_{\text{cond}}^{\tau}\!\left(\mathbf{X}, \mathbf{E}^{\tau}\right),
    \label{eq:cond_block}
\end{equation}
where $\phi_{\text{cond}}^{\tau}$ acts as an affine transformation block:
\begin{equation}
\phi_{\text{cond}}^\tau(\mathbf{X}, \mathbf{E}^\tau\bigr)
= \operatorname{Dropout}\Bigl(\operatorname{Act}\bigl(W_x^\tau \mathbf{X} + W_e^\tau \mathbf{E}^\tau + \mathbf{b}^\tau\bigr)\Bigr),
\label{eq:cond}
\end{equation}
with learnable projections $W_x^\tau\in\mathbb{R}^{H\times F}$ and $W_e^\tau\in\mathbb{R}^{H\times F_\tau}$. Following standard protocols~\citep{shao2024exploring,jin2024survey}, we assume uniform sequence lengths of past and future steps to enable direct element-wise addition. For more arbitrary cases, the zero-padding preprocessing trick can also easily ensure alignment. This process yields the conditional representations $\mathbf{X}^p, \mathbf{X}^f \in \mathbb{R}^{N \times T \times H}$.

\textbf{Gated Expert Selector.}
To mitigate semantic degeneration and channel entanglement arising from inconsistent variable effects, we incorporate a Gated Expert Selector module for each $\mathbf{X}^{\tau}$. Specifically, we employ $K$ expert projections $\{{W}_k^{\tau}\in\mathbb{R}^{H\times H}\}_{k=1}^{K}$ and a gating network $\mathbf{g}^{\tau}:\mathbb{R}^{H}\rightarrow\Delta^{K}$. The gating scores $\mathbf{g}^{\tau}\in\mathbb{R}^{K}$, which determine the contribution of each expert based on specific spatio-temporal contexts, are computed via:
\begin{equation}
    \mathbf{g}^{\tau} = \operatorname{Softmax}\!\left(W_g^{\tau}\,\mathbf{X}^{\tau}\right),
\end{equation}
where $W_g^{\tau}$ is a learnable projection matrix, along with a normalized activation function, calculates each expert's weight. The refined representation is subsequently synthesized through a weighted aggregation:
\begin{equation}
    \mathbf{X}^{{\tau}^{\prime}} = \sum_{k=1}^{K} g^{\tau}_{k}\; W_k^{\tau}\,\mathbf{X}^{\tau}.
    \label{eq:moe}
\end{equation}
This mechanism explicitly disentangles representations via a latent bottleneck and selective recombination, adaptively deriving robust inputs $\mathbf{X}^{p^{\prime}}$ and $\mathbf{X}^{f^{\prime}}$ for the backbone rather than relying on entangled mappings.

\textbf{Analysis.}
Unlike mixture-of-experts (MoEs) in modern language models~\citep{shazeer2017outrageously,xue2024openmoe} designed for efficient sparse parallelism, ours resembles early recommender-system MoEs~\citep{ma2018modeling} that capture diverse feature patterns (\eg, rain–rush-hour or fog–low-flow). It activates specific experts under local exogenous conditions, resolving inconsistency by decoupling conflicting signals and enabling specialized processing paths (More formal analysis in Appendix~\ref{appendix_analysis_select}). Fig. \ref{fig:gate_weight} shows the gating probability distributions across time steps in the \AirNineteen~experiment, highlighting the smooth composition of expert allocation. Empirically, we also validate the module's efficacy across diverse scenarios in Section.~\ref{mechanism_study}.

\begin{figure*}[htbp!]
    \centering
    \begin{minipage}[b]{0.49\columnwidth}
        \centering
        \includegraphics[width=\linewidth]{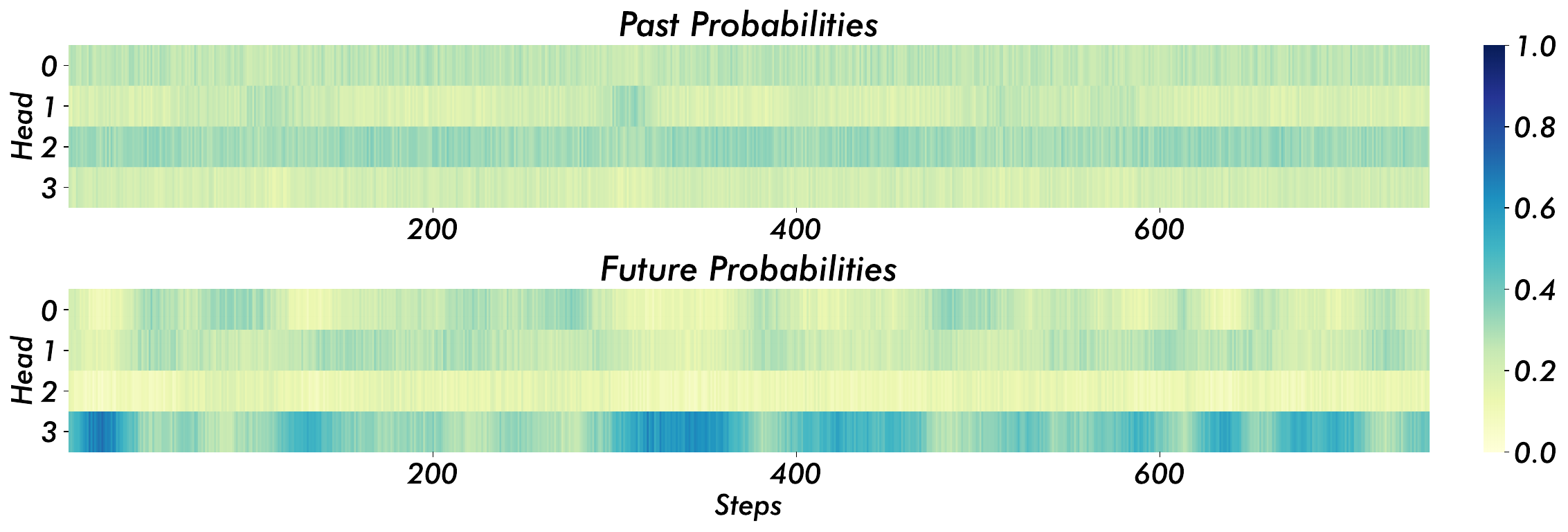}  
        \vspace{-6mm}
        \caption{Visualization of gated expert probabilities.}
        \label{fig:gate_weight}
    \end{minipage}
    \hfill  
    \begin{minipage}[b]{0.49\columnwidth}
        \centering
        \includegraphics[width=\linewidth]{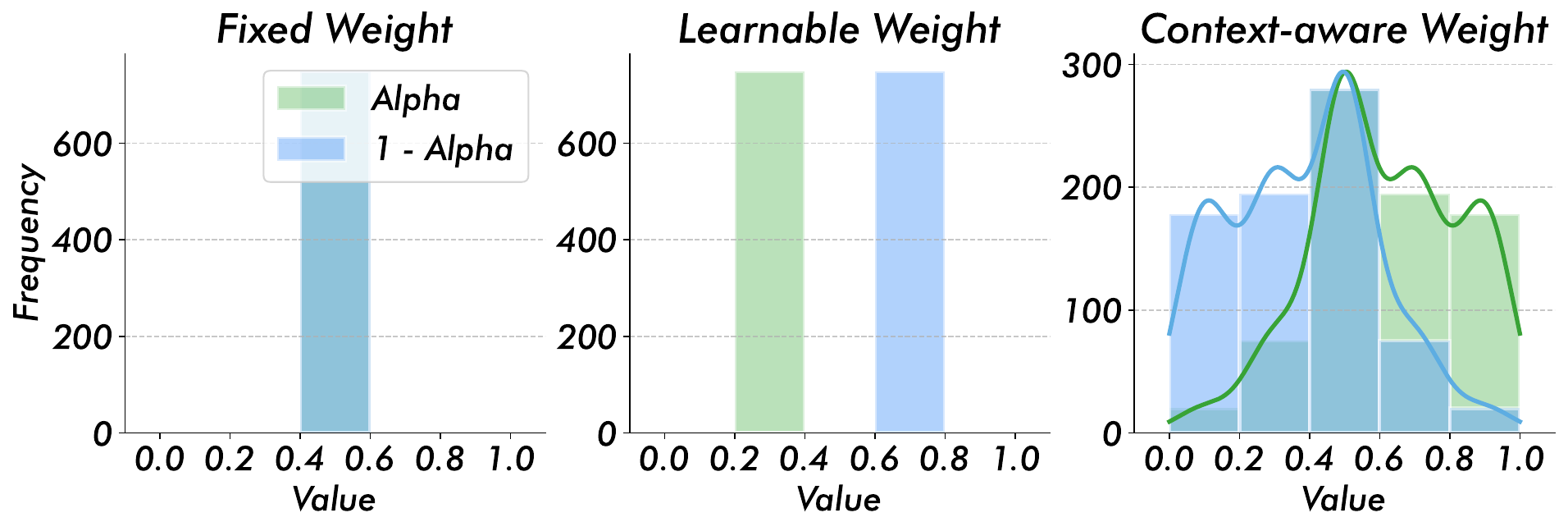}
        \vspace{-6mm}
        \caption{Weight distribution across different strategies.}
        \label{fig:balancer_weight}
    \end{minipage}
    \vspace{-4mm}
\end{figure*}

\subsection{Balance: Dual-Branch Adaptive Fusion}
\label{sec:balance}

The balancing stage addresses unbalanced type effects arising from the distributional discrepancies between past and future exogenous contexts. To effectively fuse these heterogeneous signals, we employ parallel Siamese networks for independent ST feature extraction, coupled with a context-aware balance mechanism for adaptive fusion.

\textbf{Siamese ST Encoders.} To process the distinct characteristics of past- and future-conditioned semantic representations, we employ a Siamese architecture comprising parallel encoders. Each encoder $\phi_{\text{st}}^{\tau}$, adaptable to arbitrary spatio-temporal backbones~\citep{yu2017spatio,li2017diffusion,guo2019attention,wu2019graph,Bai2020Adaptive,Gao2022On,Satorras2022Multivariate}, captures intrinsic spatial correlations and temporal dependencies via:
\begin{equation}
    \mathbf{Y}^{\tau} = \phi_{\text{st}}^{\tau}(\mathbf{X}^{{\tau}^{\prime}}),
    \label{eq:st_backbone}
\end{equation}
where $\mathbf{Y}^{p}, \mathbf{Y}^{f} \in \mathbb{R}^{N \times T_f \times 1}$ encode high-level spatio-temporal features derived from past and future exogenous contexts, respectively, facilitating subsequent fusion.

\textbf{Context-Aware Balancer.} To address the limitations of static fusion strategies, we employ a dynamic context-aware mechanism to integrate $\mathbf{Y}^{p}$ and $\mathbf{Y}^{f}$. We first construct a unified context tensor $\mathbf{Y}=\mathbf{Y}^{p} + \mathbf{Y}^{f}$, from which an instance-specific balancing weight $\alpha$ is derived via:
\begin{equation}
    \alpha = \sigma\left(\operatorname{MLP}\left( \operatorname{AvgPool}(\mathbf{Y})\right)\right).
    \label{eq:gate_alpha}
\end{equation}
Here, $\operatorname{AvgPool}(\cdot)$ aggregates spatial information to generate a channel descriptor, which passes through an MLP bottleneck with Sigmoid activation $\sigma(\cdot)$. The final prediction $\hat{\mathbf{Y}}$ is synthesized using a residual connection to facilitate information propagation:
\begin{equation}
    \hat{\mathbf{Y}} = \alpha\odot\mathbf{Y}^{p} + (1-\alpha)\odot\mathbf{Y}^{f} + \mathbf{Y}.
    \label{eq:fusion}
\end{equation}
This adaptive calibration effectively prioritizes past versus future contexts based on instance-specific dynamics, yielding a robust and balanced prediction.



\textbf{Analysis.}
Prior modeling of past and future exogenous variables typically relies on static fusion, employing either fixed linear weights or global affine transformations optimized over the training distribution~\citep{das2023long,giganti2024back,Arango2025ChronosX}. In contrast, our context-aware mechanism utilizes input-dependent non-linear mappings to approximate arbitrary fusion functions, enabling instance-specific adaptation during inference (More formal analysis in Appendix~\ref{appendix_analysis_balance}). Fig.~\ref{fig:balancer_weight} visualizes the weight distributions across inference samples in the \AirNineteen~experiment; unlike the scattered and irregular patterns observed in baselines, \model exhibits smoother and more uniform distributions, underscoring its superior dynamic adaptability. Empirically, we also validate the effectiveness of this design in Section.~\ref{mechanism_study}.

\section{Experiments}

In this section, we conduct extensive experiments to investigate the following core research questions:

\begin{itemize}[leftmargin=2.5mm,parsep=2pt] 
\item \textbf{RQ1: (Universality)} Can \model achieve consistent improvements on any existing ST backbones? 
\item \textbf{RQ2: (Effectiveness)} Can \model outperform existing other methods across various task scenarios? 
\item \textbf{RQ3: (Robustness)} Can \model maintain its performance when exogenous variables are missing or noisy? 
\item \textbf{RQ4: (Mechanism \& Rationality)} How does \model work? Which components or strategies are crucial, and are they sensitive to different designs or hyper-parameters? 
\item \textbf{RQ5: (Efficiency \& Lightweight)} How does \model compare in efficiency and parameter count to others? 
\end{itemize}

\subsection{Experimental Setup}
\noindent\textbf{Datasets.}
We evaluate on several public real-world spatio-temporal datasets~\footnote{\href{https://zenodo.org/records/7308425}{https://zenodo.org/records/7308425}}, containing hourly measurements from air quality, meteorological, and traffic monitoring stations in Madrid (1 January–30 June, 2019/2022). We study four forecasting tasks: (\romannumeral1) \AirNineteen~and \AirTwenty: predict $NO_{2}$ in air quality domain concentrations with traffic and meteorological domain variables as exogenous inputs; (\romannumeral2) \SpeedNineteen~and \TrafficTwenty: predict traffic speed / intensity using other system indicators as exogenous variables. Datasets are chronologically split into train/val/test with ratio 7:2:1. Date encodings are included as past and future exogenous variables. More dataset details are provided in Appendix~\ref{appendix_datasets}.

\noindent\textbf{Baseline.}
We compared with three comparable types of baselines in the literature: (\romannumeral1) time-series models with exogenous modeling: TiDE~\citep{das2023long}, NBEATSx~\citep{olivares2023neural}, TimeXer~\citep{wang2024timexer}, CrossLinear~\citep{zhou2025crosslinear}, DAG~\citep{qiu2025dagdualcausalnetwork}; (\romannumeral2) spatio-temporal models with exogenous integration: MAGCRN~\citep{giganti2024back}; (\romannumeral3) general exogenous modeling framework (for foundation model): ChronosX~\citep{Arango2025ChronosX}. To test \model's universality, we further instantiate it with various classic spatio-temporal backbones, including AGCRN~\citep{Bai2020Adaptive}, GWNet~\citep{wu2019graph}, GGNN~\citep{Satorras2022Multivariate}, GRUGCN~\citep{Gao2022On}, STGCN~\citep{yu2017spatio}, and DCRNN~\citep{li2017diffusion}. More details on baseline methods and backbone networks, please refer to Appendix~\ref{appendix_baseline}.

\noindent\textbf{Protocol.}
We use the past 24 steps to forecast the next 24 steps (\ie, 1-day) and report MAE, RMSE, MAPE, and MRE. To assess long-term robustness, we additionally perform 2- and 3-day rolling forecasts without retraining. Baseline models follow their recommended settings. For \model, we use 4 gated experts and hidden size 64. Each experiment is run five times; we report mean $\textcolor{gray}{\text{{$\pm$} standard deviation}}$. Best results appear in \firstres{bold blue}, second-best in \secondres{underlined green}. Full protocol details can be found in Appendix~\ref{appendix_protocol}.

\subsection{Universality Study (RQ1)}

\begin{figure*}[t!]
  \centering
  \includegraphics[width=1.0\textwidth]{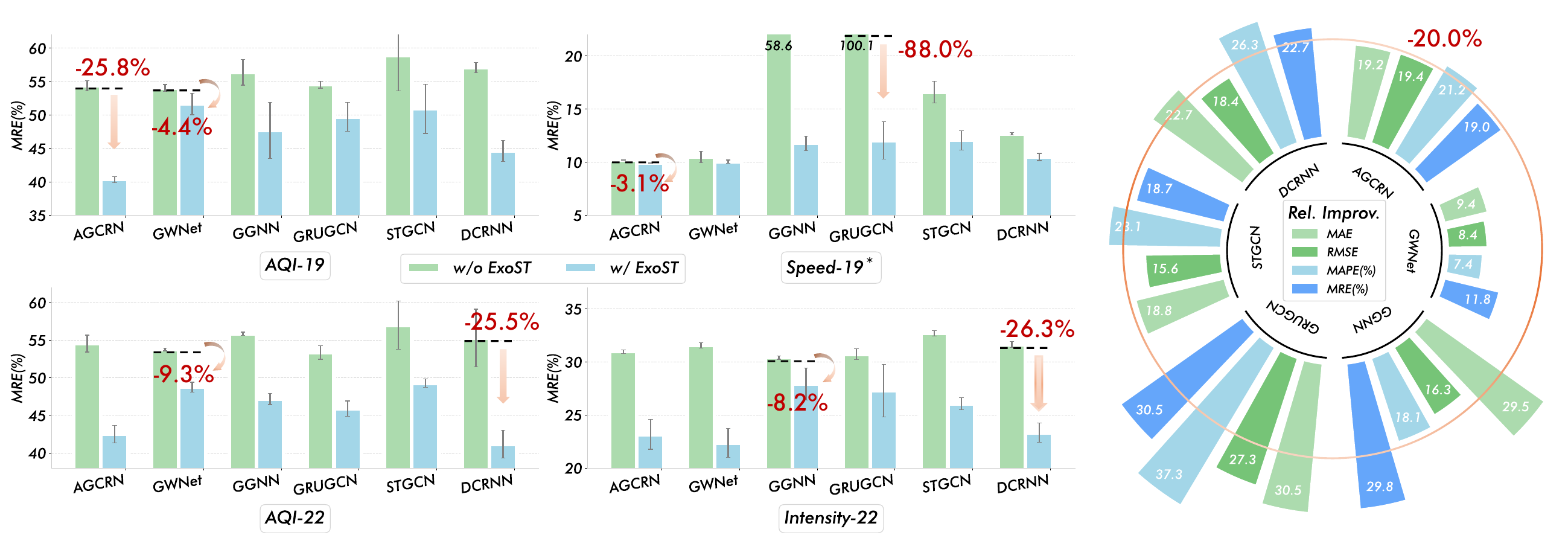}
  \vspace{-6mm}
  \caption{Left: MRE performance gains of different models on different datasets with and without \model. Right: Average relative improvement of different metrics for each model across all datasets (Full results in Appendix~\ref{appendix_universality} Table~\ref{tab:rq1_3day}).}
  \label{fig:combined_improve}
  \vspace{-5mm}
\end{figure*}

We first evaluate the universality of the ``select, then balance'' paradigm in \model across diverse ST backbones. Figure~\ref{fig:combined_improve} illustrates the absolute and average relative performance improvements of \model across four tasks (left) and six models (right) for 3-day forecasting. \ding{182} \model yields consistent performance boosts across varying architectures and datasets, validating the efficacy of our exogenous modeling paradigm. \ding{183} From the task perspective, air quality forecasting achieves substantial gains due to the strong influence of exogenous factors (\eg, traffic and meteorology), whereas traffic speed forecasting sees moderate improvements, reflecting a limited dependency on exogenous variables. \ding{184} From the model perspective, while gains vary across backbones, most exceed 20\%; notably, even the vanilla GWNet achieves a significant improvement of nearly 10\%. \ding{185} Supplementary results for 1-day and 2-day horizons (Tables~\ref{tab:rq1_2day} and~\ref{tab:rq1_1day} in Appendix \ref{appendix_universality}) further reveal that performance gains amplify as the forecasting horizon extends, underscoring the robust generalization capability of our approach.

\begin{table*}[t!]
\caption{Comparison of the overall performance of advance methods and \model.
For \SpeedNineteen~$^*$, MAE and RMSE are presented with a $10^2$ scale to facilitate detailed comparison. Best results in \firstres{bold blue}, second-best in \secondres{underlined green}.}
\label{tab:rq1_overall_performance}
\vspace{-2mm}
\centering
\small

\setlength{\tabcolsep}{2.6mm}
\renewcommand{\arraystretch}{1.2}
\resizebox{1.0\linewidth}{!}{
\begin{sc}
\begin{tabular}{llcccccccccccc}
\toprule

\multicolumn{2}{c}{\textbf{Datasets}} & 
\multicolumn{3}{c}{\textbf{AQI-19}}&\multicolumn{3}{c}{\textbf{Speed-19~$^*$}}&\multicolumn{3}{c}{\textbf{AQI-22}}&\multicolumn{3}{c}{\textbf{Intensity-22}} \\

\cmidrule(lr){1-2}\cmidrule(lr){3-5}\cmidrule(lr){6-8}\cmidrule(lr){9-11}\cmidrule(lr){12-14}

\textbf{Method} 
& \textbf{Metric} & 1-day & 2-day & 3-day & 1-day & 2-day & 3-day & 1-day & 2-day & 3-day & 1-day & 2-day & 3-day \\
\midrule

\multirow{4}{*}{\makecell[l]{\textbf{TiDE}\\\citeyearpar{das2023long}}}
& MAE 
& $14.73\textcolor{gray}{\text{\scriptsize±0.01}}$ & $15.68\textcolor{gray}{\text{\scriptsize±0.00}}$ & $15.66\textcolor{gray}{\text{\scriptsize±0.01}}$
& $6.70\textcolor{gray}{\text{\scriptsize±1.83}}$ & $7.43\textcolor{gray}{\text{\scriptsize±1.75}}$ & $7.58\textcolor{gray}{\text{\scriptsize±1.73}}$
& $10.51\textcolor{gray}{\text{\scriptsize±0.05}}$ & $11.32\textcolor{gray}{\text{\scriptsize±0.04}}$ & $11.89\textcolor{gray}{\text{\scriptsize±0.04}}$
& $150.18\textcolor{gray}{\text{\scriptsize±0.02}}$ & $157.00\textcolor{gray}{\text{\scriptsize±0.01}}$ & $161.17\textcolor{gray}{\text{\scriptsize±0.02}}$ \\

& RMSE
& $20.73\textcolor{gray}{\text{\scriptsize±0.03}}$ & $21.76\textcolor{gray}{\text{\scriptsize±0.04}}$ & $21.70\textcolor{gray}{\text{\scriptsize±0.04}}$
& $44.47\textcolor{gray}{\text{\scriptsize±5.16}}$ & $50.37\textcolor{gray}{\text{\scriptsize±4.21}}$ & $51.26\textcolor{gray}{\text{\scriptsize±4.10}}$
& $14.67\textcolor{gray}{\text{\scriptsize±0.00}}$ & $15.76\textcolor{gray}{\text{\scriptsize±0.01}}$ & $16.43\textcolor{gray}{\text{\scriptsize±0.01}}$
& $217.05\textcolor{gray}{\text{\scriptsize±0.15}}$ & $223.13\textcolor{gray}{\text{\scriptsize±0.16}}$ & $226.96\textcolor{gray}{\text{\scriptsize±0.17}}$ \\

& MAPE (\%)
& $93.01\textcolor{gray}{\text{\scriptsize±0.53}}$ & $98.80\textcolor{gray}{\text{\scriptsize±0.43}}$ & $98.60\textcolor{gray}{\text{\scriptsize±0.41}}$
& $16.38\textcolor{gray}{\text{\scriptsize±4.62}}$ & $18.02\textcolor{gray}{\text{\scriptsize±4.42}}$ & $18.37\textcolor{gray}{\text{\scriptsize±4.36}}$
& $87.51\textcolor{gray}{\text{\scriptsize±0.98}}$ & $95.07\textcolor{gray}{\text{\scriptsize±0.93}}$ & $99.22\textcolor{gray}{\text{\scriptsize±0.92}}$
& $146.50\textcolor{gray}{\text{\scriptsize±0.03}}$ & $147.09\textcolor{gray}{\text{\scriptsize±0.07}}$ & $150.07\textcolor{gray}{\text{\scriptsize±0.06}}$ \\

& MRE (\%)
& $56.96\textcolor{gray}{\text{\scriptsize±0.04}}$ & $60.67\textcolor{gray}{\text{\scriptsize±0.01}}$ & $60.67\textcolor{gray}{\text{\scriptsize±0.02}}$
& $17.44\textcolor{gray}{\text{\scriptsize±4.78}}$ & $19.33\textcolor{gray}{\text{\scriptsize±4.56}}$ & $19.72\textcolor{gray}{\text{\scriptsize±4.51}}$
& $56.27\textcolor{gray}{\text{\scriptsize±0.24}}$ & $60.56\textcolor{gray}{\text{\scriptsize±0.20}}$ & $63.62\textcolor{gray}{\text{\scriptsize±0.20}}$
& $54.43\textcolor{gray}{\text{\scriptsize±0.01}}$ & $56.82\textcolor{gray}{\text{\scriptsize±0.00}}$ & $58.21\textcolor{gray}{\text{\scriptsize±0.01}}$ \\

\cmidrule(lr){2-14}

\multirow{4}{*}{\makecell[l]{\textbf{NBEATSx}\\\citeyearpar{olivares2023neural}}}

& MAE
& $11.63\textcolor{gray}{\text{\scriptsize±0.08}}$ & $13.88\textcolor{gray}{\text{\scriptsize±0.06}}$ & $14.10\textcolor{gray}{\text{\scriptsize±0.09}}$
& $5.21\textcolor{gray}{\text{\scriptsize±0.30}}$ & $5.99\textcolor{gray}{\text{\scriptsize±0.29}}$ & $6.17\textcolor{gray}{\text{\scriptsize±0.28}}$
& $8.13\textcolor{gray}{\text{\scriptsize±0.11}}$ & $9.71\textcolor{gray}{\text{\scriptsize±0.19}}$ &
$9.97\textcolor{gray}{\text{\scriptsize±0.17}}$ &
$38.91\textcolor{gray}{\text{\scriptsize±1.68}}$ & $67.02\textcolor{gray}{\text{\scriptsize±1.78}}$ & $85.03\textcolor{gray}{\text{\scriptsize±1.43}}$ \\

& RMSE
& $18.44\textcolor{gray}{\text{\scriptsize±0.04}}$ & $20.65\textcolor{gray}{\text{\scriptsize±0.06}}$ & $21.20\textcolor{gray}{\text{\scriptsize±0.04}}$
& $39.56\textcolor{gray}{\text{\scriptsize±2.02}}$ & $45.80\textcolor{gray}{\text{\scriptsize±2.23}}$ & $46.80\textcolor{gray}{\text{\scriptsize±2.20}}$
& $12.84\textcolor{gray}{\text{\scriptsize±0.08}}$ & $14.84\textcolor{gray}{\text{\scriptsize±0.19}}$ & $15.02\textcolor{gray}{\text{\scriptsize±0.19}}$
& $70.08\textcolor{gray}{\text{\scriptsize±4.91}}$ & $117.62\textcolor{gray}{\text{\scriptsize±2.98}}$ & $141.30\textcolor{gray}{\text{\scriptsize±1.92}}$ \\

& MAPE (\%)
& $56.54\textcolor{gray}{\text{\scriptsize±1.34}}$ & $69.08\textcolor{gray}{\text{\scriptsize±1.30}}$ & $69.99\textcolor{gray}{\text{\scriptsize±1.47}}$
& $13.56\textcolor{gray}{\text{\scriptsize±0.77}}$ & $15.59\textcolor{gray}{\text{\scriptsize±0.75}}$ & $16.06\textcolor{gray}{\text{\scriptsize±0.73}}$
& $55.89\textcolor{gray}{\text{\scriptsize±1.69}}$ & $67.26\textcolor{gray}{\text{\scriptsize±2.25}}$ & $71.37\textcolor{gray}{\text{\scriptsize±2.27}}$
& $25.12\textcolor{gray}{\text{\scriptsize±0.89}}$ & $38.23\textcolor{gray}{\text{\scriptsize±0.99}}$ & $48.76\textcolor{gray}{\text{\scriptsize±0.73}}$ \\

& MRE (\%)
& $44.96\textcolor{gray}{\text{\scriptsize±0.32}}$ & $53.77\textcolor{gray}{\text{\scriptsize±0.24}}$ & $54.55\textcolor{gray}{\text{\scriptsize±0.33}}$
& $13.17\textcolor{gray}{\text{\scriptsize±0.34}}$ & $23.33\textcolor{gray}{\text{\scriptsize±0.45}}$ & $30.50\textcolor{gray}{\text{\scriptsize±0.26}}$
& $43.50\textcolor{gray}{\text{\scriptsize±0.61}}$ & $51.96\textcolor{gray}{\text{\scriptsize±1.02}}$ & $53.35\textcolor{gray}{\text{\scriptsize±0.89}}$
& $14.11\textcolor{gray}{\text{\scriptsize±0.61}}$ & $24.26\textcolor{gray}{\text{\scriptsize±0.64}}$ & $30.71\textcolor{gray}{\text{\scriptsize±0.51}}$ \\

\cmidrule(lr){2-14}

\multirow{4}{*}{\makecell[l]{\textbf{TimeXer}\\\citeyearpar{wang2024timexer}}}
& MAE
& $13.11\textcolor{gray}{\text{\scriptsize±0.02}}$ &
$13.92\textcolor{gray}{\text{\scriptsize±0.29}}$ &
$14.08\textcolor{gray}{\text{\scriptsize±0.18}}$ &
$5.07\textcolor{gray}{\text{\scriptsize±0.09}}$ &
$5.61\textcolor{gray}{\text{\scriptsize±0.11}}$ &
$5.71\textcolor{gray}{\text{\scriptsize±0.11}}$ &
$9.65\textcolor{gray}{\text{\scriptsize±0.59}}$ &
$10.13\textcolor{gray}{\text{\scriptsize±0.08}}$ &
$9.88\textcolor{gray}{\text{\scriptsize±0.46}}$ &
$128.05\textcolor{gray}{\text{\scriptsize±1.05}}$ &
$138.34\textcolor{gray}{\text{\scriptsize±4.13}}$ &
$149.21\textcolor{gray}{\text{\scriptsize±6.48}}$ \\

& RMSE
& $20.20\textcolor{gray}{\text{\scriptsize±0.03}}$ &
$20.88\textcolor{gray}{\text{\scriptsize±0.05}}$ &
$21.23\textcolor{gray}{\text{\scriptsize±0.03}}$ &
$38.84\textcolor{gray}{\text{\scriptsize±2.09}}$ &
$41.25\textcolor{gray}{\text{\scriptsize±2.23}}$ &
$41.84\textcolor{gray}{\text{\scriptsize±2.19}}$ &
$14.98\textcolor{gray}{\text{\scriptsize±0.46}}$ &
$15.57\textcolor{gray}{\text{\scriptsize±0.12}}$ &
$15.14\textcolor{gray}{\text{\scriptsize±0.59}}$ &
$191.72\textcolor{gray}{\text{\scriptsize±1.27}}$ &
$199.38\textcolor{gray}{\text{\scriptsize±8.40}}$ &
$210.42\textcolor{gray}{\text{\scriptsize±8.93}}$ \\

& MAPE (\%)
& $64.48\textcolor{gray}{\text{\scriptsize±0.27}}$ &
$65.90\textcolor{gray}{\text{\scriptsize±4.25}}$ &
$68.33\textcolor{gray}{\text{\scriptsize±3.68}}$ &
$12.42\textcolor{gray}{\text{\scriptsize±0.51}}$ &
$13.67\textcolor{gray}{\text{\scriptsize±0.55}}$ &
$13.93\textcolor{gray}{\text{\scriptsize±0.55}}$ &
$63.07\textcolor{gray}{\text{\scriptsize±7.20}}$ &
$66.30\textcolor{gray}{\text{\scriptsize±3.65}}$ &
$65.05\textcolor{gray}{\text{\scriptsize±2.85}}$ &
$191.61\textcolor{gray}{\text{\scriptsize±3.61}}$ &
$188.09\textcolor{gray}{\text{\scriptsize±1.44}}$ &
$190.11\textcolor{gray}{\text{\scriptsize±0.93}}$ \\

& MRE (\%)
& $50.71\textcolor{gray}{\text{\scriptsize±0.08}}$ &
$53.93\textcolor{gray}{\text{\scriptsize±1.11}}$ &
$54.49\textcolor{gray}{\text{\scriptsize±0.68}}$ &
$13.20\textcolor{gray}{\text{\scriptsize±0.24}}$ &
$14.59\textcolor{gray}{\text{\scriptsize±0.29}}$ &
$14.86\textcolor{gray}{\text{\scriptsize±0.29}}$ &
$51.64\textcolor{gray}{\text{\scriptsize±3.13}}$ &
$54.21\textcolor{gray}{\text{\scriptsize±0.40}}$ &
$52.86\textcolor{gray}{\text{\scriptsize±2.47}}$ &
$46.41\textcolor{gray}{\text{\scriptsize±0.34}}$ &
$49.96\textcolor{gray}{\text{\scriptsize±2.01}}$ &
$53.88\textcolor{gray}{\text{\scriptsize±2.67}}$ \\

\cmidrule(lr){2-14}

\multirow{4}{*}{\makecell[l]{\textbf{CrossLinear}\\\citeyearpar{zhou2025crosslinear}}}
& MAE
& $13.32\textcolor{gray}{\text{\scriptsize±1.32}}$ & $14.65\textcolor{gray}{\text{\scriptsize±1.83}}$ & $15.39\textcolor{gray}{\text{\scriptsize±0.76}}$
& $4.44\textcolor{gray}{\text{\scriptsize±0.05}}$ & $5.09\textcolor{gray}{\text{\scriptsize±0.06}}$ & $5.21\textcolor{gray}{\text{\scriptsize±0.06}}$
& $8.44\textcolor{gray}{\text{\scriptsize±0.23}}$ & $9.93\textcolor{gray}{\text{\scriptsize±0.21}}$ & $10.26\textcolor{gray}{\text{\scriptsize±0.25}}$
& $46.07\textcolor{gray}{\text{\scriptsize±13.80}}$ & $70.75\textcolor{gray}{\text{\scriptsize±9.89}}$ & $84.35\textcolor{gray}{\text{\scriptsize±0.67}}$ \\

& RMSE
& $19.29\textcolor{gray}{\text{\scriptsize±1.54}}$ & $21.26\textcolor{gray}{\text{\scriptsize±1.86}}$ & $22.11\textcolor{gray}{\text{\scriptsize±0.55}}$
& $37.77\textcolor{gray}{\text{\scriptsize±2.13}}$ & $40.69\textcolor{gray}{\text{\scriptsize±2.23}}$ & $41.53\textcolor{gray}{\text{\scriptsize±2.22}}$
& $13.20\textcolor{gray}{\text{\scriptsize±0.13}}$ & $14.99\textcolor{gray}{\text{\scriptsize±0.09}}$ & $15.27\textcolor{gray}{\text{\scriptsize±0.08}}$
& $81.98\textcolor{gray}{\text{\scriptsize±24.55}}$ & $124.28\textcolor{gray}{\text{\scriptsize±12.73}}$ & $142.91\textcolor{gray}{\text{\scriptsize±0.14}}$ \\

& MAPE (\%)
& $70.55\textcolor{gray}{\text{\scriptsize±0.09}}$ & $77.78\textcolor{gray}{\text{\scriptsize±0.14}}$ & $81.03\textcolor{gray}{\text{\scriptsize±0.08}}$
& $10.07\textcolor{gray}{\text{\scriptsize±0.00}}$ & $11.67\textcolor{gray}{\text{\scriptsize±0.00}}$ & $12.00\textcolor{gray}{\text{\scriptsize±0.00}}$
& $53.06\textcolor{gray}{\text{\scriptsize±0.04}}$ & $63.76\textcolor{gray}{\text{\scriptsize±0.03}}$ & $68.33\textcolor{gray}{\text{\scriptsize±0.03}}$
& $25.91\textcolor{gray}{\text{\scriptsize±0.07}}$ & $37.08\textcolor{gray}{\text{\scriptsize±0.06}}$ & $43.25\textcolor{gray}{\text{\scriptsize±0.01}}$ \\

& MRE (\%)
& $51.62\textcolor{gray}{\text{\scriptsize±0.05}}$ & $56.70\textcolor{gray}{\text{\scriptsize±0.07}}$ & $59.59\textcolor{gray}{\text{\scriptsize±0.03}}$
& $11.56\textcolor{gray}{\text{\scriptsize±0.00}}$ & $13.24\textcolor{gray}{\text{\scriptsize±0.00}}$ & $13.54\textcolor{gray}{\text{\scriptsize±0.00}}$
& $45.17\textcolor{gray}{\text{\scriptsize±0.01}}$ & $53.13\textcolor{gray}{\text{\scriptsize±0.01}}$ & $54.87\textcolor{gray}{\text{\scriptsize±0.01}}$
& $16.69\textcolor{gray}{\text{\scriptsize±0.05}}$ & $25.59\textcolor{gray}{\text{\scriptsize±0.04}}$ & $30.46\textcolor{gray}{\text{\scriptsize±0.01}}$ \\

\cmidrule(lr){2-14}

\multirow{4}{*}{\makecell[l]{\textbf{DAG}\\\citeyearpar{qiu2025dagdualcausalnetwork}}}
& MAE
& $11.48\textcolor{gray}{\text{\scriptsize±0.10}}$ & $13.39\textcolor{gray}{\text{\scriptsize±0.07}}$ & $13.55\textcolor{gray}{\text{\scriptsize±0.07}}$
& $17.35\textcolor{gray}{\text{\scriptsize±1.29}}$ & $18.05\textcolor{gray}{\text{\scriptsize±1.36}}$ & $17.85\textcolor{gray}{\text{\scriptsize±1.31}}$
& $8.32\textcolor{gray}{\text{\scriptsize±0.19}}$ & $9.84\textcolor{gray}{\text{\scriptsize±0.24}}$ & $9.95\textcolor{gray}{\text{\scriptsize±0.17}}$
& $57.68\textcolor{gray}{\text{\scriptsize±8.59}}$ & $82.09\textcolor{gray}{\text{\scriptsize±6.10}}$ & $89.70\textcolor{gray}{\text{\scriptsize±4.11}}$ \\

& RMSE
& $18.47\textcolor{gray}{\text{\scriptsize±0.12}}$ & $20.47\textcolor{gray}{\text{\scriptsize±0.10}}$ & $20.89\textcolor{gray}{\text{\scriptsize±0.10}}$
& $104.01\textcolor{gray}{\text{\scriptsize±9.44}}$ & $109.77\textcolor{gray}{\text{\scriptsize±10.34}}$ & $109.78\textcolor{gray}{\text{\scriptsize±10.34}}$
& $13.26\textcolor{gray}{\text{\scriptsize±0.11}}$ & $15.10\textcolor{gray}{\text{\scriptsize±0.23}}$ & $15.06\textcolor{gray}{\text{\scriptsize±0.22}}$
& $101.29\textcolor{gray}{\text{\scriptsize±10.47}}$ & $137.06\textcolor{gray}{\text{\scriptsize±6.26}}$ & $146.73\textcolor{gray}{\text{\scriptsize±4.37}}$ \\

& MAPE (\%)
& $53.40\textcolor{gray}{\text{\scriptsize±0.01}}$ & $63.18\textcolor{gray}{\text{\scriptsize±0.00}}$ & $63.59\textcolor{gray}{\text{\scriptsize±0.00}}$
& $39.64\textcolor{gray}{\text{\scriptsize±0.03}}$ & $41.46\textcolor{gray}{\text{\scriptsize±0.04}}$ & $41.11\textcolor{gray}{\text{\scriptsize±0.03}}$
& $54.93\textcolor{gray}{\text{\scriptsize±0.04}}$ & $66.86\textcolor{gray}{\text{\scriptsize±0.04}}$ & $70.26\textcolor{gray}{\text{\scriptsize±0.03}}$
& $33.59\textcolor{gray}{\text{\scriptsize±0.06}}$ & $45.07\textcolor{gray}{\text{\scriptsize±0.05}}$ & $48.86\textcolor{gray}{\text{\scriptsize±0.03}}$ \\

& MRE (\%)
& $44.38\textcolor{gray}{\text{\scriptsize±0.00}}$ & $51.88\textcolor{gray}{\text{\scriptsize±0.00}}$ & $52.43\textcolor{gray}{\text{\scriptsize±0.00}}$
& $45.16\textcolor{gray}{\text{\scriptsize±0.03}}$ & $46.97\textcolor{gray}{\text{\scriptsize±0.04}}$ & $46.43\textcolor{gray}{\text{\scriptsize±0.03}}$
& $44.53\textcolor{gray}{\text{\scriptsize±0.01}}$ & $52.65\textcolor{gray}{\text{\scriptsize±0.01}}$ & $53.25\textcolor{gray}{\text{\scriptsize±0.01}}$
& $20.91\textcolor{gray}{\text{\scriptsize±0.03}}$ & $29.71\textcolor{gray}{\text{\scriptsize±0.02}}$ & $32.39\textcolor{gray}{\text{\scriptsize±0.01}}$ \\

\midrule

\multirow{4}{*}{\makecell[l]{\textbf{MAGCRN}\\\citeyearpar{giganti2024back}}}
& MAE
& $\secondres{{9.82}}\textcolor{gray}{\text{\scriptsize±0.11}}$ & $\secondres{{10.52}}\textcolor{gray}{\text{\scriptsize±0.13}}$ & $\secondres{{11.03}}\textcolor{gray}{\text{\scriptsize±0.27}}$
& $\secondres{{3.87}}\textcolor{gray}{\text{\scriptsize±0.05}}$ & $\secondres{{3.88}}\textcolor{gray}{\text{\scriptsize±0.05}}$ & $\secondres{{3.89}}\textcolor{gray}{\text{\scriptsize±0.05}}$
& $\secondres{{7.41}}\textcolor{gray}{\text{\scriptsize±0.24}}$ & $\secondres{{7.80}}\textcolor{gray}{\text{\scriptsize±0.25}}$ & $\secondres{{8.06}}\textcolor{gray}{\text{\scriptsize±0.33}}$
& $\secondres{{37.47}}\textcolor{gray}{\text{\scriptsize±1.06}}$ & $\secondres{{52.51}}\textcolor{gray}{\text{\scriptsize±6.80}}$ & $\secondres{{65.32}}\textcolor{gray}{\text{\scriptsize±11.02}}$ \\

& RMSE
& $\secondres{{15.18}}\textcolor{gray}{\text{\scriptsize±0.25}}$ & $\secondres{{16.03}}\textcolor{gray}{\text{\scriptsize±0.24}}$ & $\secondres{{16.55}}\textcolor{gray}{\text{\scriptsize±0.38}}$
& $\secondres{{31.83}}\textcolor{gray}{\text{\scriptsize±0.02}}$ & $\secondres{{31.84}}\textcolor{gray}{\text{\scriptsize±0.03}}$ & $\secondres{{31.91}}\textcolor{gray}{\text{\scriptsize±0.02}}$
& $\secondres{{11.43}}\textcolor{gray}{\text{\scriptsize±0.14}}$ & $\secondres{{11.91}}\textcolor{gray}{\text{\scriptsize±0.23}}$ & $\secondres{{11.96}}\textcolor{gray}{\text{\scriptsize±0.40}}$
& $\secondres{{68.52}}\textcolor{gray}{\text{\scriptsize±4.56}}$ & $\secondres{{93.26}}\textcolor{gray}{\text{\scriptsize±9.74}}$ & $\firstres{{108.66}}\textcolor{gray}{\text{\scriptsize±17.87}}$ \\

& MAPE (\%)
& $\secondres{{38.66}}\textcolor{gray}{\text{\scriptsize±0.27}}$ & $\firstres{{42.53}}\textcolor{gray}{\text{\scriptsize±0.41}}$ & $\secondres{{45.08}}\textcolor{gray}{\text{\scriptsize±1.15}}$
& $\secondres{{9.41}}\textcolor{gray}{\text{\scriptsize±0.07}}$ & $\secondres{{9.43}}\textcolor{gray}{\text{\scriptsize±0.08}}$ & $\secondres{{9.44}}\textcolor{gray}{\text{\scriptsize±0.08}}$
& $\firstres{{43.79}}\textcolor{gray}{\text{\scriptsize±4.21}}$ & $\firstres{{45.89}}\textcolor{gray}{\text{\scriptsize±4.35}}$ & $\firstres{{49.74}}\textcolor{gray}{\text{\scriptsize±4.19}}$
& $\secondres{{20.12}}\textcolor{gray}{\text{\scriptsize±0.26}}$ & $\secondres{{27.20}}\textcolor{gray}{\text{\scriptsize±2.58}}$ & $\secondres{{34.88}}\textcolor{gray}{\text{\scriptsize±4.66}}$ \\

& MRE (\%)
& $\secondres{{37.97}}\textcolor{gray}{\text{\scriptsize±0.44}}$ & $\secondres{{40.69}}\textcolor{gray}{\text{\scriptsize±0.50}}$ & $\secondres{{42.72}}\textcolor{gray}{\text{\scriptsize±1.03}}$
& $\secondres{{10.08}}\textcolor{gray}{\text{\scriptsize±0.13}}$ & $\secondres{{10.08}}\textcolor{gray}{\text{\scriptsize±0.13}}$ & $\secondres{{10.09}}\textcolor{gray}{\text{\scriptsize±0.13}}$
& $\secondres{{39.66}}\textcolor{gray}{\text{\scriptsize±1.30}}$ & $\secondres{{41.74}}\textcolor{gray}{\text{\scriptsize±1.34}}$ & $\secondres{{43.15}}\textcolor{gray}{\text{\scriptsize±1.77}}$
& $\secondres{{13.58}}\textcolor{gray}{\text{\scriptsize±0.38}}$ & $\secondres{{19.00}}\textcolor{gray}{\text{\scriptsize±2.46}}$ & $\secondres{{23.59}}\textcolor{gray}{\text{\scriptsize±3.98}}$
\\

\midrule

\multirow{4}{*}{\makecell[l]{\textbf{ChronosX}\\\citeyearpar{Arango2025ChronosX}}}
& MAE
& $12.46\textcolor{gray}{\text{\scriptsize±0.28}}$ & $13.02\textcolor{gray}{\text{\scriptsize±0.12}}$ & $13.30\textcolor{gray}{\text{\scriptsize±0.03}}$
& $39.20\textcolor{gray}{\text{\scriptsize±0.57}}$ & $39.19\textcolor{gray}{\text{\scriptsize±0.56}}$ & $39.20\textcolor{gray}{\text{\scriptsize±0.55}}$
& $8.87\textcolor{gray}{\text{\scriptsize±0.06}}$ & $9.30\textcolor{gray}{\text{\scriptsize±0.13}}$ & $9.54\textcolor{gray}{\text{\scriptsize±0.02}}$
& $142.06\textcolor{gray}{\text{\scriptsize±3.19}}$ & $148.59\textcolor{gray}{\text{\scriptsize±3.20}}$ & $156.41\textcolor{gray}{\text{\scriptsize±4.19}}$ \\

& RMSE
& $18.37\textcolor{gray}{\text{\scriptsize±0.36}}$ & $19.17\textcolor{gray}{\text{\scriptsize±0.11}}$ & $19.41\textcolor{gray}{\text{\scriptsize±0.13}}$
& $139.33\textcolor{gray}{\text{\scriptsize±0.10}}$ & $139.36\textcolor{gray}{\text{\scriptsize±0.09}}$ & $139.43\textcolor{gray}{\text{\scriptsize±0.08}}$
& $13.23\textcolor{gray}{\text{\scriptsize±0.11}}$ & $13.87\textcolor{gray}{\text{\scriptsize±0.12}}$ & $13.92\textcolor{gray}{\text{\scriptsize±0.14}}$
& $216.75\textcolor{gray}{\text{\scriptsize±3.33}}$ & $222.80\textcolor{gray}{\text{\scriptsize±4.13}}$ & $231.04\textcolor{gray}{\text{\scriptsize±4.85}}$ \\

& MAPE (\%)
& $55.86\textcolor{gray}{\text{\scriptsize±0.72}}$ & $59.48\textcolor{gray}{\text{\scriptsize±1.72}}$ & $60.74\textcolor{gray}{\text{\scriptsize±2.43}}$
& $99.50\textcolor{gray}{\text{\scriptsize±0.44}}$ & $99.54\textcolor{gray}{\text{\scriptsize±0.44}}$ & $99.58\textcolor{gray}{\text{\scriptsize±0.42}}$
& $58.84\textcolor{gray}{\text{\scriptsize±2.39}}$ & $61.46\textcolor{gray}{\text{\scriptsize±2.74}}$ & $65.24\textcolor{gray}{\text{\scriptsize±2.63}}$
& $58.60\textcolor{gray}{\text{\scriptsize±3.79}}$ & $64.23\textcolor{gray}{\text{\scriptsize±1.65}}$ & $69.75\textcolor{gray}{\text{\scriptsize±1.09}}$ \\

& MRE (\%)
& $48.17\textcolor{gray}{\text{\scriptsize±1.08}}$ & $50.39\textcolor{gray}{\text{\scriptsize±0.49}}$ & $51.53\textcolor{gray}{\text{\scriptsize±0.13}}$
& $102.00\textcolor{gray}{\text{\scriptsize±1.52}}$ & $101.98\textcolor{gray}{\text{\scriptsize±1.46}}$ & $101.98\textcolor{gray}{\text{\scriptsize±1.41}}$
& $47.49\textcolor{gray}{\text{\scriptsize±0.34}}$ & $49.78\textcolor{gray}{\text{\scriptsize±0.71}}$ & $51.07\textcolor{gray}{\text{\scriptsize±0.11}}$
& $51.35\textcolor{gray}{\text{\scriptsize±0.92}}$ & $53.78\textcolor{gray}{\text{\scriptsize±1.16}}$ & $56.49\textcolor{gray}{\text{\scriptsize±1.52}}$ \\

\midrule
\multirow{4}{*}{\textbf{\model}}

& \cellcolor{gray!8}MAE
& \cellcolor{gray!8}$\firstres{{9.33}}\textcolor{gray}{\text{\scriptsize±0.13}}$ & 
  \cellcolor{gray!8}$\firstres{{10.04}}\textcolor{gray}{\text{\scriptsize±0.19}}$ & 
  \cellcolor{gray!8}$\firstres{{10.41}}\textcolor{gray}{\text{\scriptsize±0.12}}$ &
  \cellcolor{gray!8}$\firstres{{3.79}}\textcolor{gray}{\text{\scriptsize±0.01}}$ &
  \cellcolor{gray!8}$\firstres{{3.80}}\textcolor{gray}{\text{\scriptsize±0.01}}$ &
  \cellcolor{gray!8}$\firstres{{3.80}}\textcolor{gray}{\text{\scriptsize±0.01}}$ &
  \cellcolor{gray!8}$\firstres{{7.33}}\textcolor{gray}{\text{\scriptsize±0.21}}$ & 
  \cellcolor{gray!8}$\firstres{{7.61}}\textcolor{gray}{\text{\scriptsize±0.23}}$ &
  \cellcolor{gray!8}$\firstres{{7.94}}\textcolor{gray}{\text{\scriptsize±0.29}}$ & 
  \cellcolor{gray!8}$\firstres{{31.91}}\textcolor{gray}{\text{\scriptsize±1.35}}$ & 
  \cellcolor{gray!8}$\firstres{{50.59}}\textcolor{gray}{\text{\scriptsize±0.04}}$& 
  \cellcolor{gray!8}$\firstres{{64.13}}\textcolor{gray}{\text{\scriptsize±3.86}}$ \\

& \cellcolor{gray!8}RMSE
& \cellcolor{gray!8}$\firstres{{14.24}}\textcolor{gray}{\text{\scriptsize±0.18}}$ & 
  \cellcolor{gray!8}$\firstres{{15.12}}\textcolor{gray}{\text{\scriptsize±0.26}}$ & 
  \cellcolor{gray!8}$\firstres{{15.51}}\textcolor{gray}{\text{\scriptsize±0.16}}$ & 
  \cellcolor{gray!8}$\firstres{{31.73}}\textcolor{gray}{\text{\scriptsize±0.02}}$ &
  \cellcolor{gray!8}$\firstres{{31.78}}\textcolor{gray}{\text{\scriptsize±0.02}}$ &
  \cellcolor{gray!8}$\firstres{{31.85}}\textcolor{gray}{\text{\scriptsize±0.02}}$ &
  \cellcolor{gray!8}$\firstres{{11.14}}\textcolor{gray}{\text{\scriptsize±0.21}}$ & 
  \cellcolor{gray!8}$\firstres{{11.52}}\textcolor{gray}{\text{\scriptsize±0.49}}$ & 
  \cellcolor{gray!8}$\firstres{{11.79}}\textcolor{gray}{\text{\scriptsize±0.51}}$ &
  \cellcolor{gray!8}$\firstres{{56.49}}\textcolor{gray}{\text{\scriptsize±2.54}}$ & 
  \cellcolor{gray!8}$\firstres{{89.10}}\textcolor{gray}{\text{\scriptsize±7.36}}$ & 
  \cellcolor{gray!8}$\secondres{{109.01}}\textcolor{gray}{\text{\scriptsize±9.19}}$ \\

& \cellcolor{gray!8}MAPE (\%)
& \cellcolor{gray!8}$\firstres{{38.57}}\textcolor{gray}{\text{\scriptsize±0.32}}$ & 
  \cellcolor{gray!8}$\secondres{{42.62}}\textcolor{gray}{\text{\scriptsize±0.52}}$ & 
  \cellcolor{gray!8}$\firstres{{44.65}}\textcolor{gray}{\text{\scriptsize±0.08}}$ & 
  \cellcolor{gray!8}$\firstres{{9.30}}\textcolor{gray}{\text{\scriptsize±0.03}}$ &
  \cellcolor{gray!8}$\firstres{{9.32}}\textcolor{gray}{\text{\scriptsize±0.03}}$ &
  \cellcolor{gray!8}$\firstres{{9.31}}\textcolor{gray}{\text{\scriptsize±0.03}}$ &
  \cellcolor{gray!8}$\secondres{{44.90}}\textcolor{gray}{\text{\scriptsize±0.88}}$ & 
  \cellcolor{gray!8}$\secondres{{46.59}}\textcolor{gray}{\text{\scriptsize±0.64}}$ & 
  \cellcolor{gray!8}$\secondres{{50.27}}\textcolor{gray}{\text{\scriptsize±1.18}}$ & 
  \cellcolor{gray!8}$\firstres{{17.63}}\textcolor{gray}{\text{\scriptsize±0.97}}$ & 
  \cellcolor{gray!8}$\firstres{{26.68}}\textcolor{gray}{\text{\scriptsize±1.69}}$ & 
  \cellcolor{gray!8}$\firstres{{34.52}}\textcolor{gray}{\text{\scriptsize±2.90}}$ \\

& \cellcolor{gray!8}MRE (\%)
& \cellcolor{gray!8}$\firstres{{36.09}}\textcolor{gray}{\text{\scriptsize±0.49}}$ & 
  \cellcolor{gray!8}$\firstres{{38.84}}\textcolor{gray}{\text{\scriptsize±0.75}}$ & 
  \cellcolor{gray!8}$\firstres{{40.34}}\textcolor{gray}{\text{\scriptsize±0.47}}$ &
  \cellcolor{gray!8}$\firstres{{9.88}}\textcolor{gray}{\text{\scriptsize±0.03}}$ &
  \cellcolor{gray!8}$\firstres{{9.89}}\textcolor{gray}{\text{\scriptsize±0.03}}$ &
  \cellcolor{gray!8}$\firstres{{9.90}}\textcolor{gray}{\text{\scriptsize±0.03}}$ &
  \cellcolor{gray!8}$\firstres{{39.23}}\textcolor{gray}{\text{\scriptsize±1.11}}$ & 
  \cellcolor{gray!8}$\firstres{{40.74}}\textcolor{gray}{\text{\scriptsize±1.24}}$ & 
  \cellcolor{gray!8}$\firstres{{42.50}}\textcolor{gray}{\text{\scriptsize±1.57}}$ &
  \cellcolor{gray!8}$\firstres{{11.57}}\textcolor{gray}{\text{\scriptsize±0.49}}$ & 
  \cellcolor{gray!8}$\firstres{{18.31}}\textcolor{gray}{\text{\scriptsize±1.14}}$ & 
  \cellcolor{gray!8}$\firstres{{23.16}}\textcolor{gray}{\text{\scriptsize±1.40}}$ \\

\bottomrule

\end{tabular}
\end{sc}
}
\vspace{-4mm}
\end{table*}

\subsection{Effectiveness Study (RQ2)}

Table \ref{tab:rq1_overall_performance} presents a performance comparison of our \model against state-of-the-art time-series, spatio-temporal, and foundation models with exogenous variable modeling techniques. To ensure a fair comparison, we employ the standard AGCRN~\citep{Bai2020Adaptive} as the backbone for all methods utilizing a decoupled design. Our observations are as follows: \ding{182} Pure time-series methods (first 5) consistently yield the lowest performance, aligning with established ST benchmarks~\citep{shao2024exploring}. This deficiency stems from their inability to model spatial dependencies, resulting in suboptimal accuracy. \ding{183} While ChronosX achieves relative gains by integrating the spatio-temporal backbone, it frequently suffers from non-convergence on specific tasks (\eg, \SpeedNineteen~and \TrafficTwenty), despite extensive hyperparameter tuning. We attribute this instability to the inherent design of the underlying Chronos-type foundation models~\citep{ansarichronos}. \ding{184} MAGCRN often produces suboptimal results as it is tailored for specific scenarios yet neglects the inconsistent and unbalanced effects of exogenous variables. \ding{185} Conversely, our \model achieves superior performance across all metrics on the four tasks. By effectively selecting and balancing exogenous information, \model adaptively handles diverse input sources to maintain robust performance.

\subsection{Robustness Study (RQ3)}
In real-world scenarios, spatio-temporal systems frequently encounter unreliable exogenous data collection (\eg, signal loss from sensor anomalies), which compromises prediction accuracy and robustness. To assess the \model under such conditions, we designed a systematic experiment simulating data corruption. Specifically, we randomly replaced exogenous variables at varying degradation levels (20\%, 40\%, 60\%, and 80\%) with zeros or noise from a standard normal distribution $\mathcal{N}(0,1)$. The 3-day prediction results in Table~\ref{tab:mask_3day} reveal that: \ding{182} While performance gradually declines as the missing data ratio increases, the absolute drop remains minimal, indicating our framework maintains reasonable efficacy even with limited auxiliary information. \ding{183} Surprisingly, for certain tasks, appropriate levels of missing signals improve performance. This suggests that random corruption acts as a data augmentation strategy, forcing the model to predict under incomplete or noisy conditions, thereby enhancing generalization and robustness. Similar trends for 1-day and 2-day scenarios are reported in Appendix~\ref{appendix_robustness}.

\begin{table*}[t!]
\caption{Performance under abnormal conditions with missing exogenous variable signals (3-day scenario).
}
\vspace{-2mm}
\label{tab:mask_3day}
\renewcommand{\arraystretch}{1.2}
\resizebox{\linewidth}{!}{
\begin{sc}
\begin{tabular}{lccccccccccccc}
\toprule
\multicolumn{2}{c}{\textbf{Datasets}}
  & \multicolumn{4}{c}{\textbf{AQI-19}}
  & \multicolumn{4}{c}{\textbf{Speed-19}$^*$}
  & \multicolumn{4}{c}{\textbf{Intensity-22}} \\
\cmidrule(lr){1-2}\cmidrule(lr){3-6}\cmidrule(lr){7-10}\cmidrule(lr){11-14}

\multicolumn{2}{c}{\textbf{Masking Strategy}}
  & MAE & RMSE & MAPE(\%) & MRE(\%)
  & MAE & RMSE & MAPE(\%) & MRE(\%)
  & MAE & RMSE & MAPE(\%) & MRE(\%) \\
\midrule

\multirow{4}{*}{Zero}
  & 20\% &
  $\secondres{{10.60}}\textcolor{gray}{\text{\scriptsize±0.15}}$ & $\secondres{{15.87}}\textcolor{gray}{\text{\scriptsize±0.16}}$ & $44.83\textcolor{gray}{\text{\scriptsize±1.39}}$ & $\secondres{{41.07}}\textcolor{gray}{\text{\scriptsize±0.60}}$ &
  $3.87\textcolor{gray}{\text{\scriptsize±0.03}}$ & $31.94\textcolor{gray}{\text{\scriptsize±0.06}}$ & $9.47\textcolor{gray}{\text{\scriptsize±1.20}}$ & $10.08\textcolor{gray}{\text{\scriptsize±0.90}}$ & $64.31\textcolor{gray}{\text{\scriptsize±9.49}}$ & $\firstres{{108.23}}\textcolor{gray}{\text{\scriptsize±18.00}}$ & $\firstres{{34.19}}\textcolor{gray}{\text{\scriptsize±3.45}}$ & $\secondres{{23.23}}\textcolor{gray}{\text{\scriptsize±3.43}}$ \\
  & 40\% &
  $11.33\textcolor{gray}{\text{\scriptsize±0.02}}$ & $17.07\textcolor{gray}{\text{\scriptsize±0.09}}$ & $46.13\textcolor{gray}{\text{\scriptsize±1.45}}$ & $43.91\textcolor{gray}{\text{\scriptsize±0.09}}$ &
  $3.89\textcolor{gray}{\text{\scriptsize±0.07}}$ & $31.98\textcolor{gray}{\text{\scriptsize±0.08}}$ & $9.53\textcolor{gray}{\text{\scriptsize±1.80}}$ & $10.12\textcolor{gray}{\text{\scriptsize±1.60}}$ &
  $67.80\textcolor{gray}{\text{\scriptsize±1.82}}$ & $115.25\textcolor{gray}{\text{\scriptsize±3.73}}$ & $38.21\textcolor{gray}{\text{\scriptsize±0.21}}$ & $24.49\textcolor{gray}{\text{\scriptsize±0.65}}$ \\
  & 60\% &
  $11.44\textcolor{gray}{\text{\scriptsize±0.52}}$ & $17.30\textcolor{gray}{\text{\scriptsize±0.84}}$ & $48.30\textcolor{gray}{\text{\scriptsize±2.02}}$ & $44.34\textcolor{gray}{\text{\scriptsize±2.01}}$ &
  $3.87\textcolor{gray}{\text{\scriptsize±0.05}}$ & $31.96\textcolor{gray}{\text{\scriptsize±0.09}}$ & $9.48\textcolor{gray}{\text{\scriptsize±1.50}}$ & $10.08\textcolor{gray}{\text{\scriptsize±1.40}}$ & $68.50\textcolor{gray}{\text{\scriptsize±10.78}}$ & $118.41\textcolor{gray}{\text{\scriptsize±17.47}}$ & $36.37\textcolor{gray}{\text{\scriptsize±6.38}}$ & $24.74\textcolor{gray}{\text{\scriptsize±3.89}}$ \\
  & 80\% &
  $12.33\textcolor{gray}{\text{\scriptsize±0.00}}$ & $18.94\textcolor{gray}{\text{\scriptsize±0.00}}$ & $52.03\textcolor{gray}{\text{\scriptsize±0.00}}$ & $47.79\textcolor{gray}{\text{\scriptsize±0.00}}$ &
  $\secondres{{3.83}}\textcolor{gray}{\text{\scriptsize±0.02}}$ & $\secondres{{31.89}}\textcolor{gray}{\text{\scriptsize±0.06}}$ & $\secondres{{9.40}}\textcolor{gray}{\text{\scriptsize±0.90}}$ & $\secondres{{9.95}}\textcolor{gray}{\text{\scriptsize±0.50}}$ & $69.70\textcolor{gray}{\text{\scriptsize±10.99}}$ & $118.51\textcolor{gray}{\text{\scriptsize±18.88}}$ & $38.18\textcolor{gray}{\text{\scriptsize±6.62}}$ & $25.17\textcolor{gray}{\text{\scriptsize±3.97}}$ \\

\cmidrule(lr){2-14}

\multirow{4}{*}{Random}
  & 20\% &
  $10.67\textcolor{gray}{\text{\scriptsize±0.16}}$ & $16.17\textcolor{gray}{\text{\scriptsize±0.19}}$ & $\secondres{{44.23}}\textcolor{gray}{\text{\scriptsize±2.37}}$ & $41.32\textcolor{gray}{\text{\scriptsize±0.62}}$
  & $3.86\textcolor{gray}{\text{\scriptsize±0.03}}$ & $31.94\textcolor{gray}{\text{\scriptsize±0.08}}$ & $9.44\textcolor{gray}{\text{\scriptsize±1.40}}$ & $10.05\textcolor{gray}{\text{\scriptsize±1.00}}$ & $67.32\textcolor{gray}{\text{\scriptsize±8.42}}$ & $114.89\textcolor{gray}{\text{\scriptsize±16.03}}$ & $34.88\textcolor{gray}{\text{\scriptsize±3.82}}$ & $24.32\textcolor{gray}{\text{\scriptsize±3.04}}$ \\
  & 40\% &
  $11.30\textcolor{gray}{\text{\scriptsize±0.08}}$ & $17.09\textcolor{gray}{\text{\scriptsize±0.23}}$ & $46.57\textcolor{gray}{\text{\scriptsize±1.19}}$ & $43.78\textcolor{gray}{\text{\scriptsize±0.32}}$ & $3.84\textcolor{gray}{\text{\scriptsize±0.02}}$ & $31.92\textcolor{gray}{\text{\scriptsize±0.04}}$ & $9.42\textcolor{gray}{\text{\scriptsize±0.60}}$ & $9.99\textcolor{gray}{\text{\scriptsize±0.50}}$ &
  $\secondres{{64.43}}\textcolor{gray}{\text{\scriptsize±10.78}}$ & $110.41\textcolor{gray}{\text{\scriptsize±17.43}}$ & $\secondres{{34.34}}\textcolor{gray}{\text{\scriptsize±7.21}}$ & $23.27\textcolor{gray}{\text{\scriptsize±3.89}}$ \\
  & 60\% &
  $11.37\textcolor{gray}{\text{\scriptsize±0.05}}$ & $17.26\textcolor{gray}{\text{\scriptsize±0.22}}$ & $47.43\textcolor{gray}{\text{\scriptsize±1.15}}$ & $44.04\textcolor{gray}{\text{\scriptsize±0.18}}$ & $3.87\textcolor{gray}{\text{\scriptsize±0.05}}$ & $31.95\textcolor{gray}{\text{\scriptsize±0.09}}$ & $9.49\textcolor{gray}{\text{\scriptsize±1.40}}$ & $10.06\textcolor{gray}{\text{\scriptsize±1.40}}$ & $65.27\textcolor{gray}{\text{\scriptsize±6.06}}$ & $109.81\textcolor{gray}{\text{\scriptsize±11.90}}$ & $35.34\textcolor{gray}{\text{\scriptsize±3.27}}$ & $23.57\textcolor{gray}{\text{\scriptsize±2.19}}$ \\
  & 80\% &
  $12.33\textcolor{gray}{\text{\scriptsize±0.00}}$ & $18.94\textcolor{gray}{\text{\scriptsize±0.00}}$ & $52.03\textcolor{gray}{\text{\scriptsize±0.00}}$ & $47.79\textcolor{gray}{\text{\scriptsize±0.00}}$ & $3.88\textcolor{gray}{\text{\scriptsize±0.05}}$ & $31.93\textcolor{gray}{\text{\scriptsize±0.09}}$ & $9.48\textcolor{gray}{\text{\scriptsize±1.60}}$ & $10.08\textcolor{gray}{\text{\scriptsize±1.20}}$ & $68.13\textcolor{gray}{\text{\scriptsize±10.61}}$ & $115.96\textcolor{gray}{\text{\scriptsize±18.89}}$ & $36.61\textcolor{gray}{\text{\scriptsize±4.61}}$ & $24.60\textcolor{gray}{\text{\scriptsize±3.83}}$ \\
\midrule
\rowcolor{gray!8}
\multicolumn{2}{c}{No Masking}
  & $\firstres{{10.41}}\textcolor{gray}{\text{\scriptsize±0.12}}$ & $\firstres{{15.51}}\textcolor{gray}{\text{\scriptsize±0.16}}$ & $\firstres{{42.62}}\textcolor{gray}{\text{\scriptsize±0.52}}$ & $\firstres{{38.84}}\textcolor{gray}{\text{\scriptsize±0.75}}$
  & $\firstres{{3.80}}\textcolor{gray}{\text{\scriptsize±0.01}}$ & $\firstres{{31.85}}\textcolor{gray}{\text{\scriptsize±0.02}}$ & $\firstres{{9.31}}\textcolor{gray}{\text{\scriptsize±0.03}}$ & $\firstres{{9.90}}\textcolor{gray}{\text{\scriptsize±0.03}}$
  & $\firstres{{64.13}}\textcolor{gray}{\text{\scriptsize±3.86}}$ & $\secondres{{109.01}}\textcolor{gray}{\text{\scriptsize±9.19}}$ & $34.52\textcolor{gray}{\text{\scriptsize±2.90}}$ & $\firstres{{23.16}}\textcolor{gray}{\text{\scriptsize±1.40}}$\\
\bottomrule
\end{tabular}
\end{sc}
}
\vspace{-4mm}
\end{table*}

\subsection{Mechanism \& Rationality Study (RQ4)}\label{mechanism_study}

\textbf{Ablation Study.}
We conducted an ablation study to systematically evaluate the impact of different components on \model from both data and model perspectives.



\begin{figure}[t!]
    \centering
    \begin{minipage}[c]{0.49\textwidth}
        \centering
        \makeatletter\def\@captype{table}\makeatother 
        \caption{Ablation study from the data perspective on \AirNineteen~for 1-day forecasting.}
        \label{tab:performance_1day}
\small
\renewcommand{\arraystretch}{1.2}
\resizebox{1\linewidth}{!}{
    \begin{sc}
    \begin{tabular}{ccc|cccc}
        \toprule
        \textbf{P} & \textbf{F} & \textbf{D} & \textbf{MAE} & \textbf{RMSE} & \textbf{MAPE (\%)} & \textbf{MRE (\%)} \\
        \midrule
        \gcmark & - & - & $13.09\textcolor{gray}{\text{\scriptsize±0.65}}$ & $19.35\textcolor{gray}{\text{\scriptsize±0.52}}$ & $65.54\textcolor{gray}{\text{\scriptsize±5.87}}$ & $50.64\textcolor{gray}{\text{\scriptsize±2.50}}$ \\
        - & \gcmark & - & $9.97\textcolor{gray}{\text{\scriptsize±0.36}}$ & $15.09\textcolor{gray}{\text{\scriptsize±0.52}}$ & $41.20\textcolor{gray}{\text{\scriptsize±1.97}}$ & $38.56\textcolor{gray}{\text{\scriptsize±1.38}}$ \\
        - & - & \gcmark & $12.47\textcolor{gray}{\text{\scriptsize±0.15}}$ & $19.15\textcolor{gray}{\text{\scriptsize±0.20}}$ & $57.42\textcolor{gray}{\text{\scriptsize±5.11}}$ & $48.21\textcolor{gray}{\text{\scriptsize±0.58}}$ \\
        \gcmark & - & \gcmark & $12.20\textcolor{gray}{\text{\scriptsize±0.09}}$ & $18.72\textcolor{gray}{\text{\scriptsize±0.08}}$ & $59.06\textcolor{gray}{\text{\scriptsize±1.72}}$ & $47.16\textcolor{gray}{\text{\scriptsize±0.33}}$ \\
        - & \gcmark & \gcmark & $\secondres{{9.63}}\textcolor{gray}{\text{\scriptsize±0.00}}$ & $\secondres{{14.69}}\textcolor{gray}{\text{\scriptsize±0.00}}$ & $39.68\textcolor{gray}{\text{\scriptsize±0.03}}$ & $\secondres{{37.24}}\textcolor{gray}{\text{\scriptsize±0.02}}$ \\
        \gcmark & \gcmark & - & $9.69\textcolor{gray}{\text{\scriptsize±0.33}}$ & $14.93\textcolor{gray}{\text{\scriptsize±0.59}}$ & $\secondres{{39.42}}\textcolor{gray}{\text{\scriptsize±0.69}}$ & $37.46\textcolor{gray}{\text{\scriptsize±1.28}}$ \\
        \midrule
        \rowcolor{gray!8}
        \gcmark & \gcmark & \gcmark & $\firstres{{9.33}}\textcolor{gray}{\text{\scriptsize±0.13}}$ & $\firstres{{14.24}}\textcolor{gray}{\text{\scriptsize±0.18}}$ & $\firstres{{38.57}}\textcolor{gray}{\text{\scriptsize±0.32}}$ & $\firstres{{36.09}}\textcolor{gray}{\text{\scriptsize±0.49}}$ \\
        \bottomrule
    \end{tabular}
    \end{sc}
}
    \end{minipage}
    \hfill
    \begin{minipage}[c]{0.49\textwidth}
        \centering
        \includegraphics[width=\linewidth, height=4.2cm, keepaspectratio]{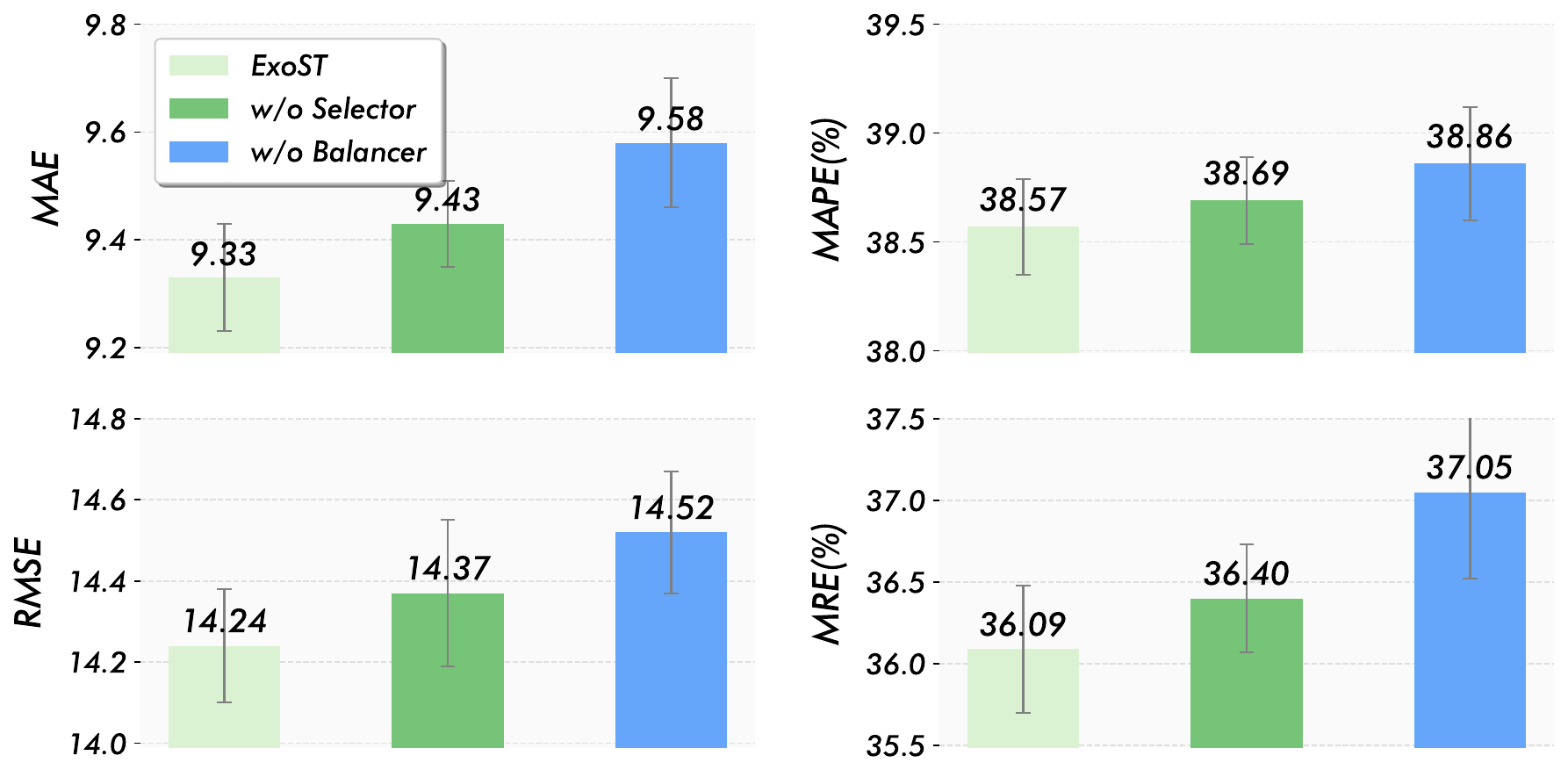}
        \vspace{-8mm}
        \caption{Ablation study on \AirNineteen~ comparing \model, \textit{w/o Selector}, and \textit{w/o Balancer} variants in 1-day horizons.}
        \label{fig:ablation_study_model_1day}
    \end{minipage}
    \vspace{-4mm}
\end{figure}

\textit{From the data perspective}, we investigated the auxiliary role of distinct exogenous variable types. Given the universality of date information, we treat it as a separate category. As shown in Table~\ref{tab:performance_1day}: \ding{182} Future exogenous variables outperform past variables, indicating that forward-looking information provides more direct predictive signals; however, combining both yields superior performance, suggesting they offer complementary patterns. \ding{183} Date exogenous variables perform poorly in isolation but drive improvements when combined with other types. \ding{184} The best performance is achieved by integrating all exogenous variables, demonstrating that \model effectively extracts and balances contributions from diverse temporal contexts. 

\textit{From the model perspective}, we examined the contribution of specific modules by constructing variants w/o the Gated Expert Selector or the Context-aware Balancer (replaced by direct element-wise fusion). As Figure~\ref{fig:ablation_study_model_1day} illustrates: \ding{182} The \textit{w/o Selector} variant suffers significant degradation across all metrics, proving the importance of addressing inconsistent variable effects. Without semantic alignment and adaptive selection, the model struggles to disentangle heterogeneous exogenous variables from endogenous ones, leading to feature space competition. \ding{183} The \textit{w/o Balancer} variant exhibits even more pronounced degradation, highlighting that temporal distribution shifts between past and future exogenous information pose a fundamental challenge. Sophisticated balancing mechanisms are key to manage noise-contaminated data and uncertainty-laden future forecasts. More experimental results are shown in Appendix~\ref{appendix_ablation}.

\begin{figure*}[t!]
    \centering
    \begin{minipage}[c]{0.49\columnwidth}
        \centering
        \includegraphics[width=\linewidth, height=4.2cm, keepaspectratio]{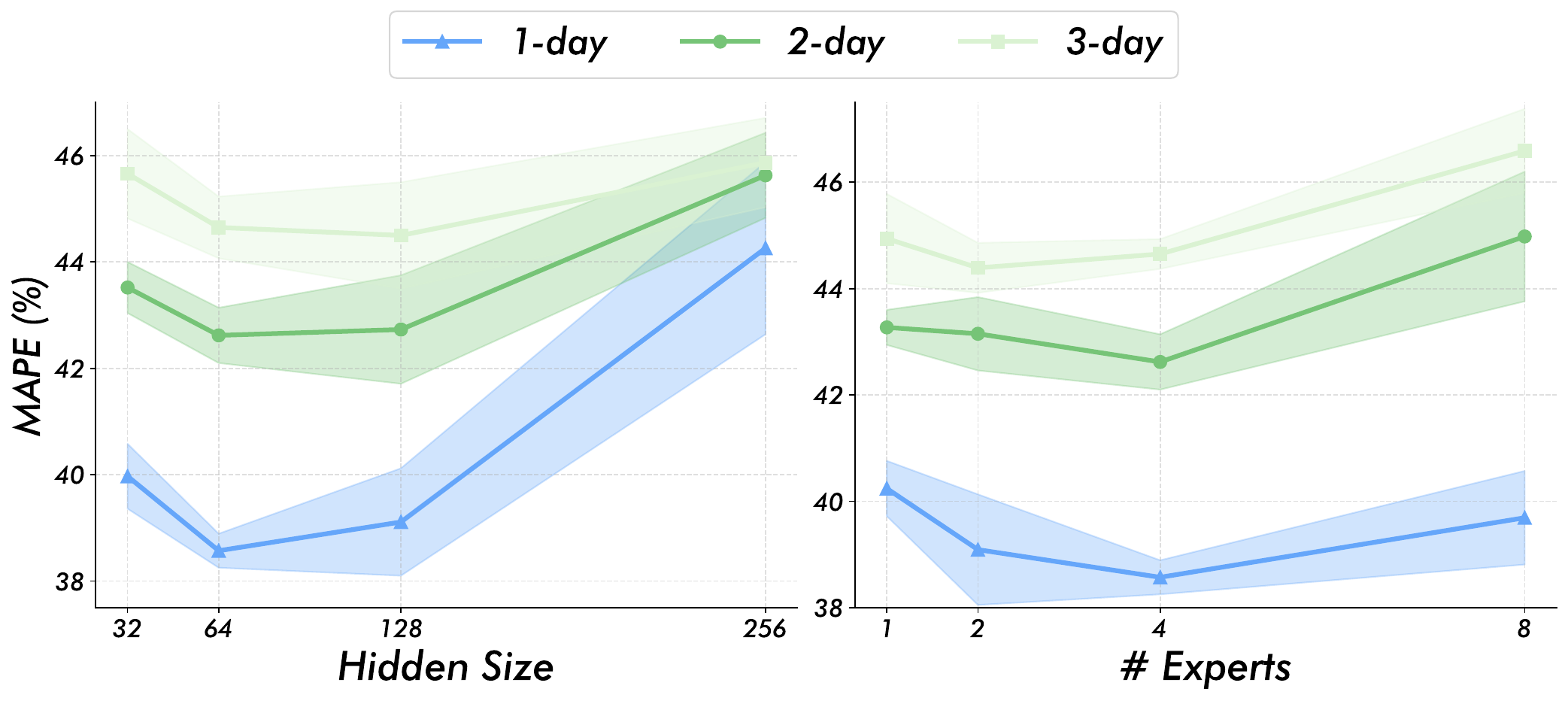}
        \caption{Different hyper-parameter performance comparison on \AirNineteen~ dataset: \textit{hidden size} of conditional representations and \textit{the number of experts} of gated expert selector across 1, 2, and 3-day forecasting horizons.}
        \label{fig:hyperparameter_study}
    \end{minipage}
    \hfill
    \begin{minipage}[c]{0.49\columnwidth}
        \centering
        \includegraphics[width=\linewidth, height=4.2cm, keepaspectratio]{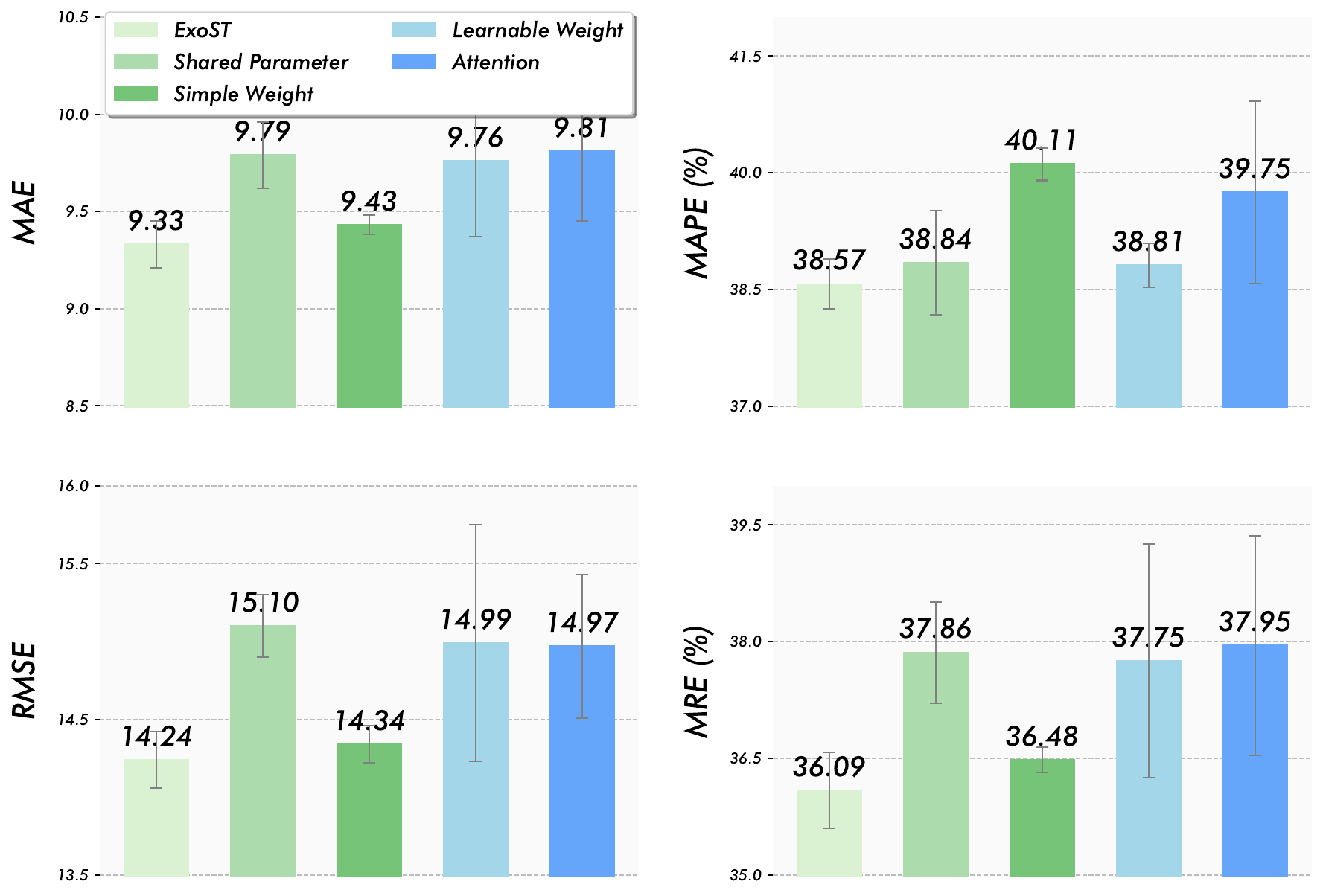}
        \caption{Different strategy performance comparison on \AirNineteen~ comparing \model, share parameter, simple weight, learned parameter and attention variants on 1-day scenario.}
        \label{fig:strategy_comparison}
    \end{minipage}
    \vspace{-4mm}
\end{figure*}

\textbf{Parameter Study.} 
We then performed a parameter sensitivity analysis to assess the impact of hyper-parameters on \model. As shown in Figure~\ref{fig:hyperparameter_study}: \ding{182} A hidden size of \textit{64} achieves optimal performance across all prediction horizons, with larger sizes causing degradation due to overfitting. \ding{183} Setting the number of experts to 4 provides the best trade-off between model capacity and computational efficiency. Fewer experts fail to capture the diverse semantic patterns required to disentangle inconsistent effects, while excessive experts lead to redundancy and overselection, diminishing the effectiveness of the mixture-of-experts gating mechanism. These findings validate the lightweight design choices of our latent-space gated expert module.


\textbf{Strategy Study.}
We compared our approach against several alternative fusion strategies to validate our design rationale:(\romannumeral1) \textit{Shared Parameter}: Siamese branches sharing parameters to learn a unified representation.(\romannumeral2) \textit{Simple Weight}: A static balancing weight ($\alpha=0.5$) fusing outputs.(\romannumeral3) \textit{Learnable Weight}: Dynamically adjusting importance via learnable parameters.(\romannumeral4) \textit{Attention}: Using cross-attention to balance contributions.As shown in Figure~\ref{fig:strategy_comparison}: \ding{182} \model outperforms all alternative strategies, confirming the effectiveness of our design.\ding{183} \textit{Simple Weight} degrades performance significantly due to its inability to adapt to varying input contexts.\ding{184} \textit{Attention} and \textit{Learnable Weight} strategies underperform while increasing computational cost; attention mechanisms introduce excessive parameter overhead, and learnable weights lack the sophisticated context modeling required to handle the nuanced differences between past and future exogenous variables. More experimental results are shown in Appendix~\ref{appendix_strategy}.


\subsection{Efficiency \& Lightweight Study (RQ5)}

\begin{wrapfigure}{r}{6cm}
\begin{center}
\vspace{-14mm}
\includegraphics[width=1.0\linewidth]{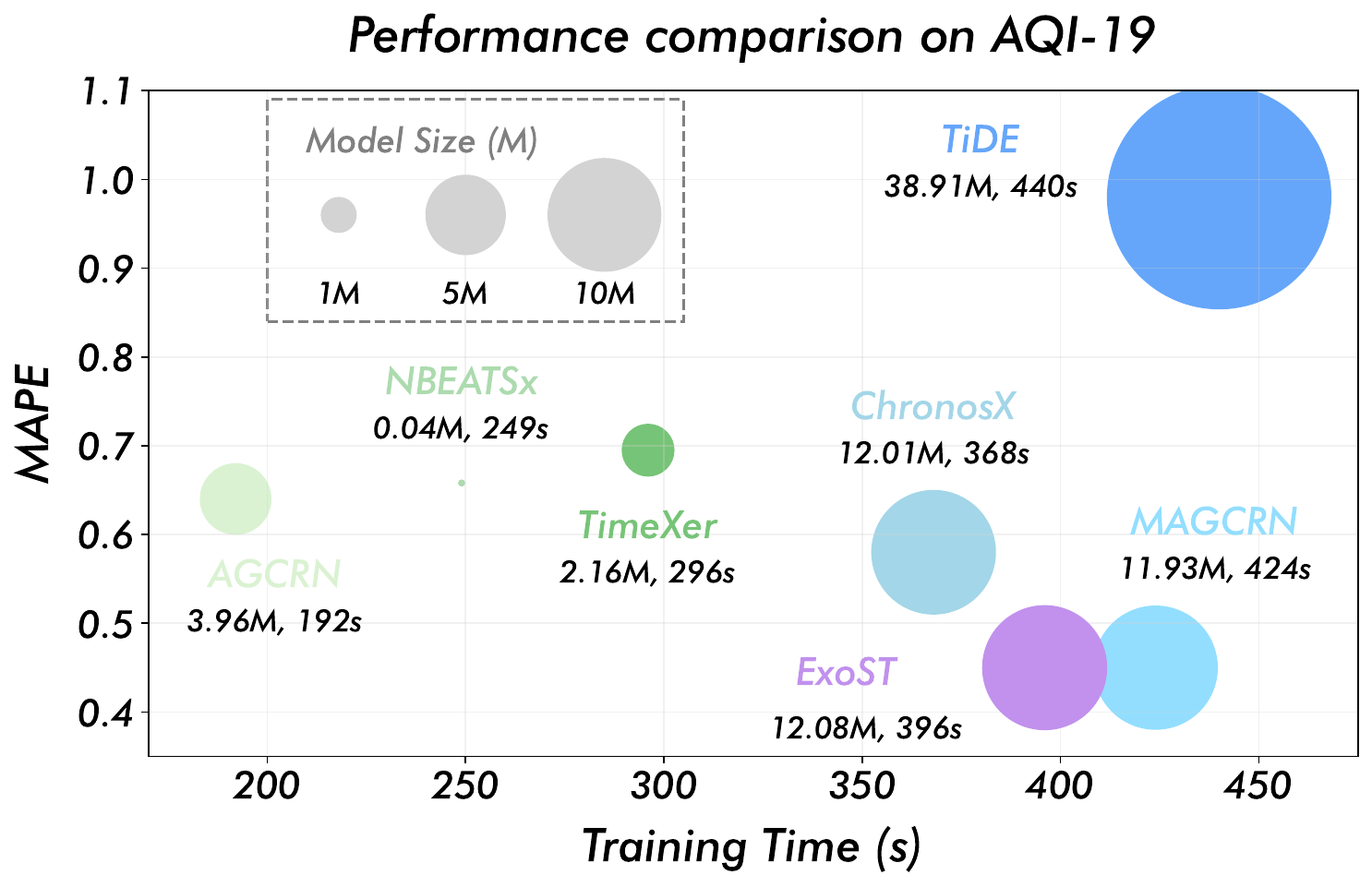}
\setlength{\abovecaptionskip}{-0.3cm}
\caption{Efficiency (first 100 epochs) \& Lightweight Study across different models.} 
\vspace{-4mm}
\label{fig:lightweight_study}
\end{center}
\end{wrapfigure} 
We evaluated the \textit{training time}, \textit{memory footprint}, and \textit{performance} of \model against other forecasting models incorporating exogenous variables on the \AirNineteen~dataset. To ensure fair comparison given varying convergence rates, we report the average training time (first 100 epochs). As shown in Figure \ref{fig:lightweight_study}, \model strikes a favorable balance between performance, efficiency, and parameter count. Specifically: \ding{182} Compared to time series models, purely lightweight designs often lead to inferior performance. 
\ding{183} Compared to ST and general exogenous variable modeling methods using the same backbone (\ie, MAGCRN, ChronosX), our approach achieves superior performance while maintaining comparable parameter counts and training times.


\section{Conclusion and Future Work}


This paper presents the first systematic study of the challenges inherent in modeling exogenous variables for spatio-temporal forecasting and explores effective implementation strategies. We propose a simple yet effective general framework, \model. This framework follows a ``select, then balance'' paradigm, which effectively addresses the problems of inconsistent variable effects and unbalanced type effects. Extensive experiments demonstrate its effectiveness, universality, and robustness. In future work, we aim to explore how to expand the modality availability of exogenous variables and their feasibility in extremely data-scarce scenarios.


\clearpage
\section*{Acknowledgment}
This work is mainly supported by the National Natural Science Foundation of China (No. 62402414). This work is also supported by the Guangzhou-HKUST(GZ) Joint Funding Program (No. 2024A03J0620), Guangzhou Municipal Science and Technology Project (No. 2023A03J0011), the Guangzhou Industrial Information and Intelligent Key Laboratory Project (No. 2024A03J0628), and a grant from State Key Laboratory of Resources and Environmental Information System, and Guangdong Provincial Key Lab of Integrated Communication, Sensing and Computation for Ubiquitous Internet of Things (No. 2023B1212010007).

\definecolor{textgray}{HTML}{6E6E73}
\makeatletter
\newcommand\applefootnote[1]{%
  \begingroup
  \renewcommand\thefootnote{}%
  \renewcommand\@makefntext[1]{\noindent##1}%
  \footnote{#1}%
  \addtocounter{footnote}{-1}%
  \endgroup
}
\makeatother

\bibliography{ref}
\bibliographystyle{plainnat}
\newpage
\appendix

\section*{Appendix}
\appendix
\counterwithin{figure}{section}
\counterwithin{table}{section}
\fancypagestyle{appendixfooter}{
  \fancyhf{} 
  \renewcommand{\headrulewidth}{0pt}
  \renewcommand{\footrulewidth}{0pt}
  \fancyfoot[L]{\hyperlink{appendix-start}{{\textit{Go to Appendix Index}}}}
  \fancyfoot[C]{\thepage}
}

\hypertarget{appendix-start}{}
\pagestyle{appendixfooter}


\vspace{1.5em}

\hrule height .8pt
\DoToC
\hrule height .8pt

\vspace{1.5em}

\clearpage

\section{More Analysis}\label{analysis}

\subsection{More Analysis on Select Module}\label{appendix_analysis_select}

We provide theoretical insights into why the proposed select stage (conditional embedding + latent-space gated experts) improves exogenous-aware spatio-temporal forecasting, addressing the challenge of {inconsistent variable effects}. Our analysis is backbone-agnostic and focuses on representation, approximation, and optimization perspectives.

\ding{70} \textit{Input-dependent linear operator and regime-adaptive processing.}
The gated expert selector in Eq.~(\ref{eq:moe}) can be reformulated as an input-dependent linear operator:
$$\mathbf{X}^{{\tau}^{\prime}} = W(\mathbf{X}^{\tau})\,\mathbf{X}^{\tau}, \quad W(\mathbf{X}^{\tau})=\sum_{k=1}^{K} g_k^{\tau}(\mathbf{X}^{\tau})\,W_k^{\tau},
\label{eq:dynamic_operator}$$
where $g^{\tau}(\cdot)$ is a simplex-valued gate. Unlike a static projection that enforces a global transformation, Eq.~(\ref{eq:dynamic_operator}) implements a {dynamic filter} whose parameters evolve with local spatio-temporal contexts encoded in $\mathbf{X}^{\tau}$. This explicitly models the inconsistent effects, where the contribution of endogenous channels varies across different weather conditions or temporal regimes.

\ding{70} \textit{Latent regime approximation.}
We postulate that the data-generating process involves a continuous latent regime variable $z \in \mathcal{Z}$ (\eg, a continuous spectrum from clear to extreme weather). The predictive distribution can be viewed as an integral over these regimes:
$$
p(\mathbf{Y}\mid \mathbf{X},\mathbf{E}) = \int_{\mathcal{Z}} p(\mathbf{Y}\mid \mathbf{X}, z)\,p(z\mid \mathbf{E})\,dz.
\label{eq:regime_integral}$$
Our module approximates this integral via discretization. The conditional embedding $\phi_{\mathrm{cond}}^{\tau}$ aligns heterogeneous sources, enabling the gate $g^{\tau}(\mathbf{X}^{\tau})$ to approximate the regime posterior $p(z\mid \mathbf{E})$. The $K$ experts $W_k^{\tau}$ then serve as support points in the function space, effectively decomposing the complex multimodal mapping into a mixture of local experts:
$$p(\mathbf{Y}\mid \mathbf{X},\mathbf{E}) \approx \sum_{k=1}^{K} g_k^{\tau}(\mathbf{X}^{\tau})\,p_k(\mathbf{Y}\mid \mathbf{X}).$$

\ding{70} \textit{Reduced gradient interference via soft routing.}
Inconsistent effects also manifest as {gradient conflicts}: samples from different contexts induce updates in opposing directions when sharing parameters. For an expert parameter $\theta_k$ (e.g., $W_k^{\tau}$), the per-sample gradient satisfies
$$
\nabla_{\theta_k}\mathcal{L}_i \approx g_{i,k}^{\tau}\,\nabla_{\theta_k}\widetilde{\mathcal{L}}_i,
\label{eq:routed_grad}
$$
where $g_{i,k}^{\tau}$ is the gate weight for sample $i$. Thus, samples primarily update the experts they activate, which {decomposes} conflicting gradients across experts and reduces negative transfer. Empirically, this is consistent with smoother gate trajectories and clearer expert specialization (Fig.~\ref{fig:gate_weight}), improving optimization stability and downstream forecasting accuracy.

Overall, the key advantage of the Select Module is that it explicitly incorporates exogenous-condition-induced variations in variable effects into both the representation learning and routing mechanisms. Specifically, the conditional embedding first fuses endogenous histories with heterogeneous, multi-source exogenous signals into context-dependent latent representations, thereby establishing a unified semantic basis for subsequent selection. Building on this latent space, the gated experts further instantiate an input-dependent dynamic operator, enabling the model to automatically activate more appropriate processing pathways across different time steps, spatial nodes, and exogenous contexts. Meanwhile, soft routing assigns the learning signal from each contextual sample primarily to its corresponding expert parameters, effectively alleviating gradient interference and negative transfer that commonly arise under shared-parameter training, which in turn promotes context-specialized expert division of labor and improves optimization efficiency. Consequently, by combining {conditional representation} with {expert selection}, the Select stage realizes adaptive modeling of inconsistent variable effects and provides the downstream backbone with more robust and controllable input features.

\subsection{More Analysis on Balance Module}\label{appendix_analysis_balance}

Although existing methods~\citep{wang2024timexer,giganti2024back,Arango2025ChronosX} have explored various strategies for balancing past and future exogenous variables, our approach introduces a fundamental distinction in its design philosophy. We theoretically scrutinize the discrepancies among these methods through the lens of functional complexity and the representational capacity of their respective function spaces:

\begin{itemize}[noitemsep, topsep=8pt, partopsep=0pt, leftmargin=6mm,parsep=8pt]
    \item[\ding{70}] \textit{Fixed Weighting Strategy:} This approach operates as a rudimentary static linear mapping, formalized as:
    $$
    f(x,y) = (1-{\alpha})x + {\alpha}y
    $$
    The representational capacity of this method is severely constrained, as the dimension of the function space is $1$, governed exclusively by the scalar hyperparameter ${\alpha}$. It imposes a rigid interpolation that is agnostic to data distribution.

    \item[\ding{70}]  \textit{Learnable Weighting Strategy:} While this method introduces optimization, it fundamentally remains a linear mapping within the family of affine transformations:
    $$
    f(x,y) = (1+{\tilde{w}_1})x + (1+{\tilde{w}_2})y
    $$
    Although the parameters are learnable, the objective is to converge upon a single, global optimal weighted balance across the entire training set. Consequently, the learned weights remain static during the inference phase and cannot adapt to the specific variations of individual samples.

    \item[\ding{70}] \textit{Context-Aware Balancing (Ours):} In stark contrast, our proposed mechanism functions as an input-dependent, non-linear mapping:
    $$
    f(x,y) = (1-w(x,y))x + w(x,y)y + (x+y) = \Phi(x,y)
    $$
    Theoretically, this formulation is capable of approximating arbitrary fusion functions. By conditioning the weights $w(x,y)$ on the input itself, the model achieves context sensitivity at inference time, allowing for the dynamic modulation of the fusion strategy based on immediate data characteristics.
\end{itemize}

Therefore, our methodology diverges significantly from prior static strategies by offering superior dynamic adaptability. To validate this theoretical advantage empirically, Section~\ref{mechanism_study} presents a comprehensive ablation study comparing our approach against degenerate fixed-weight baselines, learnable-weight strategies, and attention-based mechanisms. The empirical results robustly demonstrate the efficacy and necessity of our context-aware balancing strategy.

\section{Experimental Details}\label{appendix_setup}

\subsection{Datasets Details}
\label{appendix_datasets}

The statistics of the two real-world datasets used in this paper are summarized in Table~\ref{tab:dataset}.

\begin{table}[h]
\centering
\caption{Summary of datasets used for our experiments. M: million ($10^6$).}
\label{tab:dataset}
\begin{tabular}{@{}lcc@{}}
\toprule
\textbf{Attribute} & \textbf{Madrid-19} & \textbf{Madrid-22} \\ \midrule
Time range & 01/01/2019 – 30/06/2019 & 01/01/2022 – 30/06/2022 \\
Frames & 4344 & 4344 \\
Sampling Rate & 1 hour & 1 hour \\
Data points & 2.03 M & 2.03 M \\
Variables & 20 & 20 \\

\multirow{2}{*}[1.5ex]{\centering Variable Domain}
& \makecell[c]{Air quality, Traffic, \\ Meteorological, Date} & \makecell[c]{Air quality, Traffic, \\ Meteorological, Date} \\ \bottomrule
\end{tabular}
\end{table}

\noindent\textbf{Graph Construction for GNN-based Models.} We classify the GNN models into \textit{adaptive graph construction} and \textit{predefined graph structure} based on their graph construction approaches and perform graph construction on datasets accordingly. 
For \textit{adaptive graph construction} models including AGCRN, GraphWaveNet, GGNN, and MAGCRN, the adjacency matrix $\mathbf{A} \in \mathbb{R}^{N \times N}$ is computed by embedding the learnable node. 
Specifically, given node embeddings $\mathbf{E} \in \mathbb{R}^{N \times d_e}$ where $d_e$ is the embedding dimension, AGCRN computes the adjacency matrix as $\mathbf{A} = \text{softmax}(\text{ReLU}(\mathbf{E}\mathbf{E}^\top))$, where $\mathbf{E}\mathbf{E}^\top$ computes the similarity between all pairs of nodes, ReLU ensures non-negative similarities, and softmax normalizes the weights to create a valid adjacency matrix. 
This approach allows the model to automatically discover the latent spatial dependencies between the monitoring stations without requiring prior knowledge of their geographical relationships. 
GWNet uses separate source $\mathbf{E}_s$ and target embeddings $\mathbf{E}_t$ to compute the adjacency matrix as $\mathbf{A} = \text{softmax}(\text{ReLU}(\mathbf{E}_s\mathbf{E}_t^\top))$, allowing more flexible spatial relationship modeling. 
GGNN employs a fully connected graph structure where $\mathbf{A}_{ij} = 1$ for all $i,j \in \{1,2,\ldots,N\}$, enabling dense spatial attention across all pairs of node. 
MAGCRN inherits the adaptive graph construction capability from AGCRN, automatically learning the graph structure through node embeddings. 
For \textit{predefined graph structure} models that include STGCN, DCRNN, and GRUGCN, due to the lack of geographic location, we follow the mainstream practice~\footnote{\href{https://github.com/TorchSpatiotemporal/tsl}{Torch-Spatiotemporal}} and construct the adjacency matrix based on the Pearson correlation coefficient of the target time series.
Given the target variable $\mathbf{X}_t \in \mathbb{R}^{N \times F}$ at time $t$, we compute the correlation matrix $\mathbf{C} \in \mathbb{R}^{N \times N}$ where $\mathbf{C}_{ij} = \rho_{ij}$ represents the Pearson correlation coefficient between nodes $i$ and $j$. 
The adjacency matrix is then constructed as $\mathbf{A}_{ij} = \begin{cases} \rho{ij}, & \text{if } j \in \text{TopK}(\rho_{i,:}) \\ 0, & \text{otherwise} \end{cases}$, where the $TopK$ operation retains the $k$ most correlated connections for each node, specifically choosing $k=8$.

\vspace{0.2em}
\noindent\textbf{Variable Scales and Units.} 
For the \AirNineteen~and \AirTwenty, the target variable $\text{NO}_2$ ranges from 0 to 328 $\mu$g/m$^3$ in Madrid-19 and 0 to 625 $\mu$g/m$^3$ in Madrid-22, with mean values of 36.6 and 28.0 $\mu$g/m$^3$ respectively.
For the \SpeedNineteen~and \TrafficTwenty, the target variables have various scales: the average traffic speed in Madrid-19 ranges from 0 to 3.57 $km/h$ with a mean of approximately 1.2 $km/h$, while the traffic intensity in Madrid-22 ranges from 56.5 to 289.88 vehicles/hour with a mean of approximately 170 vehicles/hour.
Traffic variables show substantial scale differences: traffic intensity ranges from $56$ to $429~vehicles/hour$, traffic occupation time varies from 0.5\% to 5.76\%, traffic load ranges from 6.5\% to 28.12\%, and traffic average speed ranges from $0$ to $3.57 km/h$.
Exogenous variables maintain consistent scales across all tasks, with meteorological variables serving as future covariates and traffic variables (excluding the target) serving as past covariates.
To handle these diverse scales effectively, we apply StandardScaler normalization to all variables, ensuring that each variable has zero mean and unit variance.

\vspace{0.2em}
\noindent\textbf{Temporal Variable Encoding.} We employ consistent temporal encoding in four forecasting tasks using sinusoidal transformations for periodic features and one-hot encoding for categorical information.
For continuous periodic features, given a timestamp $t$, the hour encoding is calculated as $\sin(2\pi \cdot \text{hour}(t) / 24)$ and $\cos(2\pi \cdot \text{hour}(t) / 24)$, and the month encoding as $\sin(2\pi \cdot (\text{month}(t) - 1) / 12)$ and $\cos(2\pi \cdot (\text{month}(t) - 1) / 12)$.
For categorical temporal information, each weekday is represented as a binary vector $\mathbf{w} \in \{0,1\}^7$, where $\mathbf{w}_i = 1$ if the day corresponds to the $i$-th day of the week (Monday=0, Sunday=6), and $\mathbf{w}_i = 0$ otherwise.
These temporal features are added as \textit{universal exogenous} variables to four  tasks.

\subsection{Baseline Details}
\label{appendix_baseline}

In this appendix, we provide detailed descriptions of the advanced exogenous variable modeling methods and different spatio-temporal backbone models used in our default evaluation. 

\subsubsection{Exogenous Variable Modeling Methods}

These methods are grouped into three main categories based on their approach to handling exogenous variables.

\vspace{0.2em}
\textbf{TS Models w/ Exogenous Variable Integration.}

\vspace{0.5em}
\begin{itemize}[noitemsep, topsep=8pt, partopsep=0pt, leftmargin=6mm,parsep=8pt]
\setlength\itemsep{0mm}
\item \textit{TiDE}~\citep{das2023long}: \textit{TiDE} is an encoder–decoder model built on multilayer perceptrons (MLPs) that fuses historical time-series data and covariates via dense MLP residual blocks to map them into low-dimensional representations, then generates forecast sequences by incorporating future covariates. \url{https://github.com/google-research/google-research/tree/master/tide}
\item \textit{TimeXer}~\citep{wang2024timexer}: \textit{TimeXer} is a Transformer model tailored for time series forecasting with exogenous variables: it captures temporal dependencies in the endogenous series via patch-level self-attention and effectively integrates exogenous information through variable-level cross attention and learnable global endogenous tokens to produce high-quality predictions of the target series. \url{https://github.com/thuml/TimeXer}
\item \textit{NBEATSx}~\citep{olivares2023neural}: \textit{NBEATSx} extends the original N-BEATS by incorporating convolutional encoding substructures for time-varying and static exogenous variables within a hierarchical stacked residual block architecture, enabling interpretable decomposition of trend, seasonality, and exogenous factors for high-quality target-series forecasting. \url{https://github.com/cchallu/nbeatsx}
\item \textit{CrossLinear}~\citep{zhou2025crosslinear}:\textit{CrossLinear} is a Linear-based model that introduces a plug-and-play cross-correlation embedding module to capture time-invariant dependencies between endogenous and exogenous variables, utilizing patch-wise processing and a global linear head to effectively model both short-term and long-term temporal patterns for precise forecasting. \url{https://github.com/mumiao2000/CrossLinear}
\item \textit{DAG}~\citep{qiu2025dagdualcausalnetwork}: \textit{DAG} is a general framework that utilizes a Dual Causal Network along temporal and channel dimensions, it explicitly models how historical exogenous variables affect future exogenous variables (temporal) and historical endogenous variables (channel), and then injects these discovered causal relationships to improve the prediction of future endogenous variables. \url{https://github.com/decisionintelligence/DAG}
\end{itemize}

\vspace{0.2em}
\textbf{ST Models w/ Exogenous Variable Integration.}

\vspace{0.5em}
\begin{itemize}[noitemsep, topsep=8pt, partopsep=0pt, leftmargin=6mm,parsep=8pt]
\setlength\itemsep{0mm}
\item \textit{MAGCRN}~\citep{giganti2024back}: \textit{MAGCRN} is a spatio-temporal model that extends AGCRN into a dual-path architecture, using separate AGCRN blocks to encode historical observations and future exogenous covariates. \url{https://github.com/polimi-ispl/MAGCRN}
\end{itemize}

\vspace{0.2em}
\textbf{General Exogenous Variable Integration Framework.}

\vspace{0.5em}
\begin{itemize}[noitemsep, topsep=8pt, partopsep=0pt, leftmargin=6mm,parsep=8pt]
\setlength\itemsep{0mm}
\item \textit{ChronosX}~\citep{Arango2025ChronosX}: \textit{ChronosX} extends MAGCRN by integrating two exogenous adaptation modules: Instance-wise Interpolation Block (IIB) and Offset Integration Block (OIB). 
IIB adaptively interpolates exogenous inputs across time steps, while OIB injects time-dependent bias offsets into the output layer, enabling more flexible integration of future covariates. 
Due to the limited public availability of the code, we implemented the IIB and OIB modules based on the formulas provided in the referenced paper. \url{https://github.com/amazon-science/chronos-forecasting/tree/chronosx}
\end{itemize}

\subsubsection{Spatio-Temporal Backbone Models}

To ensure fair use of these models, we use a popular spatio-temporal forecasting benchmark library: \url{https://github.com/TorchSpatiotemporal/tsl}.

\begin{itemize}[noitemsep, topsep=8pt, partopsep=0pt, leftmargin=6mm,parsep=8pt]
\setlength\itemsep{0mm}
\item \textit{AGCRN} ~\citep{Bai2020Adaptive}: \textit{AGCRN} is an adaptive graph-based recurrent model that captures fine-grained spatio-temporal dependencies by jointly learning node-specific patterns and data-driven graph structures, eliminating the need for pre-defined adjacency matrices.
\item \textit{GWNet}~\citep{wu2019graph}: \textit{GWNet} is a spatial-temporal model that combines diffusion-based graph convolutions with dilated causal temporal convolutions, and incorporates a learnable self-adaptive adjacency matrix to capture latent spatial relationships from data.
\item \textit{GGNN}~\citep{Satorras2022Multivariate}: \textit{GGNN} is a recurrent graph-based model that applies gated update mechanisms to iteratively refine node representations through message passing, enabling effective modeling of long-range structural dependencies.
\item \textit{GRUGCN}~\citep{Gao2022On}: \textit{GRUGCN} is a spatio-temporal neural architecture that integrates graph convolutions for spatial feature extraction and gated recurrent units for sequential modeling, effectively capturing dynamic dependencies in traffic forecasting scenarios.
\item \textit{STGCN}~\citep{yu2017spatio}: \textit{STGCN} is a graph-based temporal forecasting model that integrates graph convolutional layers with temporal convolution modules to jointly learn spatial and temporal dependencies in structured time series data.
\item \textit{DCRNN}~\citep{li2017diffusion}: \textit{DCRNN} is a graph-based recurrent forecasting model that captures spatial dependencies via diffusion convolutions on directed graphs and models temporal dynamics through a sequence-to-sequence recurrent architecture.
\end{itemize}

\subsection{Protocol Details}
\label{appendix_protocol}

\subsubsection{Metrics Detail.}
We use different metrics such as MAE, RMSE, MRE, and MAPE. Formally, these metrics are formulated as follows:
\begin{align*}
\text{MAE} &= \frac{1}{n} \sum_{i=1}^n |y_i - \hat{y}_i|, &
\text{RMSE} &= \sqrt{\frac{1}{n} \sum_{i=1}^n (y_i - \hat{y}_i)^2}, \\
\text{MRE} &= \frac{\sum_{i=1}^n |y_i - \hat{y}_i|}{\sum_{i=1}^n |y_i|}, &
\text{MAPE} &= \frac{100\%}{n} \sum_{i=1}^n \left| \frac{\hat{y}_i - y_i}{y_i} \right|,
\end{align*}
where $n$ represents the indices of all observed samples, $y_i$ denotes the $i$-th actual sample and $\hat{y}_i$ is the corresponding prediction. 
The mean relative error is defined as the MAE divided by the $\ell_1$ norm of the target window. 

\subsubsection{Parameter Detail.} The model is trained for 500 epochs using the AdamW optimizer with a cosine annealing learning rate scheduler. 
The learning rate is initialized at $10^{-2}$ and decays to $10^{-7}$ over the training period. 
A batch size of 512 is used, incorporating mixed-precision arithmetic and Z-score normalization for target variables. 
Early stopping is implemented to prevent overfitting, halting training if the validation MAE does not improve for 30 consecutive epochs. 
All experiments are carried out on a Linux server equipped with a 1 $\times$ AMD EPYC 7763 128-Core Processor CPU with 256GB memory and 4 $\times$ NVIDIA RTX A6000 GPUs, each with 48GB memory. 


\section{More Results}\label{strategy}

\subsection{More Result on Universality Study}
\label{appendix_universality}

\begin{table*}[htbp!]
  \setlength{\tabcolsep}{4.0pt}
  \centering
  \caption{Performance comparison of different models w/ and w/o \model on common benchmarks (3-day scenario).}
  \resizebox{\linewidth}{!}{%
  \begin{sc}
  \renewcommand{\arraystretch}{1.8}
  \begin{tabular}{lccccccccccccccccccc}
    \toprule
    & Model & \multicolumn{3}{c}{{AGCRN}}
      & \multicolumn{3}{c}{{GWNet}}
      & \multicolumn{3}{c}{{GGNN}}
      & \multicolumn{3}{c}{{GRUGCN}}
      & \multicolumn{3}{c}{{STGCN}}
      & \multicolumn{3}{c}{{DCRNN}}\\
    \multicolumn{2}{c}{w/ \model} & \rxmark & \gcmark & {$\Delta$(\%)}
     & \rxmark & \gcmark & {$\Delta$(\%)}
     & \rxmark & \gcmark & {$\Delta$(\%)}
     & \rxmark & \gcmark & {$\Delta$(\%)}
     & \rxmark & \gcmark & {$\Delta$(\%)}
     & \rxmark & \gcmark & {$\Delta$(\%)}\\
    \cmidrule(lr){1-2}\cmidrule(lr){3-5}\cmidrule(lr){6-8}\cmidrule(lr){9-11}\cmidrule(lr){12-14}\cmidrule(lr){15-17}\cmidrule(lr){18-20}
\multirow{4}{*}{\AirNineteen}
  & MAE
  & $14.05\textcolor{gray}{\text{\scriptsize±0.20}}$ & $\firstres{{10.41}}\textcolor{gray}{\text{\scriptsize±0.12}}$ & \cellcolor{gray!8}\boldmath{$\downarrow 25.91$}
  & $12.67\textcolor{gray}{\text{\scriptsize±2.25}}$ & $13.33\textcolor{gray}{\text{\scriptsize±0.41}}$ & \cellcolor{gray!8}\boldmath{$\uparrow 5.21$}
  & $14.39\textcolor{gray}{\text{\scriptsize±0.22}}$ & $12.30\textcolor{gray}{\text{\scriptsize±1.08}}$ & \cellcolor{gray!8}\boldmath{$\downarrow 14.52$}
  & $14.10\textcolor{gray}{\text{\scriptsize±0.09}}$ & $12.83\textcolor{gray}{\text{\scriptsize±0.55}}$ & \cellcolor{gray!8}\boldmath{$\downarrow 9.01$}
  & $15.17\textcolor{gray}{\text{\scriptsize±1.31}}$ & $13.14\textcolor{gray}{\text{\scriptsize±0.94}}$ & \cellcolor{gray!8}\boldmath{$\downarrow 13.38$}
  & $14.75\textcolor{gray}{\text{\scriptsize±0.20}}$ & $\secondres{{11.52}}\textcolor{gray}{\text{\scriptsize±0.40}}$ & \cellcolor{gray!8}\boldmath{$\downarrow 21.84$} \\
  & RMSE
  & $20.92\textcolor{gray}{\text{\scriptsize±0.47}}$ & $\firstres{{15.51}}\textcolor{gray}{\text{\scriptsize±0.16}}$ & \cellcolor{gray!8}\boldmath{$\downarrow 25.86$}
  & $19.00\textcolor{gray}{\text{\scriptsize±3.52}}$ & $19.81\textcolor{gray}{\text{\scriptsize±0.28}}$ & \cellcolor{gray!8}\boldmath{$\uparrow 4.26$}
  & $20.88\textcolor{gray}{\text{\scriptsize±0.07}}$ & $18.57\textcolor{gray}{\text{\scriptsize±1.62}}$ & \cellcolor{gray!8}\boldmath{$\downarrow 11.06$}
  & $21.20\textcolor{gray}{\text{\scriptsize±0.04}}$ & $19.58\textcolor{gray}{\text{\scriptsize±0.54}}$ & \cellcolor{gray!8}\boldmath{$\downarrow 7.64$}
  & $21.72\textcolor{gray}{\text{\scriptsize±0.61}}$ & $19.90\textcolor{gray}{\text{\scriptsize±1.08}}$ & \cellcolor{gray!8}\boldmath{$\downarrow 8.47$}
  & $21.76\textcolor{gray}{\text{\scriptsize±0.32}}$ & $\secondres{{17.51}}\textcolor{gray}{\text{\scriptsize±0.86}}$ & \cellcolor{gray!8}\boldmath{$\downarrow 19.53$} \\
  & MAPE(\%)
  & $63.79\textcolor{gray}{\text{\scriptsize±1.32}}$ & $\firstres{{44.65}}\textcolor{gray}{\text{\scriptsize±0.08}}$ & \cellcolor{gray!8}\boldmath{$\downarrow 30.00$}
  & $63.50\textcolor{gray}{\text{\scriptsize±2.08}}$ & $62.43\textcolor{gray}{\text{\scriptsize±3.94}}$ & \cellcolor{gray!8}\boldmath{$\downarrow 1.68$}
  & $73.36\textcolor{gray}{\text{\scriptsize±1.23}}$ & $53.55\textcolor{gray}{\text{\scriptsize±8.41}}$ & \cellcolor{gray!8}\boldmath{$\downarrow 27.00$}
  & $69.99\textcolor{gray}{\text{\scriptsize±1.47}}$ & $50.96\textcolor{gray}{\text{\scriptsize±2.82}}$ & \cellcolor{gray!8}\boldmath{$\downarrow 27.20$}
  & $82.69\textcolor{gray}{\text{\scriptsize±20.81}}$ & $57.21\textcolor{gray}{\text{\scriptsize±7.64}}$ & \cellcolor{gray!8}\boldmath{$\downarrow 30.70$}
  & $73.64\textcolor{gray}{\text{\scriptsize±6.33}}$ & $\secondres{{46.56}}\textcolor{gray}{\text{\scriptsize±2.39}}$ & \cellcolor{gray!8}\boldmath{$\downarrow 36.81$} \\
  & MRE(\%)
  & $54.36\textcolor{gray}{\text{\scriptsize±0.77}}$ & $\firstres{{40.34}}\textcolor{gray}{\text{\scriptsize±0.47}}$ & \cellcolor{gray!8}\boldmath{$\downarrow 25.79$}
  & $54.05\textcolor{gray}{\text{\scriptsize±0.50}}$ & $51.65\textcolor{gray}{\text{\scriptsize±1.60}}$ & \cellcolor{gray!8}\boldmath{$\downarrow 4.44$}
  & $56.39\textcolor{gray}{\text{\scriptsize±1.93}}$ & $47.67\textcolor{gray}{\text{\scriptsize±4.20}}$ & \cellcolor{gray!8}\boldmath{$\downarrow 15.46$}
  & $54.55\textcolor{gray}{\text{\scriptsize±0.33}}$ & $49.71\textcolor{gray}{\text{\scriptsize±2.15}}$ & \cellcolor{gray!8}\boldmath{$\downarrow 8.87$}
  & $58.89\textcolor{gray}{\text{\scriptsize±5.27}}$ & $50.91\textcolor{gray}{\text{\scriptsize±3.66}}$ & \cellcolor{gray!8}\boldmath{$\downarrow 13.67$}
  & $57.09\textcolor{gray}{\text{\scriptsize±0.78}}$ & $\secondres{{44.62}}\textcolor{gray}{\text{\scriptsize±1.55}}$ & \cellcolor{gray!8}\boldmath{$\downarrow 21.84$} \\

\midrule
\multirow{4}{*}{\SpeedNineteen$*$}
  & MAE
  & $3.94\textcolor{gray}{\text{\scriptsize±0.01}}$ & $\firstres{{3.80}}\textcolor{gray}{\text{\scriptsize±0.01}}$ & \cellcolor{gray!8}\boldmath{$\downarrow 3.55$}
  & $4.03\textcolor{gray}{\text{\scriptsize±0.06}}$ & $\secondres{{3.86}}\textcolor{gray}{\text{\scriptsize±0.06}}$ & \cellcolor{gray!8}\boldmath{$\downarrow 4.22$}
  & $22.53\textcolor{gray}{\text{\scriptsize±3.92}}$ & $4.53\textcolor{gray}{\text{\scriptsize±0.38}}$ & \cellcolor{gray!8}\boldmath{$\downarrow 79.89$}
  & $38.48\textcolor{gray}{\text{\scriptsize±0.02}}$ & $4.63\textcolor{gray}{\text{\scriptsize±0.63}}$ & \cellcolor{gray!8}\boldmath{$\downarrow 87.97$}
  & $6.45\textcolor{gray}{\text{\scriptsize±0.48}}$ & $4.64\textcolor{gray}{\text{\scriptsize±0.45}}$ & \cellcolor{gray!8}\boldmath{$\downarrow 28.06$}
  & $4.87\textcolor{gray}{\text{\scriptsize±0.05}}$ & $4.03\textcolor{gray}{\text{\scriptsize±0.20}}$ & \cellcolor{gray!8}\boldmath{$\downarrow 17.25$} \\
  & RMSE
  & $\secondres{{31.88}}\textcolor{gray}{\text{\scriptsize±0.02}}$ & $\firstres{{31.85}}\textcolor{gray}{\text{\scriptsize±0.02}}$ & \cellcolor{gray!8}\boldmath{$\downarrow 0.09$}
  & $32.00\textcolor{gray}{\text{\scriptsize±0.06}}$ & $32.05\textcolor{gray}{\text{\scriptsize±0.05}}$ & \cellcolor{gray!8}\boldmath{$\uparrow 0.16$}
  & $45.89\textcolor{gray}{\text{\scriptsize±3.35}}$ & $32.89\textcolor{gray}{\text{\scriptsize±0.12}}$ & \cellcolor{gray!8}\boldmath{$\downarrow 28.32$}
  & $139.12\textcolor{gray}{\text{\scriptsize±0.01}}$ & $32.63\textcolor{gray}{\text{\scriptsize±0.45}}$ & \cellcolor{gray!8}\boldmath{$\downarrow 76.54$}
  & $44.46\textcolor{gray}{\text{\scriptsize±0.50}}$ & $33.84\textcolor{gray}{\text{\scriptsize±0.60}}$ & \cellcolor{gray!8}\boldmath{$\downarrow 23.90$}
  & $34.49\textcolor{gray}{\text{\scriptsize±0.38}}$ & $32.26\textcolor{gray}{\text{\scriptsize±0.09}}$ & \cellcolor{gray!8}\boldmath{$\downarrow 6.50$} \\
  & MAPE(\%)
  & $\secondres{{9.42}}\textcolor{gray}{\text{\scriptsize±0.03}}$ & $\firstres{{9.31}}\textcolor{gray}{\text{\scriptsize±0.03}}$ & \cellcolor{gray!8}\boldmath{$\downarrow 1.17$}
  & $9.54\textcolor{gray}{\text{\scriptsize±0.23}}$ & $9.65\textcolor{gray}{\text{\scriptsize±0.23}}$ & \cellcolor{gray!8}\boldmath{$\uparrow 1.14$}
  & $14.15\textcolor{gray}{\text{\scriptsize±0.27}}$ & $10.63\textcolor{gray}{\text{\scriptsize±0.18}}$ & \cellcolor{gray!8}\boldmath{$\downarrow 24.90$}
  & $99.86\textcolor{gray}{\text{\scriptsize±0.01}}$ & $10.67\textcolor{gray}{\text{\scriptsize±0.50}}$ & \cellcolor{gray!8}\boldmath{$\downarrow 89.33$}
  & $14.02\textcolor{gray}{\text{\scriptsize±0.46}}$ & $11.20\textcolor{gray}{\text{\scriptsize±0.81}}$ & \cellcolor{gray!8}\boldmath{$\downarrow 20.11$}
  & $11.70\textcolor{gray}{\text{\scriptsize±0.14}}$ & $9.73\textcolor{gray}{\text{\scriptsize±0.20}}$ & \cellcolor{gray!8}\boldmath{$\downarrow 16.84$} \\
  & MRE(\%)
  & $10.21\textcolor{gray}{\text{\scriptsize±0.02}}$ & $\firstres{{9.90}}\textcolor{gray}{\text{\scriptsize±0.03}}$ & \cellcolor{gray!8}\boldmath{$\downarrow 3.04$}
  & $10.49\textcolor{gray}{\text{\scriptsize±0.52}}$ & $\secondres{{10.05}}\textcolor{gray}{\text{\scriptsize±0.16}}$ & \cellcolor{gray!8}\boldmath{$\downarrow 4.20$}
  & $58.60\textcolor{gray}{\text{\scriptsize±10.19}}$ & $11.78\textcolor{gray}{\text{\scriptsize±0.69}}$ & \cellcolor{gray!8}\boldmath{$\downarrow 79.90$}
  & $100.09\textcolor{gray}{\text{\scriptsize±0.05}}$ & $12.04\textcolor{gray}{\text{\scriptsize±1.78}}$ & \cellcolor{gray!8}\boldmath{$\downarrow 87.97$}
  & $16.60\textcolor{gray}{\text{\scriptsize±1.03}}$ & $12.06\textcolor{gray}{\text{\scriptsize±0.92}}$ & \cellcolor{gray!8}\boldmath{$\downarrow 27.35$}
  & $12.68\textcolor{gray}{\text{\scriptsize±0.13}}$ & $10.49\textcolor{gray}{\text{\scriptsize±0.34}}$ & \cellcolor{gray!8}\boldmath{$\downarrow 17.27$} \\
\midrule
\multirow{4}{*}{\AirTwenty}
  & MAE & $10.19\textcolor{gray}{\text{\scriptsize±0.21}}$ & $\firstres{{7.94}}\textcolor{gray}{\text{\scriptsize±0.29}}$ & \cellcolor{gray!8}\boldmath{$\downarrow 22.08$}
 & $10.05\textcolor{gray}{\text{\scriptsize±0.04}}$ & $9.11\textcolor{gray}{\text{\scriptsize±0.12}}$ & \cellcolor{gray!8}\boldmath{$\downarrow 9.35$}
 & $10.44\textcolor{gray}{\text{\scriptsize±0.04}}$ & $8.82\textcolor{gray}{\text{\scriptsize±0.14}}$ & \cellcolor{gray!8}\boldmath{$\downarrow 15.51$}
 & $9.97\textcolor{gray}{\text{\scriptsize±0.17}}$ & $8.58\textcolor{gray}{\text{\scriptsize±0.19}}$ & \cellcolor{gray!8}\boldmath{$\downarrow 13.96$}
 & $10.65\textcolor{gray}{\text{\scriptsize±0.60}}$ & $9.21\textcolor{gray}{\text{\scriptsize±0.11}}$ & \cellcolor{gray!8}\boldmath{$\downarrow 13.51$}
 & $10.33\textcolor{gray}{\text{\scriptsize±0.72}}$ & $\secondres{{7.70}}\textcolor{gray}{\text{\scriptsize±0.34}}$ & \cellcolor{gray!8}\boldmath{$\downarrow 25.47$}\\
  & RMSE & $15.34\textcolor{gray}{\text{\scriptsize±0.02}}$ & $\firstres{{11.79}}\textcolor{gray}{\text{\scriptsize±0.51}}$ & \cellcolor{gray!8}\boldmath{$\downarrow 23.14$}
  & $15.11\textcolor{gray}{\text{\scriptsize±0.27}}$ & $13.66\textcolor{gray}{\text{\scriptsize±0.08}}$ & \cellcolor{gray!8}\boldmath{$\downarrow 9.60$}
  & $15.13\textcolor{gray}{\text{\scriptsize±0.04}}$ & $13.01\textcolor{gray}{\text{\scriptsize±0.25}}$ & \cellcolor{gray!8}\boldmath{$\downarrow 14.03$}
  & $15.02\textcolor{gray}{\text{\scriptsize±0.19}}$ & $13.29\textcolor{gray}{\text{\scriptsize±0.43}}$ & \cellcolor{gray!8}\boldmath{$\downarrow 11.51$}
  & $15.65\textcolor{gray}{\text{\scriptsize±0.50}}$ & $14.16\textcolor{gray}{\text{\scriptsize±0.17}}$ & \cellcolor{gray!8}\boldmath{$\downarrow 9.53$}
  & $15.25\textcolor{gray}{\text{\scriptsize±0.83}}$ & $\secondres{{11.81}}\textcolor{gray}{\text{\scriptsize±0.61}}$ & \cellcolor{gray!8}\boldmath{$\downarrow 22.49$}\\
  & MAPE(\%) & $56.91\textcolor{gray}{\text{\scriptsize±2.06}}$ & $\firstres{{50.27}}\textcolor{gray}{\text{\scriptsize±1.18}}$ & \cellcolor{gray!8}\boldmath{$\downarrow 11.67$}
  & $60.72\textcolor{gray}{\text{\scriptsize±1.66}}$ & $60.70\textcolor{gray}{\text{\scriptsize±1.57}}$ & \cellcolor{gray!8}\boldmath{$\downarrow 0.03$}
  & $71.58\textcolor{gray}{\text{\scriptsize±1.28}}$ & $59.79\textcolor{gray}{\text{\scriptsize±2.29}}$ & \cellcolor{gray!8}\boldmath{$\downarrow 16.47$}
  & $71.37\textcolor{gray}{\text{\scriptsize±2.27}}$ & $53.48\textcolor{gray}{\text{\scriptsize±1.16}}$ & \cellcolor{gray!8}\boldmath{$\downarrow 25.07$}
  & $73.76\textcolor{gray}{\text{\scriptsize±10.10}}$ & $59.47\textcolor{gray}{\text{\scriptsize±0.81}}$ & \cellcolor{gray!8}\boldmath{$\downarrow 19.41$}
  & $69.19\textcolor{gray}{\text{\scriptsize±6.00}}$ & $\secondres{{51.26}}\textcolor{gray}{\text{\scriptsize±3.68}}$ & \cellcolor{gray!8}\boldmath{$\downarrow 25.93$}\\
  & MRE(\%) & $54.54\textcolor{gray}{\text{\scriptsize±1.12}}$ & $\secondres{{42.50}}\textcolor{gray}{\text{\scriptsize±1.57}}$ & \cellcolor{gray!8}\boldmath{$\downarrow 22.08$}
 & $53.77\textcolor{gray}{\text{\scriptsize±0.20}}$ & $48.75\textcolor{gray}{\text{\scriptsize±0.65}}$ & \cellcolor{gray!8}\boldmath{$\downarrow 9.34$}
 & $55.85\textcolor{gray}{\text{\scriptsize±0.21}}$ & $47.18\textcolor{gray}{\text{\scriptsize±0.74}}$ & \cellcolor{gray!8}\boldmath{$\downarrow 15.52$}
 & $53.35\textcolor{gray}{\text{\scriptsize±0.89}}$ & $45.89\textcolor{gray}{\text{\scriptsize±1.02}}$ & \cellcolor{gray!8}\boldmath{$\downarrow 13.99$}
 & $56.98\textcolor{gray}{\text{\scriptsize±3.20}}$ & $49.28\textcolor{gray}{\text{\scriptsize±0.58}}$ & \cellcolor{gray!8}\boldmath{$\downarrow 13.51$}
 & $55.27\textcolor{gray}{\text{\scriptsize±3.85}}$ & $\firstres{{41.18}}\textcolor{gray}{\text{\scriptsize±1.83}}$ & \cellcolor{gray!8}\boldmath{$\downarrow 25.50$}\\
\midrule
\multirow{4}{*}{\TrafficTwenty}
  & MAE & $85.76\textcolor{gray}{\text{\scriptsize±0.47}}$ & $\secondres{{64.13}}\textcolor{gray}{\text{\scriptsize±3.86}}$ & \cellcolor{gray!8}\boldmath{$\downarrow 25.22$}
 & $87.34\textcolor{gray}{\text{\scriptsize±0.75}}$ & $\firstres{{61.87}}\textcolor{gray}{\text{\scriptsize±3.80}}$ & \cellcolor{gray!8}\boldmath{$\downarrow 29.14$}
 & $84.17\textcolor{gray}{\text{\scriptsize±0.43}}$ & $77.29\textcolor{gray}{\text{\scriptsize±1.21}}$ & \cellcolor{gray!8}\boldmath{$\downarrow 8.17$}
 & $85.03\textcolor{gray}{\text{\scriptsize±1.43}}$ & $75.55\textcolor{gray}{\text{\scriptsize±6.83}}$ & \cellcolor{gray!8}\boldmath{$\downarrow 11.16$}
 & $90.49\textcolor{gray}{\text{\scriptsize±0.75}}$ & $72.17\textcolor{gray}{\text{\scriptsize±1.54}}$ & \cellcolor{gray!8}\boldmath{$\downarrow 20.23$}
 & $87.57\textcolor{gray}{\text{\scriptsize±0.79}}$ & $64.55\textcolor{gray}{\text{\scriptsize±2.61}}$ & \cellcolor{gray!8}\boldmath{$\downarrow 26.29$}\\
  & RMSE & $144.77\textcolor{gray}{\text{\scriptsize±0.94}}$ & $\secondres{{109.01}}\textcolor{gray}{\text{\scriptsize±9.19}}$ & \cellcolor{gray!8}\boldmath{$\downarrow 28.34$}
  & $147.13\textcolor{gray}{\text{\scriptsize±3.61}}$ & $\firstres{{105.44}}\textcolor{gray}{\text{\scriptsize±5.53}}$ & \cellcolor{gray!8}\boldmath{$\downarrow 28.34$}
  & $141.73\textcolor{gray}{\text{\scriptsize±1.35}}$ & $125.02\textcolor{gray}{\text{\scriptsize±2.74}}$ & \cellcolor{gray!8}\boldmath{$\downarrow 11.83$}
  & $141.30\textcolor{gray}{\text{\scriptsize±1.92}}$ & $122.13\textcolor{gray}{\text{\scriptsize±11.30}}$ & \cellcolor{gray!8}\boldmath{$\downarrow 13.62$}
  & $148.09\textcolor{gray}{\text{\scriptsize±0.05}}$ & $117.47\textcolor{gray}{\text{\scriptsize±1.89}}$ & \cellcolor{gray!8}\boldmath{$\downarrow 20.68$}
  & $145.99\textcolor{gray}{\text{\scriptsize±1.77}}$ & $109.83\textcolor{gray}{\text{\scriptsize±2.77}}$ & \cellcolor{gray!8}\boldmath{$\downarrow 24.89$}\\
  & MAPE(\%) & $59.46\textcolor{gray}{\text{\scriptsize±3.18}}$ & $34.52\textcolor{gray}{\text{\scriptsize±2.90}}$ & \cellcolor{gray!8}\boldmath{$\downarrow 41.94$}
  & $44.80\textcolor{gray}{\text{\scriptsize±0.89}}$ & $\firstres{{31.74}}\textcolor{gray}{\text{\scriptsize±1.69}}$ & \cellcolor{gray!8}\boldmath{$\downarrow 29.15$}
  & $47.16\textcolor{gray}{\text{\scriptsize±0.66}}$ & $45.19\textcolor{gray}{\text{\scriptsize±1.92}}$ & \cellcolor{gray!8}\boldmath{$\downarrow 4.18$}
  & $48.76\textcolor{gray}{\text{\scriptsize±0.73}}$ & $44.98\textcolor{gray}{\text{\scriptsize±5.12}}$ & \cellcolor{gray!8}\boldmath{$\downarrow 7.75$}
  & $50.02\textcolor{gray}{\text{\scriptsize±2.67}}$ & $39.01\textcolor{gray}{\text{\scriptsize±0.67}}$ & \cellcolor{gray!8}\boldmath{$\downarrow 22.01$}
  & $46.25\textcolor{gray}{\text{\scriptsize±1.65}}$ & $\secondres{{34.48}}\textcolor{gray}{\text{\scriptsize±2.83}}$ & \cellcolor{gray!8}\boldmath{$\downarrow 25.45$}\\
  & MRE(\%) & $30.97\textcolor{gray}{\text{\scriptsize±0.17}}$ & $\secondres{{23.16}}\textcolor{gray}{\text{\scriptsize±1.40}}$ & \cellcolor{gray!8}\boldmath{$\downarrow 25.22$}
 & $31.54\textcolor{gray}{\text{\scriptsize±0.27}}$ & $\firstres{{22.35}}\textcolor{gray}{\text{\scriptsize±1.37}}$ & \cellcolor{gray!8}\boldmath{$\downarrow 29.14$}
 & $30.40\textcolor{gray}{\text{\scriptsize±0.16}}$ & $27.91\textcolor{gray}{\text{\scriptsize±1.50}}$ & \cellcolor{gray!8}\boldmath{$\downarrow 8.19$}
 & $30.71\textcolor{gray}{\text{\scriptsize±0.51}}$ & $27.29\textcolor{gray}{\text{\scriptsize±2.47}}$ & \cellcolor{gray!8}\boldmath{$\downarrow 11.14$}
 & $32.68\textcolor{gray}{\text{\scriptsize±0.27}}$ & $26.06\textcolor{gray}{\text{\scriptsize±0.56}}$ & \cellcolor{gray!8}\boldmath{$\downarrow 20.39$}
 & $31.62\textcolor{gray}{\text{\scriptsize±0.29}}$ & $23.31\textcolor{gray}{\text{\scriptsize±0.94}}$ & \cellcolor{gray!8}\boldmath{$\downarrow 26.27$}\\
    \bottomrule
  \end{tabular}
  \end{sc}
  }
  \label{tab:rq1_3day}
\end{table*}

\begin{table*}[htbp!]
  \setlength{\tabcolsep}{4.0pt}
  \centering
  \caption{Performance comparison of different models w/ and w/o \model on common benchmarks (2-day scenario).}
  \resizebox{\linewidth}{!}{%
  \begin{sc}
  \renewcommand{\arraystretch}{1.8}
  \begin{tabular}{lccccccccccccccccccc}
    \toprule
    & Model & \multicolumn{3}{c}{{AGCRN}}
      & \multicolumn{3}{c}{{GWNet}}
      & \multicolumn{3}{c}{{GGNN}}
      & \multicolumn{3}{c}{{GRUGCN}}
      & \multicolumn{3}{c}{{STGCN}}
      & \multicolumn{3}{c}{{DCRNN}}\\
    \multicolumn{2}{c}{w/ \model} & \rxmark & \gcmark & {$\Delta$(\%)}
     & \rxmark & \gcmark & {$\Delta$(\%)}
     & \rxmark & \gcmark & {$\Delta$(\%)}
     & \rxmark & \gcmark & {$\Delta$(\%)}
     & \rxmark & \gcmark & {$\Delta$(\%)}
     & \rxmark & \gcmark & {$\Delta$(\%)}\\
    \cmidrule(lr){1-2}\cmidrule(lr){3-5}\cmidrule(lr){6-8}\cmidrule(lr){9-11}\cmidrule(lr){12-14}\cmidrule(lr){15-17}\cmidrule(lr){18-20}
  \multirow{4}{*}{\AirNineteen} 
& MAE 
& $13.77\textcolor{gray}{\text{\scriptsize±0.47}}$ 
& $\firstres{{10.04}}\textcolor{gray}{\text{\scriptsize±0.19}}$ 
& \cellcolor{gray!8}\boldmath{$\downarrow 27.09$} 
& $12.29\textcolor{gray}{\text{\scriptsize±2.45}}$ 
& $12.75\textcolor{gray}{\text{\scriptsize±0.21}}$ 
& \cellcolor{gray!8}\boldmath{$\uparrow 3.74$} 
& $14.28\textcolor{gray}{\text{\scriptsize±0.11}}$ 
& $12.14\textcolor{gray}{\text{\scriptsize±1.08}}$ 
& \cellcolor{gray!8}\boldmath{$\downarrow 14.99$} 
& $13.88\textcolor{gray}{\text{\scriptsize±0.06}}$ 
& $12.58\textcolor{gray}{\text{\scriptsize±0.55}}$ 
& \cellcolor{gray!8}\boldmath{$\downarrow 9.37$} 
& $14.73\textcolor{gray}{\text{\scriptsize±0.66}}$ 
& $12.78\textcolor{gray}{\text{\scriptsize±0.64}}$ 
& \cellcolor{gray!8}\boldmath{$\downarrow 13.24$} 
& $14.27\textcolor{gray}{\text{\scriptsize±0.31}}$ 
& $\secondres{{11.31}}\textcolor{gray}{\text{\scriptsize±0.01}}$ 
& \cellcolor{gray!8}\boldmath{$\downarrow 20.74$} \\

& RMSE 
& $20.54\textcolor{gray}{\text{\scriptsize±0.58}}$ 
& $\firstres{{15.12}}\textcolor{gray}{\text{\scriptsize±0.26}}$ 
& \cellcolor{gray!8}\boldmath{$\downarrow 26.39$} 
& $18.87\textcolor{gray}{\text{\scriptsize±3.73}}$ 
& $19.18\textcolor{gray}{\text{\scriptsize±0.13}}$ 
& \cellcolor{gray!8}\boldmath{$\uparrow 1.64$} 
& $20.81\textcolor{gray}{\text{\scriptsize±0.38}}$ 
& $18.45\textcolor{gray}{\text{\scriptsize±1.58}}$ 
& \cellcolor{gray!8}\boldmath{$\downarrow 11.34$} 
& $20.65\textcolor{gray}{\text{\scriptsize±0.06}}$ 
& $19.47\textcolor{gray}{\text{\scriptsize±0.50}}$ 
& \cellcolor{gray!8}\boldmath{$\downarrow 5.71$} 
& $21.80\textcolor{gray}{\text{\scriptsize±0.56}}$ 
& $19.66\textcolor{gray}{\text{\scriptsize±0.67}}$ 
& \cellcolor{gray!8}\boldmath{$\downarrow 9.82$} 
& $20.97\textcolor{gray}{\text{\scriptsize±0.36}}$ 
& $\secondres{{17.55}}\textcolor{gray}{\text{\scriptsize±0.26}}$ 
& \cellcolor{gray!8}\boldmath{$\downarrow 16.31$} \\

& MAPE(\%) 
& $62.14\textcolor{gray}{\text{\scriptsize±0.54}}$ 
& $\firstres{{42.62}}\textcolor{gray}{\text{\scriptsize±0.52}}$ 
& \cellcolor{gray!8}\boldmath{$\downarrow 31.41$} 
& $60.03\textcolor{gray}{\text{\scriptsize±4.32}}$ 
& $58.83\textcolor{gray}{\text{\scriptsize±2.95}}$ 
& \cellcolor{gray!8}\boldmath{$\downarrow 2.00$} 
& $71.16\textcolor{gray}{\text{\scriptsize±1.37}}$ 
& $52.19\textcolor{gray}{\text{\scriptsize±0.62}}$ 
& \cellcolor{gray!8}\boldmath{$\downarrow 26.66$} 
& $69.08\textcolor{gray}{\text{\scriptsize±1.30}}$ 
& $49.74\textcolor{gray}{\text{\scriptsize±2.81}}$ 
& \cellcolor{gray!8}\boldmath{$\downarrow 27.99$} 
& $75.40\textcolor{gray}{\text{\scriptsize±9.03}}$ 
& $54.59\textcolor{gray}{\text{\scriptsize±6.32}}$ 
& \cellcolor{gray!8}\boldmath{$\downarrow 27.61$} 
& $70.18\textcolor{gray}{\text{\scriptsize±6.64}}$ 
& $\secondres{{43.76}}\textcolor{gray}{\text{\scriptsize±0.03}}$ 
& \cellcolor{gray!8}\boldmath{$\downarrow 37.65$} \\

& MRE(\%) 
& $53.33\textcolor{gray}{\text{\scriptsize±1.79}}$ 
& $\firstres{{38.84}}\textcolor{gray}{\text{\scriptsize±0.75}}$ 
& \cellcolor{gray!8}\boldmath{$\downarrow 27.17$} 
& $52.24\textcolor{gray}{\text{\scriptsize±1.39}}$ 
& $49.36\textcolor{gray}{\text{\scriptsize±0.82}}$ 
& \cellcolor{gray!8}\boldmath{$\downarrow 5.51$} 
& $55.63\textcolor{gray}{\text{\scriptsize±0.78}}$ 
& $46.97\textcolor{gray}{\text{\scriptsize±4.17}}$ 
& \cellcolor{gray!8}\boldmath{$\downarrow 15.58$} 
& $53.77\textcolor{gray}{\text{\scriptsize±0.24}}$ 
& $48.68\textcolor{gray}{\text{\scriptsize±2.11}}$ 
& \cellcolor{gray!8}\boldmath{$\downarrow 9.47$} 
& $57.02\textcolor{gray}{\text{\scriptsize±2.55}}$ 
& $49.45\textcolor{gray}{\text{\scriptsize±2.46}}$ 
& \cellcolor{gray!8}\boldmath{$\downarrow 13.28$} 
& $55.27\textcolor{gray}{\text{\scriptsize±1.22}}$ 
& $\secondres{{43.76}}\textcolor{gray}{\text{\scriptsize±0.03}}$ 
& \cellcolor{gray!8}\boldmath{$\downarrow 20.82$} \\

  \midrule
\multirow{4}{*}{\SpeedNineteen$*$}
& MAE
& $3.93\textcolor{gray}{\text{\scriptsize±0.01}}$ 
& $\secondres{{3.80}}\textcolor{gray}{\text{\scriptsize±0.01}}$ 
& \cellcolor{gray!8}\boldmath{$\downarrow 3.31$}
& $3.96\textcolor{gray}{\text{\scriptsize±0.06}}$ 
& $\firstres{{3.77}}\textcolor{gray}{\text{\scriptsize±0.06}}$ 
& \cellcolor{gray!8}\boldmath{$\downarrow 4.80$}
& $22.33\textcolor{gray}{\text{\scriptsize±3.90}}$ 
& $4.28\textcolor{gray}{\text{\scriptsize±0.34}}$ 
& \cellcolor{gray!8}\boldmath{$\downarrow 80.83$}
& $38.45\textcolor{gray}{\text{\scriptsize±0.02}}$ 
& $4.52\textcolor{gray}{\text{\scriptsize±0.62}}$ 
& \cellcolor{gray!8}\boldmath{$\downarrow 88.27$}
& $6.27\textcolor{gray}{\text{\scriptsize±0.53}}$ 
& $4.51\textcolor{gray}{\text{\scriptsize±0.41}}$ 
& \cellcolor{gray!8}\boldmath{$\downarrow 28.07$}
& $4.73\textcolor{gray}{\text{\scriptsize±0.06}}$ 
& $3.96\textcolor{gray}{\text{\scriptsize±0.18}}$ 
& \cellcolor{gray!8}\boldmath{$\downarrow 16.28$} \\

& RMSE
& $31.82\textcolor{gray}{\text{\scriptsize±0.02}}$ 
& $31.78\textcolor{gray}{\text{\scriptsize±0.02}}$ 
& \cellcolor{gray!8}\boldmath{$\downarrow 0.13$}
& $\firstres{{31.38}}\textcolor{gray}{\text{\scriptsize±0.02}}$ 
& $\secondres{{31.66}}\textcolor{gray}{\text{\scriptsize±0.02}}$ 
& \cellcolor{gray!8}\boldmath{$\uparrow 0.88$}
& $45.19\textcolor{gray}{\text{\scriptsize±3.38}}$ 
& $32.18\textcolor{gray}{\text{\scriptsize±0.34}}$ 
& \cellcolor{gray!8}\boldmath{$\downarrow 28.79$}
& $139.01\textcolor{gray}{\text{\scriptsize±0.01}}$ 
& $32.26\textcolor{gray}{\text{\scriptsize±0.55}}$ 
& \cellcolor{gray!8}\boldmath{$\downarrow 76.78$}
& $43.66\textcolor{gray}{\text{\scriptsize±0.77}}$ 
& $33.12\textcolor{gray}{\text{\scriptsize±0.96}}$ 
& \cellcolor{gray!8}\boldmath{$\downarrow 31.82$}
& $33.64\textcolor{gray}{\text{\scriptsize±0.41}}$ 
& $31.92\textcolor{gray}{\text{\scriptsize±0.25}}$ 
& \cellcolor{gray!8}\boldmath{$\downarrow 5.11$} \\

& MAPE(\%)
& $9.43\textcolor{gray}{\text{\scriptsize±0.02}}$ 
& $\secondres{{9.32}}\textcolor{gray}{\text{\scriptsize±0.03}}$ 
& \cellcolor{gray!8}\boldmath{$\downarrow 1.17$}
& $\firstres{{9.05}}\textcolor{gray}{\text{\scriptsize±0.23}}$ 
& $9.45\textcolor{gray}{\text{\scriptsize±0.22}}$ 
& \cellcolor{gray!8}\boldmath{$\uparrow 4.76$}
& $13.82\textcolor{gray}{\text{\scriptsize±0.29}}$ 
& $10.03\textcolor{gray}{\text{\scriptsize±0.35}}$ 
& \cellcolor{gray!8}\boldmath{$\downarrow 27.42$}
& $99.85\textcolor{gray}{\text{\scriptsize±0.01}}$ 
& $10.44\textcolor{gray}{\text{\scriptsize±0.70}}$ 
& \cellcolor{gray!8}\boldmath{$\downarrow 89.55$}
& $13.59\textcolor{gray}{\text{\scriptsize±0.58}}$ 
& $10.94\textcolor{gray}{\text{\scriptsize±0.91}}$ 
& \cellcolor{gray!8}\boldmath{$\downarrow 19.49$}
& $11.41\textcolor{gray}{\text{\scriptsize±0.14}}$ 
& $9.58\textcolor{gray}{\text{\scriptsize±0.30}}$ 
& \cellcolor{gray!8}\boldmath{$\downarrow 16.04$} \\

& MRE(\%)
& $10.21\textcolor{gray}{\text{\scriptsize±0.02}}$ 
& $\secondres{{9.89}}\textcolor{gray}{\text{\scriptsize±0.03}}$ 
& \cellcolor{gray!8}\boldmath{$\downarrow 3.13$}
& $10.29\textcolor{gray}{\text{\scriptsize±0.46}}$ 
& $\firstres{{9.83}}\textcolor{gray}{\text{\scriptsize±0.15}}$ 
& \cellcolor{gray!8}\boldmath{$\downarrow 4.68$}
& $58.11\textcolor{gray}{\text{\scriptsize±10.16}}$ 
& $11.13\textcolor{gray}{\text{\scriptsize±0.88}}$ 
& \cellcolor{gray!8}\boldmath{$\downarrow 80.84$}
& $100.08\textcolor{gray}{\text{\scriptsize±0.05}}$ 
& $11.77\textcolor{gray}{\text{\scriptsize±1.61}}$ 
& \cellcolor{gray!8}\boldmath{$\downarrow 88.24$}
& $16.09\textcolor{gray}{\text{\scriptsize±1.11}}$ 
& $11.75\textcolor{gray}{\text{\scriptsize±1.05}}$ 
& \cellcolor{gray!8}\boldmath{$\downarrow 26.97$}
& $12.32\textcolor{gray}{\text{\scriptsize±0.16}}$ 
& $10.29\textcolor{gray}{\text{\scriptsize±0.46}}$ 
& \cellcolor{gray!8}\boldmath{$\downarrow 16.48$} \\

  \midrule
  \multirow{4}{*}{\AirTwenty} 
& MAE 
& $9.91\textcolor{gray}{\text{\scriptsize±0.05}}$ 
& $\secondres{{7.61}}\textcolor{gray}{\text{\scriptsize±0.23}}$ 
& \cellcolor{gray!8}\boldmath{$\downarrow 23.21$}
& $9.39\textcolor{gray}{\text{\scriptsize±0.20}}$ 
& $8.66\textcolor{gray}{\text{\scriptsize±0.17}}$ 
& \cellcolor{gray!8}\boldmath{$\downarrow 7.78$}
& $10.10\textcolor{gray}{\text{\scriptsize±0.03}}$ 
& $8.49\textcolor{gray}{\text{\scriptsize±0.17}}$ 
& \cellcolor{gray!8}\boldmath{$\downarrow 15.94$}
& $9.71\textcolor{gray}{\text{\scriptsize±0.19}}$ 
& $8.10\textcolor{gray}{\text{\scriptsize±0.07}}$ 
& \cellcolor{gray!8}\boldmath{$\downarrow 16.57$}
& $10.00\textcolor{gray}{\text{\scriptsize±0.76}}$ 
& $8.71\textcolor{gray}{\text{\scriptsize±0.14}}$ 
& \cellcolor{gray!8}\boldmath{$\downarrow 12.90$}
& $9.93\textcolor{gray}{\text{\scriptsize±0.85}}$ 
& $\firstres{{7.35}}\textcolor{gray}{\text{\scriptsize±0.21}}$ 
& \cellcolor{gray!8}\boldmath{$\downarrow 25.97$} \\

& RMSE
& $15.08\textcolor{gray}{\text{\scriptsize±0.12}}$ 
& $\secondres{{11.52}}\textcolor{gray}{\text{\scriptsize±0.49}}$ 
& \cellcolor{gray!8}\boldmath{$\downarrow 23.61$}
& $14.51\textcolor{gray}{\text{\scriptsize±0.34}}$ 
& $13.26\textcolor{gray}{\text{\scriptsize±0.04}}$ 
& \cellcolor{gray!8}\boldmath{$\downarrow 8.60$}
& $14.93\textcolor{gray}{\text{\scriptsize±0.12}}$ 
& $12.76\textcolor{gray}{\text{\scriptsize±0.18}}$ 
& \cellcolor{gray!8}\boldmath{$\downarrow 14.52$}
& $14.84\textcolor{gray}{\text{\scriptsize±0.19}}$ 
& $13.07\textcolor{gray}{\text{\scriptsize±0.23}}$ 
& \cellcolor{gray!8}\boldmath{$\downarrow 11.91$}
& $14.98\textcolor{gray}{\text{\scriptsize±0.65}}$ 
& $13.77\textcolor{gray}{\text{\scriptsize±0.21}}$ 
& \cellcolor{gray!8}\boldmath{$\downarrow 8.08$}
& $14.85\textcolor{gray}{\text{\scriptsize±0.89}}$ 
& $\firstres{{11.47}}\textcolor{gray}{\text{\scriptsize±0.39}}$ 
& \cellcolor{gray!8}\boldmath{$\downarrow 22.79$} \\

& MAPE(\%)
& $54.78\textcolor{gray}{\text{\scriptsize±1.05}}$ 
& $\firstres{{46.59}}\textcolor{gray}{\text{\scriptsize±0.64}}$ 
& \cellcolor{gray!8}\boldmath{$\downarrow 14.95$}
& $54.79\textcolor{gray}{\text{\scriptsize±0.59}}$ 
& $56.45\textcolor{gray}{\text{\scriptsize±2.25}}$ 
& \cellcolor{gray!8}\boldmath{$\uparrow 3.03$}
& $66.35\textcolor{gray}{\text{\scriptsize±1.23}}$ 
& $55.59\textcolor{gray}{\text{\scriptsize±2.78}}$ 
& \cellcolor{gray!8}\boldmath{$\downarrow 16.19$}
& $67.26\textcolor{gray}{\text{\scriptsize±2.25}}$ 
& $48.14\textcolor{gray}{\text{\scriptsize±1.33}}$ 
& \cellcolor{gray!8}\boldmath{$\downarrow 28.40$}
& $68.35\textcolor{gray}{\text{\scriptsize±11.56}}$ 
& $54.35\textcolor{gray}{\text{\scriptsize±1.09}}$ 
& \cellcolor{gray!8}\boldmath{$\downarrow 20.43$}
& $66.35\textcolor{gray}{\text{\scriptsize±8.27}}$ 
& $\secondres{{47.63}}\textcolor{gray}{\text{\scriptsize±3.59}}$ 
& \cellcolor{gray!8}\boldmath{$\downarrow 28.21$} \\

& MRE(\%)
& $53.01\textcolor{gray}{\text{\scriptsize±0.29}}$ 
& $\secondres{{40.74}}\textcolor{gray}{\text{\scriptsize±1.24}}$ 
& \cellcolor{gray!8}\boldmath{$\downarrow 23.15$}
& $50.25\textcolor{gray}{\text{\scriptsize±1.09}}$ 
& $46.35\textcolor{gray}{\text{\scriptsize±0.92}}$ 
& \cellcolor{gray!8}\boldmath{$\downarrow 7.76$}
& $54.05\textcolor{gray}{\text{\scriptsize±0.16}}$ 
& $45.41\textcolor{gray}{\text{\scriptsize±0.93}}$ 
& \cellcolor{gray!8}\boldmath{$\downarrow 15.97$}
& $51.96\textcolor{gray}{\text{\scriptsize±1.02}}$ 
& $43.35\textcolor{gray}{\text{\scriptsize±0.36}}$ 
& \cellcolor{gray!8}\boldmath{$\downarrow 16.56$}
& $53.51\textcolor{gray}{\text{\scriptsize±4.08}}$ 
& $46.59\textcolor{gray}{\text{\scriptsize±0.76}}$ 
& \cellcolor{gray!8}\boldmath{$\downarrow 12.92$}
& $53.11\textcolor{gray}{\text{\scriptsize±4.55}}$ 
& $\firstres{{39.35}}\textcolor{gray}{\text{\scriptsize±1.11}}$ 
& \cellcolor{gray!8}\boldmath{$\downarrow 25.94$} \\

  \midrule
  \multirow{4}{*}{\TrafficTwenty} 
& MAE 
& $62.66\textcolor{gray}{\text{\scriptsize±0.59}}$ 
& $\secondres{{50.59}}\textcolor{gray}{\text{\scriptsize±0.04}}$ 
& \cellcolor{gray!8}\boldmath{$\downarrow 19.26$}
& $62.65\textcolor{gray}{\text{\scriptsize±1.19}}$ 
& $51.09\textcolor{gray}{\text{\scriptsize±4.57}}$ 
& \cellcolor{gray!8}\boldmath{$\downarrow 18.53$}
& $61.34\textcolor{gray}{\text{\scriptsize±0.83}}$ 
& $62.36\textcolor{gray}{\text{\scriptsize±1.83}}$ 
& \cellcolor{gray!8}\boldmath{$\uparrow 1.66$}
& $67.02\textcolor{gray}{\text{\scriptsize±1.78}}$ 
& $62.31\textcolor{gray}{\text{\scriptsize±3.14}}$ 
& \cellcolor{gray!8}\boldmath{$\downarrow 7.04$}
& $68.04\textcolor{gray}{\text{\scriptsize±0.65}}$ 
& $60.90\textcolor{gray}{\text{\scriptsize±1.19}}$ 
& \cellcolor{gray!8}\boldmath{$\downarrow 10.49$}
& $64.29\textcolor{gray}{\text{\scriptsize±1.37}}$ 
& $\firstres{{50.53}}\textcolor{gray}{\text{\scriptsize±3.94}}$ 
& \cellcolor{gray!8}\boldmath{$\downarrow 21.39$} \\

& RMSE
& $113.21\textcolor{gray}{\text{\scriptsize±1.13}}$ 
& $\firstres{{89.10}}\textcolor{gray}{\text{\scriptsize±7.36}}$ 
& \cellcolor{gray!8}\boldmath{$\downarrow 21.30$}
& $113.12\textcolor{gray}{\text{\scriptsize±1.07}}$ 
& $91.20\textcolor{gray}{\text{\scriptsize±7.36}}$ 
& \cellcolor{gray!8}\boldmath{$\downarrow 19.38$}
& $111.50\textcolor{gray}{\text{\scriptsize±0.40}}$ 
& $102.84\textcolor{gray}{\text{\scriptsize±2.01}}$ 
& \cellcolor{gray!8}\boldmath{$\downarrow 7.78$}
& $117.62\textcolor{gray}{\text{\scriptsize±2.98}}$ 
& $104.07\textcolor{gray}{\text{\scriptsize±6.46}}$ 
& \cellcolor{gray!8}\boldmath{$\downarrow 11.53$}
& $116.81\textcolor{gray}{\text{\scriptsize±0.13}}$ 
& $102.06\textcolor{gray}{\text{\scriptsize±2.49}}$ 
& \cellcolor{gray!8}\boldmath{$\downarrow 12.63$}
& $114.56\textcolor{gray}{\text{\scriptsize±2.74}}$ 
& $\secondres{{90.24}}\textcolor{gray}{\text{\scriptsize±6.35}}$ 
& \cellcolor{gray!8}\boldmath{$\downarrow 21.23$} \\

& MAPE(\%)
& $45.87\textcolor{gray}{\text{\scriptsize±0.60}}$ 
& $26.68\textcolor{gray}{\text{\scriptsize±1.69}}$ 
& \cellcolor{gray!8}\boldmath{$\downarrow 41.84$}
& $30.77\textcolor{gray}{\text{\scriptsize±0.48}}$ 
& $\secondres{{26.62}}\textcolor{gray}{\text{\scriptsize±3.11}}$ 
& \cellcolor{gray!8}\boldmath{$\downarrow 13.50$}
& $33.47\textcolor{gray}{\text{\scriptsize±0.86}}$ 
& $35.00\textcolor{gray}{\text{\scriptsize±3.14}}$ 
& \cellcolor{gray!8}\boldmath{$\uparrow 4.56$}
& $38.23\textcolor{gray}{\text{\scriptsize±0.99}}$ 
& $35.81\textcolor{gray}{\text{\scriptsize±1.34}}$ 
& \cellcolor{gray!8}\boldmath{$\downarrow 6.33$}
& $36.08\textcolor{gray}{\text{\scriptsize±2.07}}$ 
& $32.27\textcolor{gray}{\text{\scriptsize±1.58}}$ 
& \cellcolor{gray!8}\boldmath{$\downarrow 10.54$}
& $32.31\textcolor{gray}{\text{\scriptsize±1.36}}$ 
& $\firstres{{26.11}}\textcolor{gray}{\text{\scriptsize±2.67}}$ 
& \cellcolor{gray!8}\boldmath{$\downarrow 19.18$} \\

& MRE(\%)
& $22.68\textcolor{gray}{\text{\scriptsize±0.21}}$ 
& $\secondres{{18.31}}\textcolor{gray}{\text{\scriptsize±1.14}}$ 
& \cellcolor{gray!8}\boldmath{$\downarrow 19.27$}
& $22.68\textcolor{gray}{\text{\scriptsize±0.43}}$ 
& $18.49\textcolor{gray}{\text{\scriptsize±1.66}}$ 
& \cellcolor{gray!8}\boldmath{$\downarrow 18.48$}
& $22.20\textcolor{gray}{\text{\scriptsize±0.30}}$ 
& $22.57\textcolor{gray}{\text{\scriptsize±1.62}}$ 
& \cellcolor{gray!8}\boldmath{$\uparrow 1.67$}
& $24.26\textcolor{gray}{\text{\scriptsize±0.64}}$ 
& $22.55\textcolor{gray}{\text{\scriptsize±1.14}}$ 
& \cellcolor{gray!8}\boldmath{$\downarrow 7.04$}
& $24.63\textcolor{gray}{\text{\scriptsize±0.24}}$ 
& $22.04\textcolor{gray}{\text{\scriptsize±0.43}}$ 
& \cellcolor{gray!8}\boldmath{$\downarrow 10.51$}
& $23.27\textcolor{gray}{\text{\scriptsize±0.49}}$ 
& $\firstres{{18.29}}\textcolor{gray}{\text{\scriptsize±1.43}}$ 
& \cellcolor{gray!8}\boldmath{$\downarrow 21.41$}\\

    \bottomrule
  \end{tabular}
  \end{sc}
  }
  \label{tab:rq1_2day}
\end{table*}

\begin{table*}[htbp!]
  \setlength{\tabcolsep}{4.0pt}
  \centering
  \caption{Performance comparison of different models w/ and w/o \model on common benchmarks (1-day scenario).}
  \resizebox{\linewidth}{!}{%
  \begin{sc}
  \renewcommand{\arraystretch}{1.8}
  \begin{tabular}{lccccccccccccccccccc}
    \toprule
    & Model & \multicolumn{3}{c}{{AGCRN}}
      & \multicolumn{3}{c}{{GWNet}}
      & \multicolumn{3}{c}{{GGNN}}
      & \multicolumn{3}{c}{{GRUGCN}}
      & \multicolumn{3}{c}{{STGCN}}
      & \multicolumn{3}{c}{{DCRNN}}\\
    \multicolumn{2}{c}{w/ \model} & \rxmark & \gcmark & {$\Delta$(\%)}
     & \rxmark & \gcmark & {$\Delta$(\%)}
     & \rxmark & \gcmark & {$\Delta$(\%)}
     & \rxmark & \gcmark & {$\Delta$(\%)}
     & \rxmark & \gcmark & {$\Delta$(\%)}
     & \rxmark & \gcmark & {$\Delta$(\%)}\\
    \cmidrule(lr){1-2}\cmidrule(lr){3-5}\cmidrule(lr){6-8}\cmidrule(lr){9-11}\cmidrule(lr){12-14}\cmidrule(lr){15-17}\cmidrule(lr){18-20}
 \multirow{4}{*}{\AirNineteen} 
& MAE 
& $13.63\textcolor{gray}{\text{\scriptsize±0.45}}$ 
& $\firstres{{9.33}}\textcolor{gray}{\text{\scriptsize±0.13}}$ 
& \cellcolor{gray!8}\boldmath{$\downarrow 31.55$}
& $10.92\textcolor{gray}{\text{\scriptsize±2.01}}$ 
& $11.57\textcolor{gray}{\text{\scriptsize±0.32}}$ 
& \cellcolor{gray!8}\boldmath{$\uparrow 5.95$}
& $11.57\textcolor{gray}{\text{\scriptsize±0.37}}$ 
& $10.78\textcolor{gray}{\text{\scriptsize±0.48}}$ 
& \cellcolor{gray!8}\boldmath{$\downarrow 6.83$}
& $11.63\textcolor{gray}{\text{\scriptsize±0.08}}$ 
& $11.65\textcolor{gray}{\text{\scriptsize±0.39}}$ 
& \cellcolor{gray!8}\boldmath{$\uparrow 0.17$}
& $13.60\textcolor{gray}{\text{\scriptsize±1.30}}$ 
& $11.55\textcolor{gray}{\text{\scriptsize±0.21}}$ 
& \cellcolor{gray!8}\boldmath{$\downarrow 15.07$}
& $13.80\textcolor{gray}{\text{\scriptsize±0.46}}$ 
& $\secondres{{10.25}}\textcolor{gray}{\text{\scriptsize±0.19}}$ 
& \cellcolor{gray!8}\boldmath{$\downarrow 25.72$} \\

& RMSE
& $20.43\textcolor{gray}{\text{\scriptsize±0.89}}$ 
& $\firstres{{14.24}}\textcolor{gray}{\text{\scriptsize±0.18}}$ 
& \cellcolor{gray!8}\boldmath{$\downarrow 30.30$}
& $17.17\textcolor{gray}{\text{\scriptsize±3.31}}$ 
& $17.79\textcolor{gray}{\text{\scriptsize±0.37}}$ 
& \cellcolor{gray!8}\boldmath{$\uparrow 3.61$}
& $18.25\textcolor{gray}{\text{\scriptsize±0.20}}$ 
& $16.87\textcolor{gray}{\text{\scriptsize±0.89}}$ 
& \cellcolor{gray!8}\boldmath{$\downarrow 7.56$}
& $18.44\textcolor{gray}{\text{\scriptsize±0.04}}$ 
& $18.43\textcolor{gray}{\text{\scriptsize±0.38}}$ 
& \cellcolor{gray!8}\boldmath{$\downarrow 0.05$}
& $20.56\textcolor{gray}{\text{\scriptsize±1.21}}$ 
& $18.30\textcolor{gray}{\text{\scriptsize±0.29}}$ 
& \cellcolor{gray!8}\boldmath{$\downarrow 10.99$}
& $21.07\textcolor{gray}{\text{\scriptsize±0.43}}$ 
& $\secondres{{16.28}}\textcolor{gray}{\text{\scriptsize±0.06}}$ 
& \cellcolor{gray!8}\boldmath{$\downarrow 22.73$} \\

& MAPE(\%)
& $60.71\textcolor{gray}{\text{\scriptsize±0.41}}$ 
& $\firstres{{38.57}}\textcolor{gray}{\text{\scriptsize±0.32}}$ 
& \cellcolor{gray!8}\boldmath{$\downarrow 36.80$}
& $51.54\textcolor{gray}{\text{\scriptsize±2.00}}$ 
& $51.92\textcolor{gray}{\text{\scriptsize±2.89}}$ 
& \cellcolor{gray!8}\boldmath{$\uparrow 0.74$}
& $56.99\textcolor{gray}{\text{\scriptsize±1.81}}$ 
& $45.09\textcolor{gray}{\text{\scriptsize±5.46}}$ 
& \cellcolor{gray!8}\boldmath{$\downarrow 20.88$}
& $56.54\textcolor{gray}{\text{\scriptsize±1.34}}$ 
& $44.86\textcolor{gray}{\text{\scriptsize±2.08}}$ 
& \cellcolor{gray!8}\boldmath{$\downarrow 20.78$}
& $68.55\textcolor{gray}{\text{\scriptsize±12.26}}$ 
& $48.33\textcolor{gray}{\text{\scriptsize±3.28}}$ 
& \cellcolor{gray!8}\boldmath{$\downarrow 29.50$}
& $68.45\textcolor{gray}{\text{\scriptsize±7.25}}$ 
& $\secondres{{44.68}}\textcolor{gray}{\text{\scriptsize±0.81}}$ 
& \cellcolor{gray!8}\boldmath{$\downarrow 34.72$} \\

& MRE(\%)
& $52.74\textcolor{gray}{\text{\scriptsize±1.81}}$ 
& $\firstres{{35.85}}\textcolor{gray}{\text{\scriptsize±0.71}}$ 
& \cellcolor{gray!8}\boldmath{$\downarrow 32.03$}
& $46.47\textcolor{gray}{\text{\scriptsize±0.65}}$ 
& $44.73\textcolor{gray}{\text{\scriptsize±1.23}}$ 
& \cellcolor{gray!8}\boldmath{$\downarrow 3.74$}
& $45.17\textcolor{gray}{\text{\scriptsize±0.99}}$ 
& $41.67\textcolor{gray}{\text{\scriptsize±1.67}}$ 
& \cellcolor{gray!8}\boldmath{$\downarrow 7.75$}
& $44.96\textcolor{gray}{\text{\scriptsize±0.32}}$ 
& $45.06\textcolor{gray}{\text{\scriptsize±1.52}}$ 
& \cellcolor{gray!8}\boldmath{$\uparrow 0.22$}
& $52.59\textcolor{gray}{\text{\scriptsize±5.03}}$ 
& $44.68\textcolor{gray}{\text{\scriptsize±0.81}}$ 
& \cellcolor{gray!8}\boldmath{$\downarrow 15.04$}
& $53.37\textcolor{gray}{\text{\scriptsize±1.80}}$ 
& $\secondres{{39.63}}\textcolor{gray}{\text{\scriptsize±0.74}}$ 
& \cellcolor{gray!8}\boldmath{$\downarrow 25.75$} \\

 \midrule
\multirow{4}{*}{\SpeedNineteen$*$}
& MAE
& $3.92\textcolor{gray}{\text{\scriptsize±0.01}}$ 
& $\secondres{{3.79}}\textcolor{gray}{\text{\scriptsize±0.01}}$ 
& \cellcolor{gray!8}\boldmath{$\downarrow 3.32$}
& $3.81\textcolor{gray}{\text{\scriptsize±0.13}}$ 
& $\firstres{{3.70}}\textcolor{gray}{\text{\scriptsize±0.06}}$ 
& \cellcolor{gray!8}\boldmath{$\downarrow 2.89$}
& $21.81\textcolor{gray}{\text{\scriptsize±3.90}}$ 
& $4.03\textcolor{gray}{\text{\scriptsize±0.26}}$ 
& \cellcolor{gray!8}\boldmath{$\downarrow 81.52$}
& $38.45\textcolor{gray}{\text{\scriptsize±0.02}}$ 
& $4.39\textcolor{gray}{\text{\scriptsize±0.68}}$ 
& \cellcolor{gray!8}\boldmath{$\downarrow 88.60$}
& $5.53\textcolor{gray}{\text{\scriptsize±0.40}}$ 
& $4.34\textcolor{gray}{\text{\scriptsize±0.36}}$ 
& \cellcolor{gray!8}\boldmath{$\downarrow 21.52$}
& $4.52\textcolor{gray}{\text{\scriptsize±0.05}}$ 
& $3.81\textcolor{gray}{\text{\scriptsize±0.13}}$ 
& \cellcolor{gray!8}\boldmath{$\downarrow 15.71$} \\

& RMSE
& $31.78\textcolor{gray}{\text{\scriptsize±0.02}}$ 
& $31.73\textcolor{gray}{\text{\scriptsize±0.02}}$ 
& \cellcolor{gray!8}\boldmath{$\downarrow 0.16$}
& $\firstres{{30.92}}\textcolor{gray}{\text{\scriptsize±0.04}}$ 
& $31.42\textcolor{gray}{\text{\scriptsize±0.04}}$ 
& \cellcolor{gray!8}\boldmath{$\uparrow 1.61$}
& $43.31\textcolor{gray}{\text{\scriptsize±3.43}}$ 
& $\secondres{{31.08}}\textcolor{gray}{\text{\scriptsize±0.12}}$ 
& \cellcolor{gray!8}\boldmath{$\downarrow 28.30$}
& $139.01\textcolor{gray}{\text{\scriptsize±0.01}}$ 
& $31.76\textcolor{gray}{\text{\scriptsize±0.45}}$ 
& \cellcolor{gray!8}\boldmath{$\downarrow 77.15$}
& $38.60\textcolor{gray}{\text{\scriptsize±0.65}}$ 
& $32.19\textcolor{gray}{\text{\scriptsize±0.60}}$ 
& \cellcolor{gray!8}\boldmath{$\downarrow 16.61$}
& $32.47\textcolor{gray}{\text{\scriptsize±0.12}}$ 
& $31.11\textcolor{gray}{\text{\scriptsize±0.09}}$ 
& \cellcolor{gray!8}\boldmath{$\downarrow 4.19$} \\

& MAPE(\%)
& $9.42\textcolor{gray}{\text{\scriptsize±0.00}}$ 
& $9.30\textcolor{gray}{\text{\scriptsize±0.03}}$ 
& \cellcolor{gray!8}\boldmath{$\downarrow 1.27$}
& $\firstres{{8.59}}\textcolor{gray}{\text{\scriptsize±0.21}}$ 
& $9.27\textcolor{gray}{\text{\scriptsize±0.21}}$ 
& \cellcolor{gray!8}\boldmath{$\uparrow 7.33$}
& $12.86\textcolor{gray}{\text{\scriptsize±0.31}}$ 
& $9.42\textcolor{gray}{\text{\scriptsize±0.18}}$ 
& \cellcolor{gray!8}\boldmath{$\downarrow 26.75$}
& $99.84\textcolor{gray}{\text{\scriptsize±0.01}}$ 
& $10.12\textcolor{gray}{\text{\scriptsize±0.50}}$ 
& \cellcolor{gray!8}\boldmath{$\downarrow 89.87$}
& $12.19\textcolor{gray}{\text{\scriptsize±0.57}}$ 
& $10.55\textcolor{gray}{\text{\scriptsize±0.81}}$ 
& \cellcolor{gray!8}\boldmath{$\downarrow 13.45$}
& $10.95\textcolor{gray}{\text{\scriptsize±0.13}}$ 
& $\secondres{{9.25}}\textcolor{gray}{\text{\scriptsize±0.20}}$ 
& \cellcolor{gray!8}\boldmath{$\downarrow 15.52$} \\

& MRE(\%)
& $10.20\textcolor{gray}{\text{\scriptsize±0.02}}$ 
& $\secondres{{9.88}}\textcolor{gray}{\text{\scriptsize±0.03}}$ 
& \cellcolor{gray!8}\boldmath{$\downarrow 3.14$}
& $9.91\textcolor{gray}{\text{\scriptsize±0.16}}$ 
& $\firstres{{9.64}}\textcolor{gray}{\text{\scriptsize±0.16}}$ 
& \cellcolor{gray!8}\boldmath{$\downarrow 2.72$}
& $56.77\textcolor{gray}{\text{\scriptsize±10.14}}$ 
& $10.48\textcolor{gray}{\text{\scriptsize±0.69}}$ 
& \cellcolor{gray!8}\boldmath{$\downarrow 81.52$}
& $100.07\textcolor{gray}{\text{\scriptsize±0.05}}$ 
& $11.42\textcolor{gray}{\text{\scriptsize±1.78}}$ 
& \cellcolor{gray!8}\boldmath{$\downarrow 87.59$}
& $14.38\textcolor{gray}{\text{\scriptsize±1.03}}$ 
& $11.30\textcolor{gray}{\text{\scriptsize±0.92}}$ 
& \cellcolor{gray!8}\boldmath{$\downarrow 21.42$}
& $11.76\textcolor{gray}{\text{\scriptsize±0.12}}$ 
& $9.91\textcolor{gray}{\text{\scriptsize±0.34}}$ 
& \cellcolor{gray!8}\boldmath{$\downarrow 15.73$} \\

 \midrule
 \multirow{4}{*}{\AirTwenty} 
 & MAE 
& $9.55\textcolor{gray}{\text{\scriptsize±0.24}}$ 
& $\secondres{{7.33}}\textcolor{gray}{\text{\scriptsize±0.21}}$ 
& \cellcolor{gray!8}\boldmath{$\downarrow 23.25$} 
& $8.58\textcolor{gray}{\text{\scriptsize±0.41}}$ 
& $7.98\textcolor{gray}{\text{\scriptsize±0.19}}$ 
& \cellcolor{gray!8}\boldmath{$\downarrow 6.99$}
& $8.60\textcolor{gray}{\text{\scriptsize±0.01}}$ 
& $8.15\textcolor{gray}{\text{\scriptsize±0.11}}$ 
& \cellcolor{gray!8}\boldmath{$\downarrow 5.23$}
& $8.13\textcolor{gray}{\text{\scriptsize±0.11}}$ 
& $7.55\textcolor{gray}{\text{\scriptsize±0.09}}$ 
& \cellcolor{gray!8}\boldmath{$\downarrow 7.13$}
& $9.17\textcolor{gray}{\text{\scriptsize±0.88}}$ 
& $7.99\textcolor{gray}{\text{\scriptsize±0.14}}$ 
& \cellcolor{gray!8}\boldmath{$\downarrow 12.87$}
& $9.11\textcolor{gray}{\text{\scriptsize±1.32}}$ 
& $\firstres{{6.81}}\textcolor{gray}{\text{\scriptsize±0.30}}$ 
& \cellcolor{gray!8}\boldmath{$\downarrow 25.25$} \\

& RMSE
& $14.58\textcolor{gray}{\text{\scriptsize±0.48}}$ 
& $\secondres{{11.14}}\textcolor{gray}{\text{\scriptsize±0.21}}$ 
& \cellcolor{gray!8}\boldmath{$\downarrow 23.59$} 
& $13.44\textcolor{gray}{\text{\scriptsize±0.62}}$ 
& $12.26\textcolor{gray}{\text{\scriptsize±0.10}}$ 
& \cellcolor{gray!8}\boldmath{$\downarrow 8.78$}
& $13.20\textcolor{gray}{\text{\scriptsize±0.12}}$ 
& $12.43\textcolor{gray}{\text{\scriptsize±0.25}}$ 
& \cellcolor{gray!8}\boldmath{$\downarrow 5.83$}
& $12.84\textcolor{gray}{\text{\scriptsize±0.08}}$ 
& $12.17\textcolor{gray}{\text{\scriptsize±0.29}}$ 
& \cellcolor{gray!8}\boldmath{$\downarrow 5.22$}
& $13.86\textcolor{gray}{\text{\scriptsize±0.77}}$ 
& $12.65\textcolor{gray}{\text{\scriptsize±0.12}}$ 
& \cellcolor{gray!8}\boldmath{$\downarrow 8.73$}
& $13.97\textcolor{gray}{\text{\scriptsize±1.56}}$ 
& $\firstres{{10.72}}\textcolor{gray}{\text{\scriptsize±0.58}}$ 
& \cellcolor{gray!8}\boldmath{$\downarrow 23.26$} \\

& MAPE(\%)
& $\firstres{{31.80}}\textcolor{gray}{\text{\scriptsize±0.70}}$ 
& $44.90\textcolor{gray}{\text{\scriptsize±0.88}}$ 
& \cellcolor{gray!8}\boldmath{$\uparrow 41.19$} 
& $49.43\textcolor{gray}{\text{\scriptsize±1.83}}$ 
& $51.38\textcolor{gray}{\text{\scriptsize±1.95}}$ 
& \cellcolor{gray!8}\boldmath{$\uparrow 3.94$}
& $55.74\textcolor{gray}{\text{\scriptsize±1.04}}$ 
& $53.39\textcolor{gray}{\text{\scriptsize±2.41}}$ 
& \cellcolor{gray!8}\boldmath{$\downarrow 4.21$}
& $55.89\textcolor{gray}{\text{\scriptsize±1.69}}$ 
& $45.92\textcolor{gray}{\text{\scriptsize±1.50}}$ 
& \cellcolor{gray!8}\boldmath{$\downarrow 17.76$}
& $61.74\textcolor{gray}{\text{\scriptsize±11.81}}$ 
& $50.05\textcolor{gray}{\text{\scriptsize±2.16}}$ 
& \cellcolor{gray!8}\boldmath{$\downarrow 18.93$}
& $59.84\textcolor{gray}{\text{\scriptsize±11.34}}$ 
& $\secondres{{43.94}}\textcolor{gray}{\text{\scriptsize±3.49}}$ 
& \cellcolor{gray!8}\boldmath{$\downarrow 26.57$} \\

& MRE(\%)
& $51.10\textcolor{gray}{\text{\scriptsize±1.28}}$ 
& $\secondres{{39.23}}\textcolor{gray}{\text{\scriptsize±1.11}}$ 
& \cellcolor{gray!8}\boldmath{$\downarrow 23.23$} 
& $45.94\textcolor{gray}{\text{\scriptsize±2.19}}$ 
& $42.69\textcolor{gray}{\text{\scriptsize±1.02}}$ 
& \cellcolor{gray!8}\boldmath{$\downarrow 7.07$}
& $46.01\textcolor{gray}{\text{\scriptsize±0.06}}$ 
& $43.63\textcolor{gray}{\text{\scriptsize±0.60}}$ 
& \cellcolor{gray!8}\boldmath{$\downarrow 5.17$}
& $43.50\textcolor{gray}{\text{\scriptsize±0.61}}$ 
& $40.41\textcolor{gray}{\text{\scriptsize±0.50}}$ 
& \cellcolor{gray!8}\boldmath{$\downarrow 7.10$}
& $49.06\textcolor{gray}{\text{\scriptsize±4.70}}$ 
& $42.76\textcolor{gray}{\text{\scriptsize±0.75}}$ 
& \cellcolor{gray!8}\boldmath{$\downarrow 12.84$}
& $48.77\textcolor{gray}{\text{\scriptsize±7.09}}$ 
& $\firstres{{36.44}}\textcolor{gray}{\text{\scriptsize±1.62}}$ 
& \cellcolor{gray!8}\boldmath{$\downarrow 25.30$} \\

 \midrule
 \multirow{4}{*}{\TrafficTwenty} 
& MAE 
& $33.25\textcolor{gray}{\text{\scriptsize±0.83}}$ 
& $\secondres{{31.91}}\textcolor{gray}{\text{\scriptsize±1.35}}$ 
& \cellcolor{gray!8}\boldmath{$\downarrow 4.03$}
& $34.00\textcolor{gray}{\text{\scriptsize±3.74}}$ 
& $35.94\textcolor{gray}{\text{\scriptsize±4.97}}$ 
& \cellcolor{gray!8}\boldmath{$\uparrow 5.71$}
& $\firstres{{29.74}}\textcolor{gray}{\text{\scriptsize±1.21}}$ 
& $42.17\textcolor{gray}{\text{\scriptsize±0.48}}$ 
& \cellcolor{gray!8}\boldmath{$\uparrow 41.86$}
& $38.91\textcolor{gray}{\text{\scriptsize±1.68}}$ 
& $42.04\textcolor{gray}{\text{\scriptsize±3.37}}$ 
& \cellcolor{gray!8}\boldmath{$\uparrow 8.04$}
& $43.06\textcolor{gray}{\text{\scriptsize±0.92}}$ 
& $45.77\textcolor{gray}{\text{\scriptsize±0.15}}$ 
& \cellcolor{gray!8}\boldmath{$\uparrow 6.29$}
& $35.91\textcolor{gray}{\text{\scriptsize±2.16}}$ 
& $32.78\textcolor{gray}{\text{\scriptsize±2.80}}$ 
& \cellcolor{gray!8}\boldmath{$\downarrow 8.72$} \\

& RMSE
& $56.56\textcolor{gray}{\text{\scriptsize±0.99}}$ 
& $\secondres{{56.49}}\textcolor{gray}{\text{\scriptsize±2.54}}$ 
& \cellcolor{gray!8}\boldmath{$\downarrow 0.12$} 
& $57.40\textcolor{gray}{\text{\scriptsize±4.89}}$ 
& $63.08\textcolor{gray}{\text{\scriptsize±8.58}}$ 
& \cellcolor{gray!8}\boldmath{$\uparrow 9.90$}
& $\firstres{{48.88}}\textcolor{gray}{\text{\scriptsize±1.95}}$ 
& $68.39\textcolor{gray}{\text{\scriptsize±1.71}}$ 
& \cellcolor{gray!8}\boldmath{$\uparrow 39.91$}
& $70.08\textcolor{gray}{\text{\scriptsize±4.91}}$ 
& $68.89\textcolor{gray}{\text{\scriptsize±5.45}}$ 
& \cellcolor{gray!8}\boldmath{$\downarrow 1.70$}
& $69.51\textcolor{gray}{\text{\scriptsize±0.85}}$ 
& $77.07\textcolor{gray}{\text{\scriptsize±2.32}}$ 
& \cellcolor{gray!8}\boldmath{$\uparrow 10.88$}
& $60.87\textcolor{gray}{\text{\scriptsize±4.59}}$ 
& $58.63\textcolor{gray}{\text{\scriptsize±4.43}}$ 
& \cellcolor{gray!8}\boldmath{$\downarrow 3.68$} \\

& MAPE(\%)
& $18.53\textcolor{gray}{\text{\scriptsize±0.75}}$ 
& $\firstres{{17.63}}\textcolor{gray}{\text{\scriptsize±0.97}}$ 
& \cellcolor{gray!8}\boldmath{$\downarrow 4.86$} 
& $17.78\textcolor{gray}{\text{\scriptsize±1.97}}$ 
& $20.28\textcolor{gray}{\text{\scriptsize±3.36}}$ 
& \cellcolor{gray!8}\boldmath{$\uparrow 14.06$}
& $19.26\textcolor{gray}{\text{\scriptsize±0.95}}$ 
& $22.95\textcolor{gray}{\text{\scriptsize±0.95}}$ 
& \cellcolor{gray!8}\boldmath{$\uparrow 19.16$}
& $25.12\textcolor{gray}{\text{\scriptsize±0.89}}$ 
& $24.94\textcolor{gray}{\text{\scriptsize±2.91}}$ 
& \cellcolor{gray!8}\boldmath{$\downarrow 0.72$}
& $24.23\textcolor{gray}{\text{\scriptsize±1.85}}$ 
& $25.15\textcolor{gray}{\text{\scriptsize±2.09}}$ 
& \cellcolor{gray!8}\boldmath{$\uparrow 3.79$}
& $19.88\textcolor{gray}{\text{\scriptsize±2.19}}$ 
& $\secondres{{17.71}}\textcolor{gray}{\text{\scriptsize±1.93}}$ 
& \cellcolor{gray!8}\boldmath{$\downarrow 10.91$} \\

& MRE(\%)
& $12.05\textcolor{gray}{\text{\scriptsize±0.30}}$ 
& $\secondres{{11.57}}\textcolor{gray}{\text{\scriptsize±0.49}}$ 
& \cellcolor{gray!8}\boldmath{$\downarrow 3.98$}
& $12.32\textcolor{gray}{\text{\scriptsize±1.36}}$ 
& $13.03\textcolor{gray}{\text{\scriptsize±1.80}}$ 
& \cellcolor{gray!8}\boldmath{$\uparrow 5.76$}
& $\firstres{{10.78}}\textcolor{gray}{\text{\scriptsize±0.44}}$ 
& $15.28\textcolor{gray}{\text{\scriptsize±0.18}}$ 
& \cellcolor{gray!8}\boldmath{$\uparrow 41.74$}
& $14.11\textcolor{gray}{\text{\scriptsize±0.61}}$ 
& $15.24\textcolor{gray}{\text{\scriptsize±1.22}}$ 
& \cellcolor{gray!8}\boldmath{$\uparrow 7.96$}
& $15.61\textcolor{gray}{\text{\scriptsize±0.33}}$ 
& $16.59\textcolor{gray}{\text{\scriptsize±0.06}}$ 
& \cellcolor{gray!8}\boldmath{$\uparrow 6.28$}
& $13.01\textcolor{gray}{\text{\scriptsize±0.78}}$ 
& $11.89\textcolor{gray}{\text{\scriptsize±1.01}}$ 
& \cellcolor{gray!8}\boldmath{$\downarrow 8.61$}\\

    \bottomrule
  \end{tabular}
  \end{sc}
  }
  \label{tab:rq1_1day}
\end{table*}

We provide more experimental results on universality (\ie, for 1-day and 2-day scenarios), as shown in Tables~\ref{tab:rq1_2day} and~\ref{tab:rq1_1day}. We observe that while the \model exhibits some negative gains in the 1-day scenario, it exhibits improved performance as the prediction length increases, demonstrating the superior generalization of our method. The remaining observations show similar analytical results to those in the main text.

\subsection{More Result on Robustness Study}
\label{appendix_robustness}

\begin{table*}[t!]
\caption{Performance under abnormal conditions with missing exogenous variable signals (1-day scenario).
}
\label{tab:mask_1day}
\renewcommand{\arraystretch}{1.6}
\resizebox{\linewidth}{!}{
\begin{sc}
\begin{tabular}{lccccccccccccc}
\toprule
\multicolumn{2}{c}{\textbf{Datasets}}
  & \multicolumn{4}{c}{\AirNineteen}
  & \multicolumn{4}{c}{\SpeedNineteen$*$}
  & \multicolumn{4}{c}{\TrafficTwenty} \\
\cmidrule(lr){1-2}\cmidrule(lr){3-6}\cmidrule(lr){7-10}\cmidrule(lr){11-14}

\multicolumn{2}{c}{\textbf{Masking Strategy}}
  & MAE & RMSE & MAPE(\%) & MRE(\%)
  & MAE & RMSE & MAPE(\%) & MRE(\%)
  & MAE & RMSE & MAPE(\%) & MRE(\%) \\
\midrule

\multirow{4}{*}{Zero}

  & 20\% &

  $9.98\textcolor{gray}{\text{\scriptsize±0.24}}$ & $\secondres{{15.19}}\textcolor{gray}{\text{\scriptsize±0.06}}$ & $40.86\textcolor{gray}{\text{\scriptsize±0.02}}$ & $38.59\textcolor{gray}{\text{\scriptsize±0.01}}$ &

  $3.81\textcolor{gray}{\text{\scriptsize±0.01}}$ & $31.80\textcolor{gray}{\text{\scriptsize±0.04}}$ & $9.28\textcolor{gray}{\text{\scriptsize±0.01}}$ & $9.91\textcolor{gray}{\text{\scriptsize±0.01}}$ &

  $33.74\textcolor{gray}{\text{\scriptsize±3.24}}$ & $62.25\textcolor{gray}{\text{\scriptsize±7.82}}$ & $\secondres{{17.81}}\textcolor{gray}{\text{\scriptsize±1.15}}$ & $12.23\textcolor{gray}{\text{\scriptsize±1.18}}$ \\

  & 40\% &

  $10.41\textcolor{gray}{\text{\scriptsize±0.26}}$ & $16.22\textcolor{gray}{\text{\scriptsize±0.52}}$ & $41.00\textcolor{gray}{\text{\scriptsize±0.34}}$ & $40.28\textcolor{gray}{\text{\scriptsize±1.03}}$ &

  $\secondres{{3.79}}\textcolor{gray}{\text{\scriptsize±0.01}}$ & $31.77\textcolor{gray}{\text{\scriptsize±0.02}}$ & $\firstres{{9.23}}\textcolor{gray}{\text{\scriptsize±0.01}}$ & $\firstres{{9.84}}\textcolor{gray}{\text{\scriptsize±0.01}}$ &

  $33.70\textcolor{gray}{\text{\scriptsize±0.40}}$ & $58.59\textcolor{gray}{\text{\scriptsize±2.51}}$ & $19.63\textcolor{gray}{\text{\scriptsize±0.77}}$ & $12.22\textcolor{gray}{\text{\scriptsize±0.14}}$ \\

  & 60\% &

  $10.33\textcolor{gray}{\text{\scriptsize±0.22}}$ & $16.26\textcolor{gray}{\text{\scriptsize±0.56}}$ & $41.71\textcolor{gray}{\text{\scriptsize±0.01}}$ & $39.98\textcolor{gray}{\text{\scriptsize±0.01}}$ &

  $3.80\textcolor{gray}{\text{\scriptsize±0.01}}$ & $31.75\textcolor{gray}{\text{\scriptsize±0.05}}$ & $9.31\textcolor{gray}{\text{\scriptsize±0.15}}$ & $9.90\textcolor{gray}{\text{\scriptsize±0.12}}$ &

  $36.22\textcolor{gray}{\text{\scriptsize±0.39}}$ & $65.67\textcolor{gray}{\text{\scriptsize±3.71}}$ & $19.90\textcolor{gray}{\text{\scriptsize±0.68}}$ & $13.13\textcolor{gray}{\text{\scriptsize±0.14}}$ \\

  & 80\% &

  $10.99\textcolor{gray}{\text{\scriptsize±0.15}}$ & $17.10\textcolor{gray}{\text{\scriptsize±0.29}}$ & $47.36\textcolor{gray}{\text{\scriptsize±4.28}}$ & $42.50\textcolor{gray}{\text{\scriptsize±0.58}}$ &

  $3.81\textcolor{gray}{\text{\scriptsize±0.03}}$ & $31.78\textcolor{gray}{\text{\scriptsize±0.06}}$ & $9.41\textcolor{gray}{\text{\scriptsize±0.13}}$ & $9.93\textcolor{gray}{\text{\scriptsize±0.08}}$ &

  $35.96\textcolor{gray}{\text{\scriptsize±1.23}}$ & $64.86\textcolor{gray}{\text{\scriptsize±4.13}}$ & $20.03\textcolor{gray}{\text{\scriptsize±1.02}}$ & $13.04\textcolor{gray}{\text{\scriptsize±0.44}}$ \\

\cmidrule(lr){2-14}

\multirow{4}{*}{Random}

  & 20\% &

  $\secondres{{9.96}}\textcolor{gray}{\text{\scriptsize±0.12}}$ & $15.40\textcolor{gray}{\text{\scriptsize±0.28}}$ & $\secondres{{40.04}}\textcolor{gray}{\text{\scriptsize±0.02}}$ & $\secondres{{38.53}}\textcolor{gray}{\text{\scriptsize±0.01}}$ &

  $3.80\textcolor{gray}{\text{\scriptsize±0.01}}$ & $31.80\textcolor{gray}{\text{\scriptsize±0.06}}$ & $\secondres{{9.25}}\textcolor{gray}{\text{\scriptsize±0.01}}$ & $\secondres{{9.88}}\textcolor{gray}{\text{\scriptsize±0.02}}$ &

  $32.91\textcolor{gray}{\text{\scriptsize±2.61}}$ & $60.30\textcolor{gray}{\text{\scriptsize±5.91}}$ & $18.01\textcolor{gray}{\text{\scriptsize±0.93}}$ & $11.93\textcolor{gray}{\text{\scriptsize±0.94}}$ \\

  & 40\% &

  $10.40\textcolor{gray}{\text{\scriptsize±0.31}}$ & $16.12\textcolor{gray}{\text{\scriptsize±0.33}}$ & $41.52\textcolor{gray}{\text{\scriptsize±0.41}}$ & $40.24\textcolor{gray}{\text{\scriptsize±0.01}}$ &

  $3.82\textcolor{gray}{\text{\scriptsize±0.03}}$ & $31.79\textcolor{gray}{\text{\scriptsize±0.04}}$ & $9.41\textcolor{gray}{\text{\scriptsize±0.01}}$ & $9.95\textcolor{gray}{\text{\scriptsize±0.08}}$ &

  $\firstres{{31.40}}\textcolor{gray}{\text{\scriptsize±1.10}}$ & $\firstres{{54.03}}\textcolor{gray}{\text{\scriptsize±1.13}}$ & $17.89\textcolor{gray}{\text{\scriptsize±1.27}}$ & $\firstres{{11.38}}\textcolor{gray}{\text{\scriptsize±0.40}}$ \\

  & 60\% &

  $10.66\textcolor{gray}{\text{\scriptsize±0.28}}$ & $16.91\textcolor{gray}{\text{\scriptsize±0.53}}$ & $43.23\textcolor{gray}{\text{\scriptsize±0.55}}$ & $41.23\textcolor{gray}{\text{\scriptsize±0.01}}$ &

  $3.80\textcolor{gray}{\text{\scriptsize±0.04}}$ & $31.76\textcolor{gray}{\text{\scriptsize±0.05}}$ & $9.31\textcolor{gray}{\text{\scriptsize±0.15}}$ & $9.88\textcolor{gray}{\text{\scriptsize±0.09}}$ &

  $34.34\textcolor{gray}{\text{\scriptsize±2.71}}$ & $61.96\textcolor{gray}{\text{\scriptsize±9.15}}$ & $18.84\textcolor{gray}{\text{\scriptsize±1.25}}$ & $12.44\textcolor{gray}{\text{\scriptsize±0.98}}$ \\

  & 80\% &

  $11.13\textcolor{gray}{\text{\scriptsize±0.14}}$ & $17.67\textcolor{gray}{\text{\scriptsize±0.19}}$ & $43.96\textcolor{gray}{\text{\scriptsize±0.01}}$ & $43.05\textcolor{gray}{\text{\scriptsize±0.55}}$ &

  $\firstres{{3.78}}\textcolor{gray}{\text{\scriptsize±0.01}}$ & $\firstres{{31.72}}\textcolor{gray}{\text{\scriptsize±0.03}}$ & $\firstres{{9.23}}\textcolor{gray}{\text{\scriptsize±0.01}}$ & $\firstres{{9.84}}\textcolor{gray}{\text{\scriptsize±0.02}}$ &

  $34.68\textcolor{gray}{\text{\scriptsize±2.80}}$ & $64.10\textcolor{gray}{\text{\scriptsize±8.08}}$ & $19.05\textcolor{gray}{\text{\scriptsize±0.96}}$ & $12.57\textcolor{gray}{\text{\scriptsize±1.02}}$ \\

\midrule

\rowcolor{gray!8}
\multicolumn{2}{c}{No Masking}

  & $\firstres{{9.33}}\textcolor{gray}{\text{\scriptsize±0.13}}$ & $\firstres{{14.24}}\textcolor{gray}{\text{\scriptsize±0.18}}$ & $\firstres{{38.57}}\textcolor{gray}{\text{\scriptsize±0.32}}$ & $\firstres{{36.09}}\textcolor{gray}{\text{\scriptsize±0.49}}$ &

  $\secondres{{3.79}}\textcolor{gray}{\text{\scriptsize±0.01}}$ & $\secondres{{31.73}}\textcolor{gray}{\text{\scriptsize±0.02}}$ & $9.30\textcolor{gray}{\text{\scriptsize±0.03}}$ & $\secondres{{9.88}}\textcolor{gray}{\text{\scriptsize±0.03}}$ &

  $\secondres{{31.91}}\textcolor{gray}{\text{\scriptsize±1.35}}$ & $\secondres{{56.49}}\textcolor{gray}{\text{\scriptsize±2.54}}$ & $\firstres{{17.63}}\textcolor{gray}{\text{\scriptsize0.97}}$ & $\secondres{{11.57}}\textcolor{gray}{\text{\scriptsize±0.49}}$ \\

\bottomrule

\end{tabular}
\end{sc}
}

\end{table*}

\begin{table*}[t!]
\caption{Performance under abnormal conditions with missing exogenous variable signals (2-day scenario).
}
\label{tab:mask_2day}
\renewcommand{\arraystretch}{1.6}
\resizebox{\linewidth}{!}{
\begin{sc}
\begin{tabular}{lccccccccccccc}
\toprule
\multicolumn{2}{c}{\textbf{Datasets}}
  & \multicolumn{4}{c}{\AirNineteen}
  & \multicolumn{4}{c}{\SpeedNineteen$*$}
  & \multicolumn{4}{c}{\TrafficTwenty} \\
\cmidrule(lr){1-2}\cmidrule(lr){3-6}\cmidrule(lr){7-10}\cmidrule(lr){11-14}

\multicolumn{2}{c}{\textbf{Masking Strategy}}
  & MAE & RMSE & MAPE(\%) & MRE(\%)
  & MAE & RMSE & MAPE(\%) & MRE(\%)
  & MAE & RMSE & MAPE(\%) & MRE(\%) \\
\midrule

\multirow{4}{*}{Zero}
  & 20\% &
  $10.48\textcolor{gray}{\text{\scriptsize±0.21}}$ & $\secondres{{15.73}}\textcolor{gray}{\text{\scriptsize±0.23}}$ & $44.37\textcolor{gray}{\text{\scriptsize±1.28}}$ & $40.54\textcolor{gray}{\text{\scriptsize±0.82}}$ &
  $3.84\textcolor{gray}{\text{\scriptsize±0.01}}$ & $31.86\textcolor{gray}{\text{\scriptsize±0.05}}$ & $9.35\textcolor{gray}{\text{\scriptsize±0.04}}$ & $9.99\textcolor{gray}{\text{\scriptsize±0.04}}$ &
  $\firstres{{50.06}}\textcolor{gray}{\text{\scriptsize±6.65}}$ & $\firstres{{89.00}}\textcolor{gray}{\text{\scriptsize±13.90}}$ & $\firstres{{26.06}}\textcolor{gray}{\text{\scriptsize±2.03}}$ & $\firstres{{18.12}}\textcolor{gray}{\text{\scriptsize±2.40}}$ \\
  & 40\% &
  $11.05\textcolor{gray}{\text{\scriptsize±0.07}}$ & $16.78\textcolor{gray}{\text{\scriptsize±0.13}}$ & $44.64\textcolor{gray}{\text{\scriptsize±1.19}}$ & $42.76\textcolor{gray}{\text{\scriptsize±0.28}}$ &
  $\secondres{{3.82}}\textcolor{gray}{\text{\scriptsize±0.03}}$ & $31.85\textcolor{gray}{\text{\scriptsize±0.05}}$ & $\secondres{{9.33}}\textcolor{gray}{\text{\scriptsize±0.06}}$ & $9.95\textcolor{gray}{\text{\scriptsize±0.08}}$ &
  $53.31\textcolor{gray}{\text{\scriptsize±0.02}}$ & $94.52\textcolor{gray}{\text{\scriptsize±0.57}}$ & $29.38\textcolor{gray}{\text{\scriptsize±0.70}}$ & $19.29\textcolor{gray}{\text{\scriptsize±0.01}}$ \\
  & 60\% &
  $11.22\textcolor{gray}{\text{\scriptsize±0.45}}$ & $17.21\textcolor{gray}{\text{\scriptsize±0.75}}$ & $46.81\textcolor{gray}{\text{\scriptsize±1.28}}$ & $43.42\textcolor{gray}{\text{\scriptsize±1.72}}$ &
  $3.83\textcolor{gray}{\text{\scriptsize±0.02}}$ & $31.84\textcolor{gray}{\text{\scriptsize±0.04}}$ & $9.38\textcolor{gray}{\text{\scriptsize±0.08}}$ & $9.97\textcolor{gray}{\text{\scriptsize±0.06}}$ &
  $54.33\textcolor{gray}{\text{\scriptsize±7.93}}$ & $96.30\textcolor{gray}{\text{\scriptsize±14.73}}$ & $28.86\textcolor{gray}{\text{\scriptsize±3.89}}$ & $19.66\textcolor{gray}{\text{\scriptsize±2.87}}$ \\
  & 80\% &
  $12.28\textcolor{gray}{\text{\scriptsize±0.00}}$ & $19.05\textcolor{gray}{\text{\scriptsize±0.00}}$ & $51.12\textcolor{gray}{\text{\scriptsize±0.00}}$ & $47.52\textcolor{gray}{\text{\scriptsize±0.00}}$ &
  $\secondres{{3.82}}\textcolor{gray}{\text{\scriptsize±0.02}}$ & $31.83\textcolor{gray}{\text{\scriptsize±0.05}}$ & $9.41\textcolor{gray}{\text{\scriptsize±0.10}}$ & $\secondres{{9.94}}\textcolor{gray}{\text{\scriptsize±0.06}}$ &
  $56.17\textcolor{gray}{\text{\scriptsize±5.74}}$ & $99.93\textcolor{gray}{\text{\scriptsize±10.67}}$ & $29.99\textcolor{gray}{\text{\scriptsize±3.51}}$ & $20.33\textcolor{gray}{\text{\scriptsize±2.07}}$ \\
\cmidrule(lr){2-14}

\multirow{4}{*}{Random}
  & 20\% &
  $\secondres{{10.44}}\textcolor{gray}{\text{\scriptsize±0.09}}$ & $15.93\textcolor{gray}{\text{\scriptsize±0.04}}$ & $\secondres{{43.41}}\textcolor{gray}{\text{\scriptsize±1.98}}$ & $\secondres{{40.41}}\textcolor{gray}{\text{\scriptsize±0.34}}$ &
  $3.83\textcolor{gray}{\text{\scriptsize±0.00}}$ & $31.87\textcolor{gray}{\text{\scriptsize±0.07}}$ & $\firstres{{9.32}}\textcolor{gray}{\text{\scriptsize±0.05}}$ & $9.96\textcolor{gray}{\text{\scriptsize±0.02}}$ &
  $51.75\textcolor{gray}{\text{\scriptsize±7.53}}$ & $92.27\textcolor{gray}{\text{\scriptsize±15.45}}$ & $26.69\textcolor{gray}{\text{\scriptsize±2.90}}$ & $18.73\textcolor{gray}{\text{\scriptsize±2.73}}$ \\
  & 40\% &
  $10.96\textcolor{gray}{\text{\scriptsize±0.12}}$ & $16.74\textcolor{gray}{\text{\scriptsize±0.27}}$ & $45.10\textcolor{gray}{\text{\scriptsize±1.23}}$ & $42.43\textcolor{gray}{\text{\scriptsize±0.47}}$ &
  $3.83\textcolor{gray}{\text{\scriptsize±0.02}}$ & $31.85\textcolor{gray}{\text{\scriptsize±0.04}}$ & $9.42\textcolor{gray}{\text{\scriptsize±0.08}}$ & $9.97\textcolor{gray}{\text{\scriptsize±0.06}}$ &
  $50.93\textcolor{gray}{\text{\scriptsize±9.79}}$ & $89.54\textcolor{gray}{\text{\scriptsize±18.50}}$ & $26.92\textcolor{gray}{\text{\scriptsize±5.07}}$ & $18.44\textcolor{gray}{\text{\scriptsize±3.54}}$ \\
  & 60\% &
  $11.06\textcolor{gray}{\text{\scriptsize±0.08}}$ & $16.95\textcolor{gray}{\text{\scriptsize±0.24}}$ & $45.94\textcolor{gray}{\text{\scriptsize±1.28}}$ & $42.81\textcolor{gray}{\text{\scriptsize±0.32}}$ &
  $\secondres{{3.82}}\textcolor{gray}{\text{\scriptsize±0.02}}$ & $31.84\textcolor{gray}{\text{\scriptsize±0.04}}$ & $9.37\textcolor{gray}{\text{\scriptsize±0.07}}$ & $9.96\textcolor{gray}{\text{\scriptsize±0.05}}$ &
  $51.76\textcolor{gray}{\text{\scriptsize±4.47}}$ & $92.00\textcolor{gray}{\text{\scriptsize±10.11}}$ & $27.44\textcolor{gray}{\text{\scriptsize±2.26}}$ & $18.73\textcolor{gray}{\text{\scriptsize±1.62}}$ \\
  & 80\% &
  $12.28\textcolor{gray}{\text{\scriptsize±0.00}}$ & $19.05\textcolor{gray}{\text{\scriptsize±0.00}}$ & $51.12\textcolor{gray}{\text{\scriptsize±0.00}}$ & $47.52\textcolor{gray}{\text{\scriptsize±0.00}}$ &
  $\secondres{{3.82}}\textcolor{gray}{\text{\scriptsize±0.01}}$ & $\secondres{{31.82}}\textcolor{gray}{\text{\scriptsize±0.06}}$ & $\firstres{{9.32}}\textcolor{gray}{\text{\scriptsize±0.04}}$ & $\secondres{{9.94}}\textcolor{gray}{\text{\scriptsize±0.03}}$ &
  $53.20\textcolor{gray}{\text{\scriptsize±6.37}}$ & $94.41\textcolor{gray}{\text{\scriptsize±14.30}}$ & $28.13\textcolor{gray}{\text{\scriptsize±2.32}}$ & $19.26\textcolor{gray}{\text{\scriptsize±2.30}}$ \\
\midrule

\rowcolor{gray!8}
\multicolumn{2}{c}{No Masking}
  & $\firstres{{10.04}}\textcolor{gray}{\text{\scriptsize±0.19}}$ & $\firstres{{15.12}}\textcolor{gray}{\text{\scriptsize±0.26}}$ & $\firstres{{42.62}}\textcolor{gray}{\text{\scriptsize±0.52}}$ & $\firstres{{38.84}}\textcolor{gray}{\text{\scriptsize±0.75}}$ &
  $\firstres{{3.80}}\textcolor{gray}{\text{\scriptsize±0.01}}$ & $\firstres{{31.78}}\textcolor{gray}{\text{\scriptsize±0.02}}$ & $\firstres{{9.32}}\textcolor{gray}{\text{\scriptsize±0.03}}$ & $\firstres{{9.89}}\textcolor{gray}{\text{\scriptsize±0.03}}$ &
  $\secondres{{50.59}}\textcolor{gray}{\text{\scriptsize±0.04}}$ & $\secondres{{89.10}}\textcolor{gray}{\text{\scriptsize±7.36}}$ & $\secondres{{26.68}}\textcolor{gray}{\text{\scriptsize±1.69}}$ & $\secondres{{18.31}}\textcolor{gray}{\text{\scriptsize±1.14}}$\\
\bottomrule
\end{tabular}
\end{sc}
}
\end{table*}

We provide more experimental results on robustness (\ie, for 1-day and 2-day scenarios) in Tables~\ref{tab:mask_1day} and~\ref{tab:mask_2day}. We observe similar analysis as in the main text.

\subsection{More Result on Mechanism Study}

\subsubsection{Ablation Study.}
\label{appendix_ablation}

\begin{table}[htbp!] 
    \centering
    \begin{minipage}{0.48\linewidth}
        \centering
        
  \centering
  \caption{Ablation study from the data perspective on \AirNineteen~for 2-day forecasting, where P, F, and D represent Past, Future, and Data exogenous variables respectively.}
  \label{tab:performance_2day}
  \small
  \renewcommand{\arraystretch}{1.6}
  \resizebox{1.0\linewidth}{!}{
  \begin{tabular}{ccc|cccc}
    \toprule
    \textbf{P} & \textbf{F} & \textbf{D} & \textbf{MAE} & \textbf{RMSE} & \textbf{MAPE (\%)} & \textbf{MRE (\%)} \\
    \midrule
    \gcmark & - & - & $14.71\textcolor{gray}{\text{\scriptsize±0.98}}$ & $21.53\textcolor{gray}{\text{\scriptsize±0.82}}$ & $74.15\textcolor{gray}{\text{\scriptsize±8.44}}$ & $56.91\textcolor{gray}{\text{\scriptsize±3.79}}$ \\
    - & \gcmark & - & $10.71\textcolor{gray}{\text{\scriptsize±0.40}}$ & $16.05\textcolor{gray}{\text{\scriptsize±0.55}}$ & $45.33\textcolor{gray}{\text{\scriptsize±2.40}}$ & $41.44\textcolor{gray}{\text{\scriptsize±1.54}}$ \\
    - & - & \gcmark & $14.42\textcolor{gray}{\text{\scriptsize±0.81}}$ & $21.34\textcolor{gray}{\text{\scriptsize±1.01}}$ & $69.23\textcolor{gray}{\text{\scriptsize±8.53}}$ & $55.81\textcolor{gray}{\text{\scriptsize±3.10}}$ \\
    \gcmark & - & \gcmark & $14.35\textcolor{gray}{\text{\scriptsize±0.24}}$ & $21.24\textcolor{gray}{\text{\scriptsize±0.52}}$ & $71.57\textcolor{gray}{\text{\scriptsize±1.48}}$ & $55.52\textcolor{gray}{\text{\scriptsize±0.92}}$ \\
    - & \gcmark & \gcmark & $10.32\textcolor{gray}{\text{\scriptsize±0.02}}$ & $\secondres{{15.61}}\textcolor{gray}{\text{\scriptsize±0.02}}$ & $43.22\textcolor{gray}{\text{\scriptsize±0.13}}$ & $39.93\textcolor{gray}{\text{\scriptsize±0.08}}$ \\
    \gcmark & \gcmark & - & $\secondres{{10.27}}\textcolor{gray}{\text{\scriptsize±0.33}}$ & $15.67\textcolor{gray}{\text{\scriptsize±0.57}}$ & $\firstres{{42.59}}\textcolor{gray}{\text{\scriptsize±0.92}}$ & $\secondres{{39.73}}\textcolor{gray}{\text{\scriptsize±1.27}}$ \\
    \midrule
    \rowcolor{gray!8}
    \gcmark & \gcmark & \gcmark & $\firstres{{10.04}}\textcolor{gray}{\text{\scriptsize±0.19}}$ & $\firstres{{15.12}}\textcolor{gray}{\text{\scriptsize±0.26}}$ & $\secondres{{42.62}}\textcolor{gray}{\text{\scriptsize±0.52}}$ & $\firstres{{38.84}}\textcolor{gray}{\text{\scriptsize±0.75}}$ \\
    \bottomrule
  \end{tabular}
  }

    \end{minipage}
    \hfill 
    \begin{minipage}{0.48
    \linewidth}
        \centering
  \centering
  \caption{Ablation study from the data perspective on \AirNineteen~for 3-day forecasting, where P, F, and D represent Past, Future, and Data exogenous variables respectively.}
  
  \label{tab:performance_3day}
  \small
  \renewcommand{\arraystretch}{1.6}
  \resizebox{1.0\linewidth}{!}{
  \begin{tabular}{ccc|cccc}
    \toprule
    \textbf{P} & \textbf{F} & \textbf{U} & \textbf{MAE} & \textbf{RMSE} & \textbf{MAPE (\%)} & \textbf{MRE (\%)} \\
    \midrule
    \gcmark & - & - & $14.26\textcolor{gray}{\text{\scriptsize±0.78}}$ & $21.13\textcolor{gray}{\text{\scriptsize±0.77}}$ & $71.30\textcolor{gray}{\text{\scriptsize±6.43}}$ & $55.23\textcolor{gray}{\text{\scriptsize±3.02}}$ \\
    - & \gcmark & - & $10.94\textcolor{gray}{\text{\scriptsize±0.28}}$ & $16.25\textcolor{gray}{\text{\scriptsize±0.38}}$ & $46.23\textcolor{gray}{\text{\scriptsize±1.60}}$ & $42.38\textcolor{gray}{\text{\scriptsize±1.08}}$ \\
    - & - & \gcmark & $14.55\textcolor{gray}{\text{\scriptsize±0.63}}$ & $21.42\textcolor{gray}{\text{\scriptsize±0.60}}$ & $70.16\textcolor{gray}{\text{\scriptsize±9.39}}$ & $56.33\textcolor{gray}{\text{\scriptsize±2.44}}$ \\
    \gcmark & - & \gcmark & $14.20\textcolor{gray}{\text{\scriptsize±0.10}}$ & $20.83\textcolor{gray}{\text{\scriptsize±0.35}}$ & $69.91\textcolor{gray}{\text{\scriptsize±0.12}}$ & $55.02\textcolor{gray}{\text{\scriptsize±0.38}}$ \\
    - & \gcmark & \gcmark & $10.54\textcolor{gray}{\text{\scriptsize±0.10}}$ & $\secondres{{15.83}}\textcolor{gray}{\text{\scriptsize±0.12}}$ & $\secondres{{44.59}}\textcolor{gray}{\text{\scriptsize±0.19}}$ & $40.84\textcolor{gray}{\text{\scriptsize±0.39}}$ \\
    \gcmark & \gcmark & - & $\secondres{{10.49}}\textcolor{gray}{\text{\scriptsize±0.28}}$ & $15.85\textcolor{gray}{\text{\scriptsize±0.47}}$ & $\firstres{{43.71}}\textcolor{gray}{\text{\scriptsize±0.70}}$ & $\secondres{{40.62}}\textcolor{gray}{\text{\scriptsize±1.08}}$ \\
    \midrule
    \rowcolor{gray!8}
    \gcmark & \gcmark & \gcmark & $\firstres{{10.41}}\textcolor{gray}{\text{\scriptsize±0.12}}$ & $\firstres{{15.51}}\textcolor{gray}{\text{\scriptsize±0.16}}$ & $44.65\textcolor{gray}{\text{\scriptsize±0.08}}$ & $\firstres{{40.34}}\textcolor{gray}{\text{\scriptsize±0.47}}$ \\
    \bottomrule
  \end{tabular}
  }
    \end{minipage}
\end{table}

From the data perspective, we also evaluated the impact of exogenous variables on \model's performance for 2-day and 3-day forecasting on the \AirNineteen~, as shown in Tables~\ref{tab:performance_2day} and Tables~\ref{tab:performance_3day}. We have further observation:
\ding{182} Consistent with the 1-day findings, future covariates outperform past covariates, while data variables alone produce limited accuracy but enhance performance when combined with others, with the best results achieved by integrating all variables, demonstrating \model's ability to synthesize diverse temporal information.
\ding{184} In 2-day forecasting, combining past and future covariates significantly improves percentage-based metrics, indicating a balanced contribution from historical and forward-looking signals.
\ding{183} For 3-day forecasting, combining future and data covariates almost matches the full model, highlighting the significant role of the immediate temporal context in short-term predictions.  
\ding{185} Across 1-day to 3-day horizons, the most critical trend is increasing the importance of the past covariates as the forecast horizon extends, complementing the dominant predictive power of future covariates.





\subsubsection{Strategy Study.}
\label{appendix_strategy}


\begin{figure*}[htbp!]
    \centering
    \begin{subfigure}{0.45\linewidth}
        \centering
        \includegraphics[width=\linewidth]{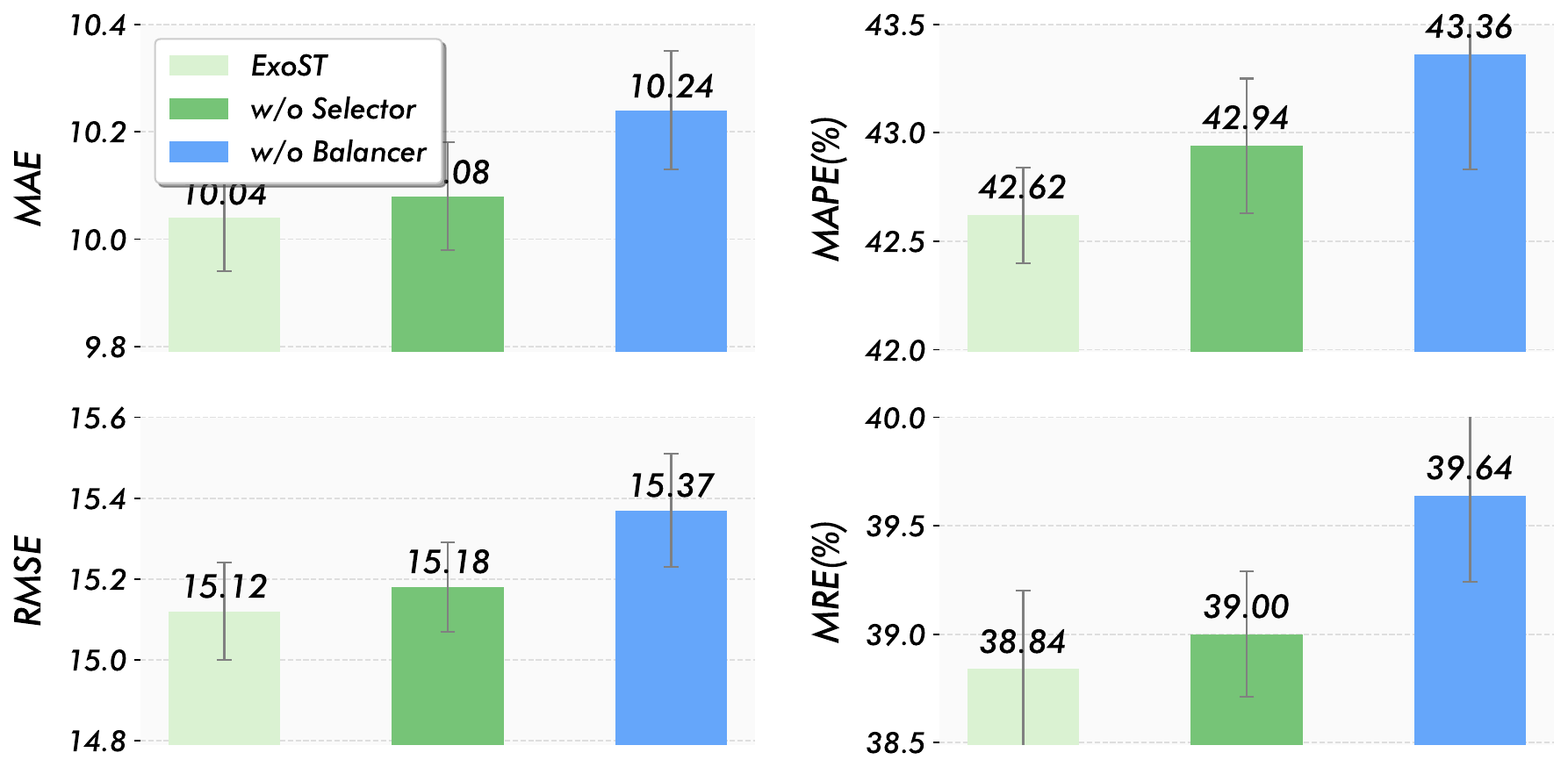}
        \caption{Ablation (2-day)}
        \label{fig:ablation_study_model_2day}
    \end{subfigure}
    \hfill 
    \begin{subfigure}{0.45\linewidth}
        \centering
        \includegraphics[width=\linewidth]{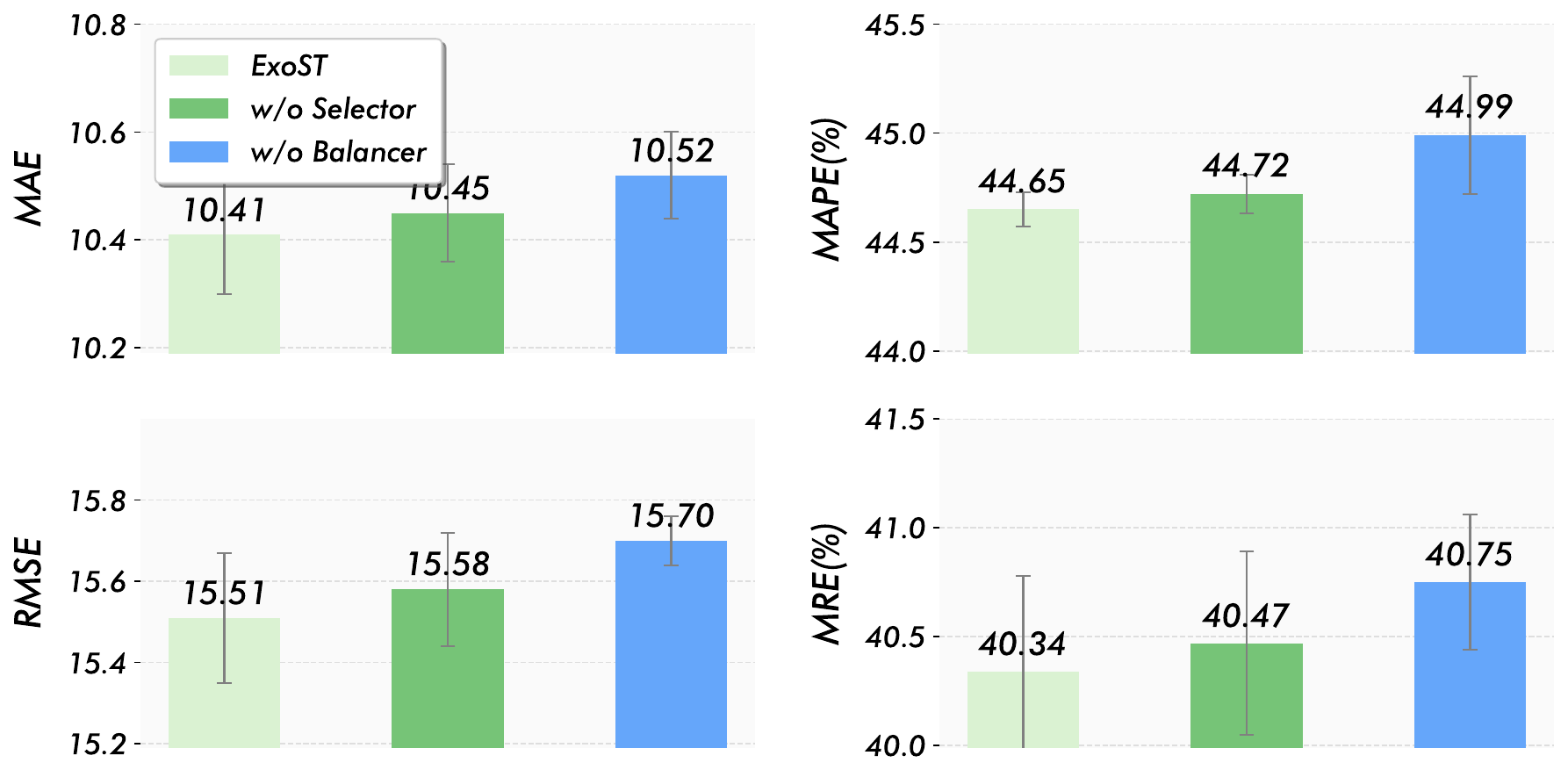}
        \caption{Ablation (3-day)}
        \label{fig:ablation_study_model_3day}
    \end{subfigure}

    \vspace{0.3cm} 

    \begin{subfigure}{0.45\linewidth}
        \centering
        \includegraphics[width=\linewidth]{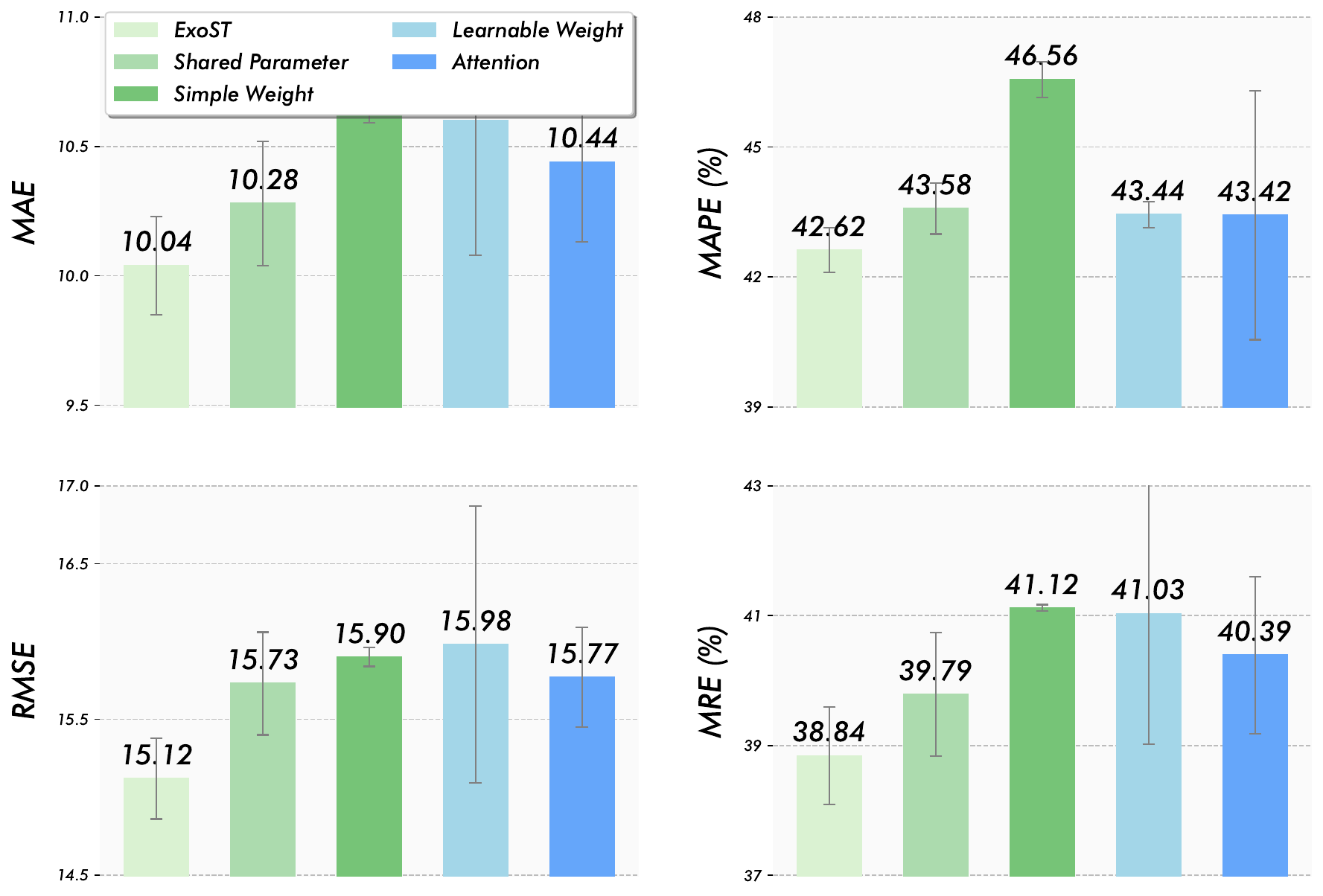}
        \caption{Strategy (2-day)}
        \label{fig:strategy_comparison2}
    \end{subfigure}
    \hfill
    \begin{subfigure}{0.45\linewidth}
        \centering
        \includegraphics[width=\linewidth]{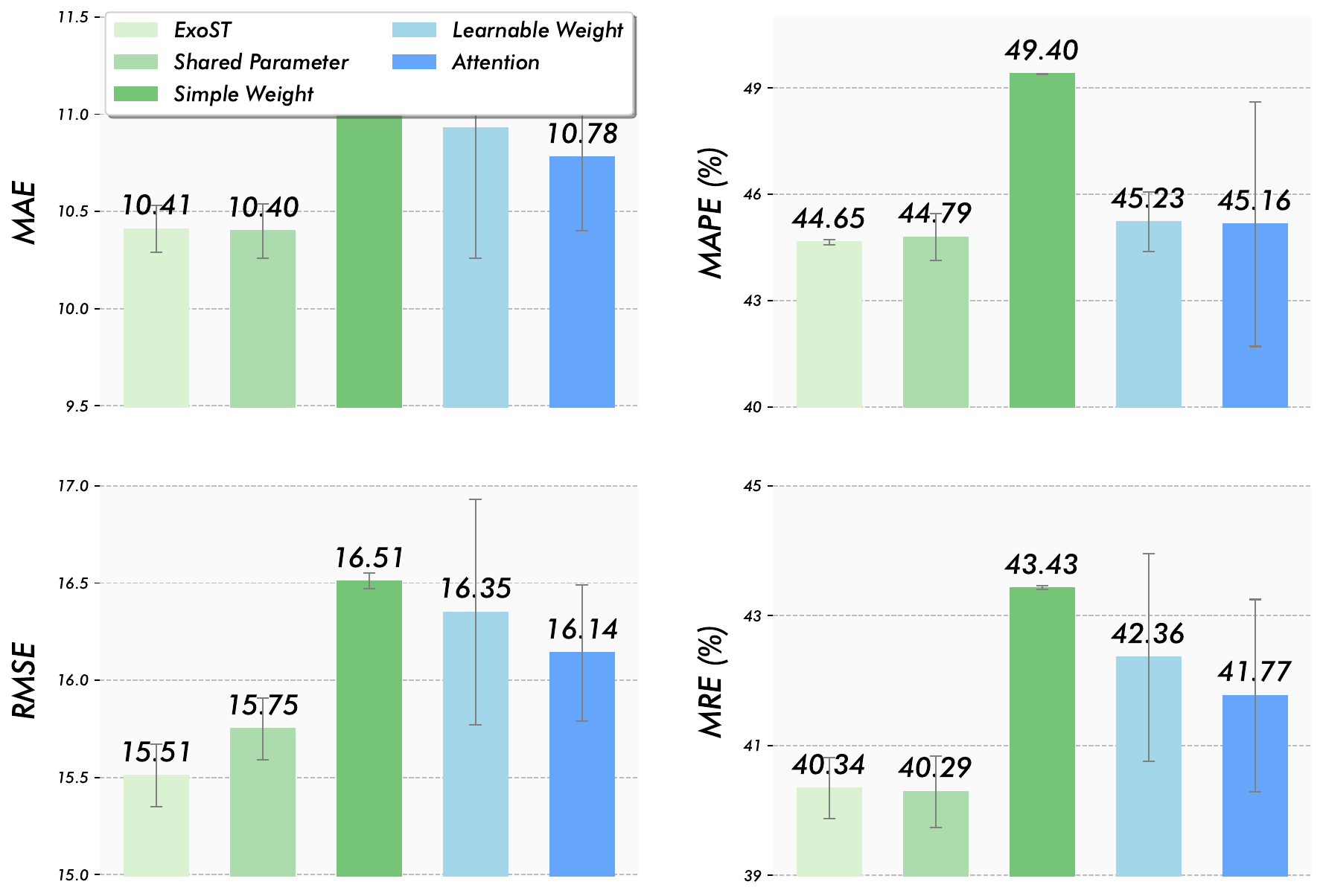}
        \caption{Strategy (3-day)}
        \label{fig:strategy_comparison3}
    \end{subfigure}

    \caption{Combined analysis on \AirNineteen~dataset. (a)-(b) Ablation studies; (c)-(d) Strategy comparisons.}
    \label{fig:combined_analysis}
\end{figure*}

We first formalize the four strategies in detail through formulas:

\begin{itemize}[noitemsep, topsep=8pt, partopsep=0pt, leftmargin=6mm,parsep=8pt]
\setlength\itemsep{0mm}
    \item \textit{Sharing Parameter}:
A single spatio-temporal encoder is re-used for both exogenous streams:
$$
\mathbf{Y}^{p}= \phi_{\text{st}}^{\text{shared}}\!\bigl(\mathbf{X}^{p'}\bigr),\qquad
\mathbf{Y}^{f}= \phi_{\text{st}}^{\text{shared}}\!\bigl(\mathbf{X}^{f'}\bigr)
$$
The forecast is obtained by averaging the two outputs:
$$
\hat{\mathbf{Y}}=\frac{1}{2}\mathbf{Y}^{p}+\frac{1}{2}\mathbf{Y}^{f}.
$$
    \item \textit{Simple Weight}:
Each stream is processed by its own encoder~\ref{sec:select}.  
The final prediction employs fixed, equal weights:
$$
\hat{\mathbf{Y}}=\frac{1}{2}\,\mathbf{Y}^{p}+\frac{1}{2}\,\mathbf{Y}^{f}.
$$
    \item \textit{Learnable Weight}:
Let $\mathbf{w}_{init}\in\mathbb{R}^{2}$ be a learnable vector, normalized by softmax:
$$
\mathbf{w}= \operatorname{Softmax}(\mathbf{w_{init}}),\qquad w_0+w_1=1.
$$
Given the two outputs of the backbone, the fusion is
$$
\hat{\mathbf{Y}}=w_0\,\mathbf{Y}^{p}+w_1\,\mathbf{Y}^{f}+\bigl(\mathbf{Y}^{p}+\mathbf{Y}^{f}\bigr),
$$
where the last term is a residual connection.
    \item \textit{Attention}:
Let $\mathbf{Y}^{p},\mathbf{Y}^{f}$ denote the intermediate hidden states produced by the two branches before the final read-out layer.  
For each directional attention we use shared projection matrices  
$\mathbf{W}_{q},\mathbf{W}_{k},\mathbf{W}_{v}\in\mathbb{R}^{H\times H}$.
\textit{Future $\rightarrow$ Past}

$$
\begin{aligned}
\mathbf{Q}^{f}&=\mathbf{Y}^{f}\mathbf{W}_{q},\quad
\mathbf{K}^{p}=\mathbf{Y}^{p}\mathbf{W}_{k},\quad
\mathbf{V}^{p}=\mathbf{Y}^{p}\mathbf{W}_{v},\\[2pt]
\boldsymbol{\alpha}_{f\rightarrow p}&=
\operatorname{softmax}\!\left(
\frac{\sum_{h=1}^{H}\bigl(\mathbf{Q}^{f}_{:,t,h}\!\cdot\!\mathbf{K}^{p}_{:,t,h}\bigr)}{\sqrt{H}}
\right),\\[2pt]
\mathbf{A}_{f\rightarrow p}&=\boldsymbol{\alpha}_{f\rightarrow p}\odot\mathbf{V}^{p},\\[4pt]
\mathbf{Y}^{p}_{\text{enh}}&=\mathbf{Y}^{p}+\mathbf{A}_{f\rightarrow p}.
\end{aligned}
$$

\textit{Past $\rightarrow$ Future}
$$
\begin{aligned}
\mathbf{Q}^{p}&=\mathbf{Y}^{p}\mathbf{W}_{q},\quad
\mathbf{K}^{f}=\mathbf{Y}^{f}\mathbf{W}_{k},\quad
\mathbf{V}^{f}=\mathbf{Y}^{f}\mathbf{W}_{v},\\[2pt]
\boldsymbol{\alpha}_{p\rightarrow f}&=
\operatorname{softmax}\!\left(
\frac{\sum_{h=1}^{H}\bigl(\mathbf{Q}^{p}_{:,t,h}\!\cdot\!\mathbf{K}^{f}_{:,t,h}\bigr)}{\sqrt{H}}
\right),\\[2pt]
\mathbf{A}_{p\rightarrow f}&=\boldsymbol{\alpha}_{p\rightarrow f}\odot\mathbf{V}^{f},\\[4pt]
\mathbf{Y}^{f}_{\text{enh}}&=\mathbf{Y}^{f}+\mathbf{A}_{p\rightarrow f}.
\end{aligned}
$$

\textit{Final Prediction}
$$
\hat{\mathbf{Y}}=\frac{1}{2}\mathbf{Y}^{p}_{\text{enh}}+\frac{1}{2}\mathbf{Y}^{f}_{\text{enh}},
$$
\end{itemize}

As Figures~\ref{fig:combined_analysis} show, we can observe that: \ding{182} In the 2-day horizon, \model maintains its advantage, while \textit{Simple Weight} and \textit{Learnable Weight} exhibit amplified weaknesses: the former's static weighting fails to handle increased uncertainty in longer forecasts, and the latter, despite its adaptable weights, amplifies noise from future exogenous signals due to lacking day-to-day context modeling.
\ding{183} Although all methods show increasing errors across horizons of 1-day to 3-day, \model consistently suppresses error accumulation better than alternatives.

\section{More Discussion}\label{disscuison}

\subsection{Current Limitation}

In this paper, we propose a new framework for modeling exogenous variables in spatio-temporal forecasting: \model. This ``select, then balance'' design addresses the inconsistent effects of variables and the imbalanced effects of types. Extensive experiments demonstrate the superiority of our method in terms of effectiveness, universality, and robustness. 
Although we have made some attempts in this direction, there are still some limitations that need to be addressed:

\begin{itemize}[noitemsep, topsep=8pt, partopsep=0pt, leftmargin=6mm,parsep=8pt]
\setlength\itemsep{0mm}
    \item First, our method assumes that spatio-temporal exogenous variables are available in real-time and are synchronized with the target system. In reality, it may not be possible to acquire real-time exogenous information at all times. Even when available, differing equipment may lead to inconsistent and mismatched sampling rates. While we conducted preliminary experiments using random masking to simulate asynchronous and non-real-time data, and our model showed some robustness, a deeper investigation into this issue is left for future work.
    \item Second, our model currently assumes that spatio-temporal exogenous variables are numerical sensor signals. However, real-world scenarios often involve diverse, heterogeneous exogenous information, such as language and visual information~\citep{chen2024terra,luo2024towards,chen2025climateiqa}. Recent research has explored using Large Language Models (LLMs) for time series~\citep{jin2024position,pan2024s,suntest,jia2024gpt4mts,chang2025llm4ts,liu2025timecma} and spatio-temporal forecasting~\citep{liu2024spatial,li2024urbangpt,liu2025st,xu2025gpt4tfp}, hoping that LLMs can gain numerical prediction capabilities through various alignment techniques. However, some studies have raised significant doubts about these approaches~\citep{tan2024language}. Different from these views, we believe that language can be more effectively viewed as sparse, discrete exogenous auxiliary information. The key challenge lies in how to leverage this information to enhance various ST tasks.
\end{itemize}

\subsection{Future Direction}

Based on these analyses, we identify two key areas for future work:

\begin{itemize}[noitemsep, topsep=8pt, partopsep=0pt, leftmargin=6mm,parsep=8pt]
\setlength\itemsep{0mm}
    \item For spatio-temporal forecasting tasks, we believe the core challenge is to use the semantic representations of exogenous variables to provide an auxiliary spatio-temporal backbone network with interventional information, allowing it to respond to dynamic changes. This approach is distinct from using language models as the primary prediction backbone.
    \item For spatio-temporal reasoning tasks~\citep{han2025large,li2025stbench,hettige2025modular,ni2026streasoner}, we believe the core essence is to activate the intrinsic capabilities of language models to analyze the underlying causes of ST changes and provide logical knowledge-based answers.
\end{itemize}

\end{document}